\pdfoutput=1

\documentclass[11pt]{article}

\usepackage[final]{acl}

\usepackage{amsmath,amsfonts,bm}

\def\eqref#1{equation~\ref{#1}}

\def\1{\bm{1}}

\def\vo{{\bm{o}}}

\def\vx{{\bm{x}}}

\def\mD{{\bm{D}}}

\def\mU{{\bm{U}}}

\DeclareMathAlphabet{\mathsfit}{\encodingdefault}{\sfdefault}{m}{sl}
\SetMathAlphabet{\mathsfit}{bold}{\encodingdefault}{\sfdefault}{bx}{n}

\usepackage{array}
\usepackage{xspace}

\newcolumntype{L}[1]{>{\raggedright\let\newline\\\arraybackslash\hspace{0pt}}m{#1}}
\newcolumntype{C}[1]{>{\centering\let\newline\\\arraybackslash\hspace{0pt}}m{#1}}
\newcolumntype{R}[1]{>{\raggedleft\let\newline\\\arraybackslash\hspace{0pt}}m{#1}}

\makeatletter
\def\adl@drawiv#1#2#3{%
        \hskip.5\tabcolsep
        \xleaders#3{#2.5\@tempdimb #1{1}#2.5\@tempdimb}%
                #2\z@ plus1fil minus1fil\relax
        \hskip.5\tabcolsep}
\newcommand{\cdashlinelr}[1]{%
  \noalign{\vskip\aboverulesep
           \global\let\@dashdrawstore\adl@draw
           \global\let\adl@draw\adl@drawiv}
  \cdashline{#1}
  \noalign{\global\let\adl@draw\@dashdrawstore
           \vskip\belowrulesep}}
\makeatother

\makeatletter
\DeclareRobustCommand\onedot{\futurelet\@let@token\@onedot}
\def\@onedot{\ifx\@let@token.\else.\null\fi\xspace}

\def\eg{e.g\onedot,\xspace} 

\makeatother

\newcommand{\modelfine}{\textsc{FineTune}\xspace}
\newcommand{\modeladapter}{\textsc{Adapter}\xspace}
\newcommand{\modelpropetl}{\textsc{ProPETL}\xspace}
\newcommand{\modelcompact}{\textsc{Compacter++}\xspace}
\newcommand{\modellora}{\textsc{LoRA}\xspace}
\newcommand{\modelia}{\((IA)^3\)\xspace}
\newcommand{\modelfineMT}{\textsc{FineTune-m}\xspace}
\newcommand{\modeladapterMT}{\textsc{Adapter-m}\xspace}
\newcommand{\modelpropetlMT}{\textsc{ProPETL-m}\xspace}

\newcommand{\modelhyper}{\textsc{HyperFormer}\xspace}
\newcommand{\modelhyperPP}{\textsc{HyperFormer++}\xspace}

\newcommand{\modelours}{\textsc{ScaLearn}\xspace}
\newcommand{\modeladapterfusion}{\textsc{AdapterFusion}\xspace}
\newcommand{\modeladaptersoup}{\textsc{AdapterSoup}\xspace}
\newcommand{\modeladapteroursVecShare}{\textsc{ScaLearn++}\xspace}
\newcommand{\modeladapteroursVecSplit}{\textsc{ScaLearn}\xspace}
\newcommand{\modeladapteroursScalarShare}{\textsc{ScaLearnUniform++}\xspace}
\newcommand{\modeladapteroursScalarSplit}{\textsc{ScaLearnUniform}\xspace}

\usepackage{times}
\usepackage{latexsym}

\usepackage[T1]{fontenc}

\usepackage[utf8]{inputenc}

\usepackage{microtype}

\usepackage{inconsolata}

\usepackage{xcolor}

\usepackage{graphicx}
\usepackage{tabularx}
\usepackage{amsmath}
\usepackage{amssymb}
\usepackage{booktabs}
\usepackage{multirow}
\usepackage{caption}
\usepackage{subcaption}
\usepackage{algorithm}
\usepackage{algpseudocode}
\usepackage{amsfonts}
\usepackage{verbatim}
\usepackage{tikzducks}
\usepackage{adjustbox}
\usepackage{arydshln}
\usepackage{rotating,booktabs}
\usepackage{makecell}
\title{\textsc{ScaLearn}: Simple and Highly Parameter-Efficient Task Transfer \\by Learning to Scale}

\newcommand{\rev}[1]{\textcolor{black}{#1}}

\author{
Markus Frohmann\textsuperscript{\textnormal{1,2}}~~~~
Carolin Holtermann\textsuperscript{\textnormal{3}}~~~~
Shahed Masoudian\textsuperscript{\textnormal{1,2}}\\
~\textbf{Anne Lauscher}\textsuperscript{\textnormal{3}}~~~~
\textbf{Navid Rekabsaz}\textsuperscript{\textnormal{4}} \\
~\textsuperscript{1} Johannes Kepler University Linz, ~\textsuperscript{2} Linz Institute of Technology, AI Lab\\
~\textsuperscript{3} Data Science Group, Universität Hamburg
\\
~\textsuperscript{4} Thomson Reuters Labs, Zug, Switzerland
\\
~\small{\texttt{\{markus.frohmann, shahed.masoudian\}@jku.at}}, 
~\small{\texttt{\{carolin.holtermann, anne.lauscher\}@uni-hamburg.de}} \\
~\small{\texttt{navid.rekabsaz@thomsonreuters.com}}\\
}

\begin{document}
\maketitle
\begin{abstract}
Multi-task learning (MTL) has shown considerable practical benefits, particularly when using language models (LMs).
While this is commonly achieved by learning $n$ tasks under a joint optimization procedure, some methods, such as AdapterFusion, divide the problem into two stages: (i) task learning, where knowledge specific to a task is encapsulated within sets of parameters (\eg adapters), and (ii) transfer, where this already learned knowledge is leveraged for a target task. 
This separation of concerns provides numerous benefits (e.g., promoting reusability). 
However, current two-stage MTL introduces a substantial number of additional parameters.
We address this issue by leveraging the usefulness of linearly scaling the output representations of source adapters for transfer learning.
We introduce \modelours, a simple and highly parameter-efficient two-stage MTL method that capitalizes on the knowledge of the source tasks by learning a minimal set of scaling parameters that enable effective 
transfer to a target task. 
Our experiments on three benchmarks (GLUE, SuperGLUE, and HumSet) and two encoder LMs show that \modelours 
consistently outperforms strong baselines with a small number of transfer parameters ($\sim\!0.35\%$ of those of AdapterFusion).
Remarkably, we observe that \modelours maintains its strong abilities even when further reducing parameters, achieving 
competitive results with \emph{only 8 transfer parameters} per target task.
Our proposed approach thus demonstrates the power of simple scaling as a promise for more efficient task transfer.\footnote{Our code is available at \url{https://github.com/CPJKU/ScaLearn}.}
\end{abstract}

\section{Introduction}
\label{sec:introduction}
\begin{figure*}[t]
    \centering
        \includegraphics[width=1\textwidth]{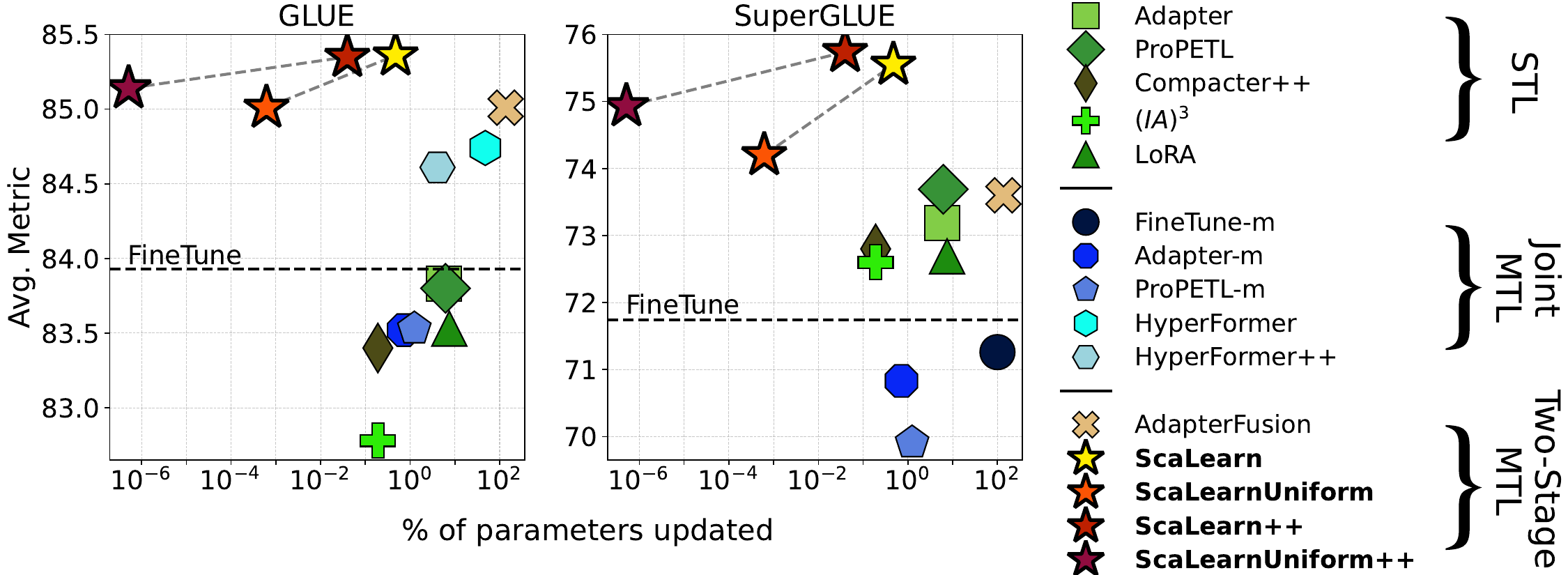}
    \caption{Performance and parameter-efficiency of single task learning (STL), and joint/two-stage MTL methods, evaluated on GLUE~\citep{GLUE} and SuperGLUE~\citep{SuperGLUE} using $\text{RoBERTa\textsubscript{BASE}}$~\citep{roberta}. \rev{The reported values for the two-stage MTL methods only consider the ones in the respective transfer layers.} The full details of the learnable parameters and performance results are provided in \S\ref{sec:results}.}
    \label{fig:fig.parameter-efficiency}
\end{figure*}

With the wide availability of pre-trained language models (LMs) as the backbone of language processing, multi-task learning (MTL) has shown significant benefits, especially for tasks with possible conceptual commonalities~\citep{ruder-mtl-overview-2, zhang-mtl-overview-3, DBLP:journals/jmlr/RaffelSRLNMZLL20}. 
The traditional paradigm in MTL is to formulate a joint optimization objective based on a set of tasks and train a single model to simultaneously learn and transfer the knowledge relevant to the tasks. This \emph{joint MTL} approach can be realized by fine-tuning an LM~\citep{liu-mtdnn, DBLP:conf/icml/Stickland019}, or, more recently, by using parameter-efficient, often modularized, MTL approaches~\citep{hyperformer, zeng-etal-2023-one, DBLP:conf/iclr/PilaultEP21, asai-attempt, ponti-etal-2023-combining, caccia2022multi}. 

As an alternative to the joint MTL paradigm, some works, such as \modeladapterfusion~\citep{pfeiffer-etal-2021-adapterfusion}, clearly distinguish task training from transfer learning, assigning dedicated parameters to each of these aspects. In this paradigm, referred to as \emph{two-stage MTL}, first each \emph{source task} is trained separately and stored into a separate module like an adapter~\citep{houlsby2019parameterefficient}, and then a task transfer layer is trained for a given \emph{target task} using information from an \emph{arbitrary} set of source tasks. This separation of concerns between task and transfer learning offers valuable benefits: (1) Learning a separate transfer layer for each target task in a two-stage MTL approach reduces the potentially destructive effects of transfer learning on specific tasks, as the transfer layer parameters corresponding to each target task can independently decide what information should be used from the available source tasks. As shown in our experiments with encoder LMs, this supports the effectiveness of transfer learning, making it less sensitive to task selection. (2) Since the source tasks can simply be taken from already trained modules (no need for re-training), two-stage approaches promote reusability -- a principle of Green AI~\citep{10.1145/3477495.3531766,10.1145/3381831}. Further, they provide a practical solution to cases involving issues such as data privacy and/or fairness constraints, as a pre-trained module can readily provide the (\eg already debiased) functionality of the source task even without the need to have access to its training data~\citep{lauscher-etal-2021-sustainable-modular,kumar-etal-2023-parameter}. 

Despite these benefits, current two-stage MTL solutions introduce significantly more learnable parameters compared to recent joint MTL ones, exacerbated by the linear increase in the number of parameters with the number of target tasks. In our experiment setup with eight target tasks using $\text{RoBERTa}_\text{BASE}$~\citep{roberta}, \modeladapterfusion introduces $\sim\!134\%$ new parameters for transfer learning, while \modelhyperPP~\citep{hyperformer} conducts joint MTL by adding $\sim\!4\%$ ($\approx5$M) trainable parameters (details in Table~\ref{tab:parameter-efficiency} and \S\ref{sec:results}). This high number of parameters is in stark contrast to the promise of Green AI given by the modularized nature of two-stage MTL.

\paragraph{Contributions.} We build on insights gained from analyzing the effects of scaling the output representations of adapters, and introduce \modelours, a novel two-stage MTL method that learns to transfer the knowledge of the source adapters using a small set of scaling parameters. For a given target task, \modelours introduces parameters that scale the output representation of each source adapter and combine the resulting \emph{scaled} representations by simply taking the element-wise sum.
This approach results in high parameter-efficiency, such that -- following the mentioned experiment setting -- \modelours only adds $\sim\!0.47\%$ ($\approx0.5$M) parameters.
We further introduce an even more parameter-efficient variant via uniform scaling (\modeladapteroursScalarSplit), where each scaling vector is reduced to a single scaling parameter. Finally, by sharing parameters across layers, we achieve our most efficient variation (\modeladapteroursScalarShare), only containing 64 parameters for transfer learning.

We conduct a large set of transfer learning experiments on the GLUE~\citep{GLUE}, SuperGLUE~\citep{SuperGLUE}, and HumSet~\citep{fekih-etal-2022-humset} benchmarks using encoder LMs, namely the popular $\text{RoBERTa}$~\citep{roberta} and $\text{XLM-R}$~\citep{conneau-etal-2020-unsupervised-xlmr} models, both in their \texttt{base} and \texttt{large} configurations.

Figure~\ref{fig:fig.parameter-efficiency} summarizes our results on GLUE and SuperGLUE, showing that \modelours, while providing high efficiency and the benefits of the two-stage MTL paradigm, consistently outperforms the baselines. The overall performance of \modelours remains highly competitive 
in its more parameter-efficient variations.
Our results also show the advantage of two-stage models in avoiding destructive effects during transfer learning.
Overall, with \modelours we leverage the power of scaling as a viable, non-destructive, simple-to-implement, and highly parameter-efficient solution to the current shortcomings of existing MTL methods, paving the future for more effective and efficient task transfer.

\if In this work, we propose a highly parameter-efficient and effective two-stage MTL method by scaling the output representations of source adapters \rev{using encoder LMs}.
Building on an analysis of scaling the output representations of adapters, we introduce \modelours, a novel two-stage MTL method that learns to transfer the knowledge of the source adapters using a small set of scaling parameters. For a given target task, \modelours introduces parameters that scale the output representation of each source adapter and combine the resulting \emph{scaled} representations by simply taking the element-wise sum.
\modelours learns to apply a (linear) scaling transformation without imposing any constraint on the relation of the scaling coefficients across source tasks using common gradient descent methods.
This approach results in high parameter-efficiency, such that -- following the mentioned experiment setting -- \modelours only adds $\sim\!0.47\%$ ($\approx0.5$M) parameters.
We further introduce an even more parameter-efficient variant via uniform scaling (\modeladapteroursScalarSplit), where each scaling vector is reduced to a single scaling parameter. Finally, by sharing parameters across layers, we achieve our most efficient variation (\modeladapteroursScalarShare), only containing 64 parameters for transfer learning.
Moreover, unlike other approaches relying, \eg on prefix-tuning~\citep{wang-MPT, asai-attempt}, \modelours does not rely on pre-selected source tasks but works with \emph{any} set of source tasks.

We conduct a large set of transfer learning experiments on the GLUE~\citep{GLUE}, SuperGLUE~\citep{SuperGLUE}, and HumSet~\citep{fekih-etal-2022-humset} benchmarks using encoder LMs, namely the popular $\text{RoBERTa}$~\citep{roberta} and $\text{XLM-R}$~\citep{conneau-etal-2020-unsupervised-xlmr} models, both in their \texttt{base} and \texttt{large} configurations.

While large language models (LLMs) have shown strong capabilities across a spectrum of tasks~\citep{gpt3,llama2}, they also tend to be very resource-intensive~\citep{DBLP:journals/corr/abs-2401-00625,DBLP:journals/corr/abs-2309-14393}. In contrast, popular encoder LMs contain considerably fewer parameters, and are widely used for a variety of tasks, \eg representation learning~\citep{DBLP:conf/nips/KusupatiBRWSRHC22}), information retrieval~\citep{DBLP:journals/corr/abs-2211-14876, bge_embedding}, sentence segmentation~\citep{DBLP:conf/acl/MinixhoferPV23}, inter alia, making them especially important for real-time use cases.
Furthermore, we compare the parameter-efficiency and performance of \modelours with strong joint and two-stage MTL baselines.
Figure~\ref{fig:fig.parameter-efficiency} summarizes our results on GLUE and SuperGLUE, showing that \modelours, while providing high efficiency and the benefits of the two-stage MTL paradigm, consistently outperforms the baselines. Interestingly, the overall performance of \modelours remains highly competitive and only marginally different in its more parameter-efficient variations. Our results also show the advantage of two-stage models in avoiding destructive effects during transfer learning, particularly on the SuperGLUE and HumSet benchmarks (cf. \S~\ref{sec:results}). 
Finally, \modelours exhibits strong performance in few-shot settings, outperforming regular adapters and \modeladapterfusion when trained only on a handful of data points. Overall, with \modelours we leverage the power of scaling as a viable, non-destructive, simple-to-implement, and highly parameter-efficient solution to the current shortcomings of two-stage and joint MTL methods, paving the future for more effective and efficient task transfer.\fi

\section{Background}\label{sec:background}
In task transfer learning, we consider a pre-trained LM as well as two sets $S$ and $T$, representing the source and target tasks, respectively. The aim of MTL is to leverage the information of tasks in $S$  to improve the generalization on tasks in $T$.

\vspace{0.5em}  
\noindent\textbf{Single Task Learning (STL).} In this basic setting, a separate set of parameters is optimized on each task ($S=T$) without any knowledge transfer between tasks. STL can be done by fine-tuning the LM parameters or by introducing more parameter-efficient modules into the model, such as adapter modules (Pfeiffer adapters~\citep{houlsby2019parameterefficient, pfeiffer-etal-2021-adapterfusion}, \modelpropetl~\citep{zeng-etal-2023-one}, or \modelcompact~\citep{compacter}),  $(IA)^3$~\citep{liu-ia3}, prefix-tuning~\citep{li-liang-2021-prefix}, or LoRA~\citep{hu2022lora}, each with $\Theta_{s}$ parameters for each task $s$.

\vspace{0.5em}  
\noindent\textbf{Joint MTL.} This approach is commonly done by having a unified model for all tasks ($S=T$), and a joint optimization objective that simultaneously optimizes the model using samples from all tasks~\citep{ruder-mtl-overview-2}. 
The general joint MTL objective can be formulated as ${\mathcal{L}_{\text{joint}}=\sum_{s=1}^{|S|} \alpha_s \mathcal{L}_{s}}$, 
where $\alpha_s$ is the sampling weight of task $s$. This optimization objective can be used to fine-tune the parameters of an LM~\citep{liu-mtdnn, DBLP:conf/icml/Stickland019, DBLP:journals/jmlr/RaffelSRLNMZLL20}, or those of a modularized architecture~\citep{hyperformer, DBLP:conf/iclr/PilaultEP21, ponti-etal-2023-combining}. Despite the benefit of having one unified model, the joint loss often causes tasks to compete with each other for learning capacity, leading to the \emph{task interference problem}~\citep{scalarization-neurips, mccloskey1989catastrophic, kirkpatrick-catastrophic}. This makes the joint MTL paradigm particularly sensitive to the selection of tasks~\citep{scalarization-neurips}, while various methods in the literature have aimed to address this issue (e.g., \citet{kendall2018multi, DBLP:conf/iclr/PilaultEP21}; a brief review is provided in \S~\ref{sec:related}).

\vspace{0.5em}  
\noindent\textbf{Two-stage MTL.} In contrast to joint MTL, two-stage MTL methods optimize each target task independently, bypassing the issue of task interference~\citep{pfeiffer-etal-2021-adapterfusion}. Similarly to STL, a parameter-efficient module is first learned for each \emph{source} task $s$ with parameters $\Theta_s$. In principle, two-stage MTL methods can simply use already pre-trained modules (such as adapters), saving the costs of re-training modules on each task. This facilitates the re-use of existing parameter-efficient modules for each source task,\footnote{E.g., through sharing platforms such as AdapterHub (\url{https://adapterhub.ml/})~\citep{adapterhub}.} which may vary in performance and/or take into account additional constraints such as fairness and bias mitigation~\citep{modularDeepL,kumar-etal-2023-parameter,lauscher-etal-2021-sustainable-modular}. Moreover, it also removes the need for accessing the training data of the source tasks (\eg due to data privacy) so far as the source task's functionality is solely provided via parameter-efficient modules. 
Next, given $|S|$ (pre-trained and frozen) source task modules, two-stage MTL methods define and optimize a transfer layer for each target task to leverage the knowledge of source tasks to solve the target task. This stage introduces $\Omega_t$ new parameters for each target task $t$. 


\modeladapterfusion~\citep{pfeiffer-etal-2021-adapterfusion} introduces an implementation of the two-stage approach with strong performance~\citep{modularDeepL}.
It uses an attention mechanism as its transfer layer, inserted into each LM layer after the source adapters. More specifically, given the output vector of each source adapter $s$ in each layer $l$, referred to as $\vo^{l}_{s}$, the attention layer (with target task $t$ as query and source tasks $S$ as keys and values) learns to assign a weight $\omega^{l}_{s}$ to each source task. The final output of the target task $t$ in this layer is calculated as:

\begin{equation}
    \label{eq:adapterfusion}
    \vo^{l}_{t} = \sum_{s=1}^{|S|} \omega^{l}_{s}\vo^{l}_{s},~~~~ \text{where}~ \sum_{s=1}^{|S|} \omega^{l}_{s}=1
\end{equation}

Regardless of how the weights are calculated, the method can be seen as a weighted summation of source output vectors, where the weights form a categorical probability distribution. 


\section{\modelours \xspace-- Learning to Scale for Knowledge Transfer}
\label{sec:method}

To understand the effect of scaling the output representations of adapters, we conducted initial experiments on scaling them, both in isolation and when combining two of them. In these experiments, we observed that (1)~scaling output vectors is an effective method for controlling the (partial or full) activation of the knowledge contained in an adapter module; (2)~an optimal configuration of the scaling parameter will, in many cases, lead to superior results on the target task; (3)~the optimal weights do not necessarily sum up to 1. 
These findings stand in contrast to the established practice of forcing the coefficients to sum up to 1 (\eg as in \modeladapterfusion; cf. Equation~\ref{eq:adapterfusion}).
We provide comprehensive results and analyses in Appendix~\ref{sec:analyses}.
Overall, these observations provide strong motivation for a method to \emph{combine} representations from several adapters by scaling their output representations.

Based on that, we present \modelours, a novel two-stage transfer learning method to combine the knowledge of source adapters by scaling their output representations.
Our core contribution regards the transfer layer, built on the output of the tasks' modular networks. Similar to \citet{pfeiffer-etal-2021-adapterfusion}, we utilize adapter modules for the task learning layer. In particular, the output representation of the adapter of source task $s$ at layer $l$ is defined as: ${\vo^{l}_{s}=\mU^l_s(\text{ReLU}(\mD^l_s(\vx^{l}_{s}))) + \vx^{l}_{s}}$, where $\vx^{l}_{s}$ is the input vector, and $\mU^l_s$ and $\mD^l_s$ denote the up- and down-projection parameter matrices, respectively.

Our introduced \modelours linearly scales and combines the output representations of source adapters, $\vo^{l}_{1}, \dots, \vo^{l}_{|S|}$, to achieve the objective of target task $t$. 

We define two variations of the scaling operation: \emph{non-uniform} which applies a scaling vector to each output vector using the element-wise product (\modelours), and the more parameter-efficient \emph{uniform} that scales each vector only with a scalar parameter (\modeladapteroursScalarSplit). These variations are formulated below:

\begin{equation}\label{eq:eq3}
    \begin{aligned}
        \modelours:~\vo^{l}_{t} &= \sum_{s=1}^{|S|} \bm{\omega}^{l}_{s}\odot\vo^{l}_{s} \\
        \modeladapteroursScalarSplit:~\vo^{l}_{t} &= \sum_{s=1}^{|S|} \omega^{l}_{s}\vo^{l}_{s},
    \end{aligned}
\end{equation}

where $\odot$ denotes the Hadamard product, and $\bm{\omega}^{l}_{s}$ and $\omega^{l}_{s}$ are learnable vector and scalar parameters, respectively. 
Inspired by previous studies~\citep{compacter, zeng-etal-2023-one, albert, bert-syntactic-layer}, we further increase parameter-efficiency by learning shared scaling parameters among all layers, formulated as follows: 
\begin{equation}\label{eq:eq4}
    \begin{aligned}
        \modeladapteroursVecShare:~\vo^{l}_{t} &= \sum_{s=1}^{|S|} \bm{\omega}_{s}\odot\vo^{l}_{s} \\
        \modeladapteroursScalarShare:~\vo^{l}_{t} &= \sum_{s=1}^{|S|} \omega_{s}\vo^{l}_{s},
    \end{aligned}
\end{equation}

where, similarly, $\bm{\omega}_{s}$ and $\omega_{s}$ are learnable vector and scalar parameters, but shared among all layers. In all the mentioned methods, to optimize the transfer parameters $\Omega$, we use gradient descent as an easy-to-implement and straightforward solution. 
On the basis of our experiments, we find that our approach provides highly competitive results on a wide range of tasks (cf.  \S~\ref{sec:results}). Furthermore, \modelours models do \emph{not} force any distributional properties on the $\omega$ values, as commonly imposed in previous work~\cite{pfeiffer-etal-2021-adapterfusion,adapterSoup,scalarization-neurips} through functions such as softmax and average. 

\vspace{0.5em}  
\noindent\textbf{Parameter-efficiency of \modelours.} To have a clear view of the parameter-efficiency of the models, we continue by analyzing the number of learnable parameters in the transfer layer. The \modelours variant introduces ${d\!\times\!L\!\times\!|S|}$ transfer parameters for a single target task, where $d$ is the embedding size and $L$ denotes the number of layers. The total number of parameters for all target tasks then becomes ${d\!\times\!L\!\times\!|S|\!\times\!|T|}$. Moving to \modeladapteroursScalarSplit, this number reduces to ${L\!\times\!|S|\!\times\!|T|}$. The \modeladapteroursVecShare spares the $L$ term and has ${d\!\times\!|S|\!\times\!|T|}$ transfer parameters. Finally, the most parameter-efficient variant \modeladapteroursScalarShare only adds ${|S|\!\times\!|T|}$ parameters. For each task, new task head parameters are learned jointly with the transfer parameters.

For comparison, the number of transfer parameters of \modeladapterfusion is $3\!\times\!d^2\!\times\!L\!\times\!|T|$ (discarding bias \rev{and task head} parameters), corresponding to the query, key, and value matrices of the attention mechanism. 
Comparing the formulas, we observe that our methods are far more parameter-efficient, since in practice $|S| \ll d$, and hence the ${d\!\times\!L}$ term in \modelours becomes much smaller than ${d^2}$ in \modeladapterfusion.
Compared to the joint MTL paradigm, despite the linear increase of parameters
with $|T|$, our \modelours$\!$* models still provide high parameter-efficiency. This stems from the fact that $|T| \ll d$, and hence reducing the effect of $d$ -- which is fully eliminated in the uniform variants -- leaves a stronger impact on parameter-efficiency.



\section{Experiment Setup}
\label{sec:setup}
\vspace{0.5em}  
\noindent\textbf{Tasks and datasets.}
We conduct our experiments on the GLUE and SuperGLUE benchmarks, respectively, each consisting of $8$ tasks, as well as on the HumSet benchmark~\citep{fekih-etal-2022-humset}. HumSet is a multilingual classification dataset for humanitarian crisis response that consists of $5$ tasks. Additionally, we use a combination of \emph{all} GLUE and SuperGLUE tasks resulting in 15 datasets\footnote{The RTE task is contained in GLUE and SuperGLUE.}.
 It has been shown that tasks from GLUE and SuperGLUE particularly benefit from multi-task learning, given their partially overlapping task formulations and highly varying dataset sizes \citep{devlin-etal-2019-bert, DBLP:conf/icml/Stickland019,asai-attempt, wang-MPT}. Complete details regarding the benchmarks including their train/validation/test splits are provided in Appendix~\ref{sec:all-hyperparams}.

\vspace{0.5em}  
\noindent\textbf{LM backbones.} We use the encoder LMs $\text{RoBERTa}_\text{BASE}$ and $\text{RoBERTa}_\text{LARGE}$~\citep{roberta} on GLUE and SuperGLUE. For the experiments on HumSet, following \citet{fekih-etal-2022-humset} we utilize the commonly used multilingual encoder LMs $\text{XLM-R\textsubscript{BASE}}$ and $\text{XLM-R\textsubscript{LARGE}}$~\citep{conneau-etal-2020-unsupervised-xlmr} as it consists of multiple languages.

We put our focus on encoder LMs since they have been studied extensively and are still widely used for a variety of tasks, \eg representation learning~\citep{DBLP:conf/nips/KusupatiBRWSRHC22, DBLP:journals/corr/abs-2211-14876, bge_embedding}, sentence segmentation~\citep{DBLP:conf/acl/MinixhoferPV23}, and as language encoder as part of multi-modal architectures~\citep{multimodal3, multimodal2, multimodal1}, inter alia, especially in real-time use cases due to their efficiency and comparatively low computational demands.

\begin{table}[t]
\centering
\scalebox{1}{
\scriptsize
\begin{tabular}{l@{\hskip8pt}l @{\hskip4pt}r@{\hskip1pt}l r @{\hskip1pt}l}
    \toprule
    \textbf{Type} & \textbf{Model} & \multicolumn{2}{c}{\textbf{\makecell{Parameters\\(one task)}}} & \multicolumn{2}{c}{\textbf{\makecell{Parameters\\(all tasks)}}} \\\midrule
    \multirow{5}{*}[0pt]{\makecell[l]{STL}} & \modelfine & 100.00\% & (125M) & 800.00\% & (125M) \\
    & \modeladapter & 0.72\% & (895K) & 5.74\% & (7M) \\
    & \modelpropetl & 0.77\% & (959K) & 6.16\% & (8M) \\
    & \modelcompact & 0.02\% & (29K) & 0.19\% & (235K) \\
    & \rev{\modelia} & \rev{0.05\%} & \rev{(57K)} & \rev{0.37\%} & \rev{(455K)} \\
    & \modellora & 0.93\% & (1.2M) & 7.50\% & (9.4M) \\
    \midrule
    \multirow{5}{*}[0pt]{\makecell[l]{Joint\\MTL}} & \modelfineMT & \multicolumn{2}{c}{-} & 100.00\% & (125M) \\
    & \modeladapterMT & \multicolumn{2}{c}{-} & 0.72\% & (895K) \\
    & \modelpropetlMT & \multicolumn{2}{c}{-} & 1.24\% & (1.5M) \\
    & \modelhyper & \multicolumn{2}{c}{-} & 47.67\% & (59M) \\
    & \modelhyperPP & \multicolumn{2}{c}{-} & 4.09\% & (5M) \\
    \midrule
    & & \multicolumn{2}{c}{\textbf{\makecell{Transfer ($\Omega_t$)\\(target task $t$)}}} & \multicolumn{2}{c}{\textbf{\makecell{Transfer ($\Omega$)\\(all target tasks)}}} \\\cdashlinelr{3-6}
    \multirow{5}{*}[0pt]{\makecell[l]{Two-\\Stage\\MTL}} & 
    \modeladapterfusion & 17.05\% & (21M) & 136.40\% & (170M) \\
    & \modeladapteroursVecSplit & 0.06\% & (74K) & 0.47\% & (590K) \\
    & \modeladapteroursScalarSplit & 0.00\% & (96) & 0.00\% & (768) \\
    & \modeladapteroursVecShare & 0.00\% & (6K) & 0.04\% & (49K) \\
    & \modeladapteroursScalarShare & 0.00\% & (8) & 0.00\% & (64) \\
    \bottomrule
\end{tabular}
}
\caption{Percentage and trainable parameters per model (\rev{excluding task head parameters}) when training on 8 tasks (as in GLUE/SuperGLUE) using $\text{RoBERTa\textsubscript{BASE}}$.
}
\label{tab:parameter-efficiency}
\end{table}

\vspace{0.5em}  
\noindent\textbf{Models and baselines.} We conduct experiments on four variants of our model, namely \textbf{\modelours}, \textbf{\modeladapteroursScalarSplit}, \textbf{\modeladapteroursVecShare}, and \textbf{\modeladapteroursScalarShare}. As a direct baseline, we compare our models with \textbf{\modeladapterfusion}, a common two-stage MTL method that shares similar conceptual properties.
\rev{We also compare our models with \textbf{\modeladaptersoup}~\citep{adapterSoup}, performing weight-space averaging over adapter weights of the $5$ most similar tasks according to their sentence similarity, adapted to our setup (cf. Appendix~\ref{sec:all-hyperparams}).}
In all two-stage MTL methods, source and target tasks are the same, containing the tasks of the underlying benchmark. 
\rev{For each target task, they learn a transfer layer (except for \modeladaptersoup) and a new task head.}

We also select a set of strong STL baselines: \textbf{\modelfine}, fully fine-tuning the LM, \textbf{\modeladapter}~\cite{houlsby2019parameterefficient} learning an adapter module for each task, \textbf{\modelpropetl}~\citep{zeng-etal-2023-one} a more memory-efficient variation based on parameter sparsification and \textbf{\modelcompact}~\citep{compacter} a highly parameter-efficient variation using parameter-sharing between layers. 
\rev{In addition, we train $\mathbf{(IA)^3}$~\citep{liu-ia3}, learning scaling vectors applied to the key and value matrices and intermediate activations in the LM's feed-forward layer, and \textbf{\modellora}~\citep{hu2022lora}, learning low-rank updates to the model's weight matrices.}

Furthermore, we conduct experiments on several joint MTL baselines, namely \textbf{\modelfineMT}, \textbf{\modeladapterMT}, and \textbf{\modelpropetlMT}, the fully fine-tuned, adapter-based, and ProPETL-based joint MTL variants, respectively; and, finally, \textbf{\modelhyper} and \textbf{\modelhyperPP}~\citep{karimi-mahabadi-etal-2021-parameter}.
\modelfineMT updates all LM parameters, \modeladapterMT adds a single adapter module shared for all tasks, and \modelpropetlMT combines sparse layer- and task-specific masks through a logical OR operation. Based on task-specific embeddings, \modelhyper and \modelhyperPP generate module parameters by a shared hypernetwork. In all adapter-based models, we use a reduction factor of $16$, and, following \citet{pfeiffer-etal-2021-adapterfusion}, insert the modules after the feed-forward layer of the LM. Furthermore, to allow a fair comparison, we adapt \modelpropetlMT, \modelhyper, and \modelhyperPP to this setting by inserting the adapters only after each feed-forward block. 
To accommodate possible variations in performance, we train each model on multiple seeds, and report the mean and standard deviation over multiple runs.

The full details of the experiment setup regarding the benchmarks and their 
splits, infrastructure,
training, and hyperparameters are provided in \S~\ref{sec:all-hyperparams}. To further enable the reproducibility of our results, our code, including documentation, is available at 
\url{https://github.com/CPJKU/ScaLearn} under the MIT license.


\section{Results}
\label{sec:results}

\begin{table*}[t]
   \centering
   \scalebox{1}{
    \resizebox{1\linewidth}{!}{%
    \begin{tabular}{lllllllll|l}
    \toprule
    \textbf{Model} & \multicolumn{1}{c}{\textbf{MNLI}} & \multicolumn{1}{c}{\textbf{QQP}}  & \multicolumn{1}{c}{\textbf{QNLI}} & \multicolumn{1}{c}{\textbf{SST-2}} & \multicolumn{1}{c}{\textbf{STS-B}} & \multicolumn{1}{c}{\textbf{MRPC}} & \multicolumn{1}{c}{\textbf{RTE}}  & \multicolumn{1}{c|}{\textbf{CoLA}} & \multicolumn{1}{c}{\textbf{Avg.}} \\\midrule
    
    \modelfine & $86.61_{0.51}$ & $90.32_{0.15}$ & $91.78_{0.28}$ & $93.33_{0.48}$ & $90.53_{0.22}$ & $86.94_{1.52}$ & $73.47_{2.05}$ & $58.46_{4.03}$ & $83.93_{0.60}$ \\
    \modeladapter & $86.50_{0.33}$ & $90.18_{0.11}$ & $92.25_{0.19}$ & $93.65_{0.71}$ & $90.23_{0.41}$ & $86.64_{1.07}$ & $72.89_{2.54}$ & $58.28_{2.50}$ & $83.83_{0.48}$ \\
    \modelpropetl & $86.19_{0.25}$ & $88.88_{0.48}$ & $92.05_{0.80}$ & $93.81_{0.72}$ & $90.03_{0.35}$ & $85.93_{1.22}$ & $74.19_{2.03}$ & $59.29_{2.07}$ & $83.80_{0.42}$ \\
    \modelcompact & $85.62_{0.42}$ & $88.84_{0.70}$ & $91.79_{0.39}$ & $93.58_{0.34}$ & $89.67_{0.54}$ & $87.21_{0.61}$ & $72.02_{2.21}$ & $58.49_{2.58}$ & $83.40_{0.45}$ \\
    \rev{\modelia} & $\rev{83.78_{0.88}}$ & $\rev{88.37_{0.20}}$ & $\rev{90.57_{0.38}}$ & $\rev{93.35_{0.30}}$ & $\rev{89.93_{0.30}}$ & $\rev{87.11_{1.14}}$ & $\rev{72.56_{2.23}}$ & $\rev{56.57_{5.39}}$ & $\rev{82.78_{1.36}}$ \\
    \modellora & $86.52_{0.10}$ & $89.86_{0.33}$ & $92.25_{0.13}$ & $\underline{94.19}_{0.53}$ & $90.66_{0.31}$ & $87.03_{0.62}$ & $70.40_{8.33}$ & $57.55_{2.18}$ & $83.56_{1.56}$ \\    \midrule
    \modelfineMT & $84.95_{0.36}$ & $89.76_{0.12}$ & $90.91_{0.07}$ & $92.58_{0.76}$ & $86.14_{0.53}$ & $83.42_{0.50}$ & $80.99_{2.54}$ & $49.12_{1.74}$ & $82.23_{0.41}$ \\
    \modeladapterMT & $86.03_{0.18}$ & $89.69_{0.01}$ & $91.58_{0.30}$ & $93.35_{0.41}$ & $88.71_{0.49}$ & $86.76_{0.92}$ & $80.26_{1.96}$ & $51.79_{1.23}$ & $83.52_{0.32}$ \\
    \modelpropetlMT & $85.23_{0.45}$ & $87.82_{0.16}$ & $91.37_{0.52}$ & $93.88_{0.44}$ & $90.27_{0.22}$ & $86.36_{1.82}$ & $78.58_{0.90}$ & $54.71_{1.12}$ & $83.53_{0.31}$ \\
    \modelhyper & $86.08_{0.46}$ & $89.13_{0.23}$ & $91.81_{0.07}$ & $93.16_{0.99}$ & $90.63_{0.32}$ & $87.01_{0.87}$ & $82.79_{1.68}$ & $57.30_{2.21}$ & $84.74_{0.39}$ \\
    \modelhyperPP & $86.38_{0.18}$ & $88.81_{0.29}$ & $91.99_{0.17}$ & $93.27_{0.11}$ & $90.80_{0.12}$ & $87.83_{1.42}$ & \underline{$83.75$}$_{0.78}$ & $54.05_{3.30}$ & $84.61_{0.46}$ \\
    \midrule
    \modeladapterfusion & $86.82_{0.04}$ & $90.23_{0.01}$ & $92.48_{0.15}$ & $93.23_{0.95}$ & $90.37_{0.20}$ & \underline{$\bm{88.41}$}$_{0.49}$ & $79.49_{2.21}$ & $59.04_{1.69}$ & $85.01_{0.37}$ \\
    \rev{\modeladaptersoup} & $\rev{63.47_{0.37}}$ & $\rev{81.63_{0.23}}$ & $\rev{78.00_{0.20}}$ & $\rev{90.75_{0.24}}$ & $\rev{80.17_{0.18}}$ & $\rev{75.00_{1.18}}$ & $\rev{62.09_{0.64}}$ & $\rev{41.06_{1.68}}$ & $\rev{71.52_{0.59}}$ \\
    \modeladapteroursVecSplit & $86.97_{0.09}$ & $90.32_{0.10}$ & $92.51_{0.17}$ & $93.88_{0.18}$ & $\underline{\bm{90.96}}_{0.16}$ & $87.75_{0.58}$ & $\bm{82.06}_{1.37}$ & $58.47_{1.76}$ & $\underline{\bm{85.36}}_{0.55}$ \\
    \modeladapteroursScalarSplit & $86.93_{0.10}$ & $\underline{\bm{90.38}}_{0.11}$ & $\underline{\bm{92.53}}_{0.28}$ & $93.58_{0.20}$ & $90.08_{0.07}$ & $87.57_{0.86}$ & $80.07_{1.18}$ & $59.04_{1.05}$ & $85.02_{0.49}$ \\
    \modeladapteroursVecShare & $\underline{\bm{87.06}}_{0.03}$ & $90.04_{0.12}$ & $92.03_{1.10}$ & $\bm{94.15}_{0.30}$ & $90.62_{0.13}$ & $88.21_{0.63}$ & $80.87_{1.05}$ & $\underline{\bm{59.82}}_{0.78}$ & $85.35_{0.52}$ \\
    \modeladapteroursScalarShare & $86.98_{0.17}$ & $\underline{\bm{90.38}}_{0.01}$ & $\underline{\bm{92.53}}_{0.28}$ & $94.11_{0.07}$ & $90.18_{0.19}$ & $87.43_{0.63}$ & $80.04_{0.99}$ & $59.45_{0.67}$ & $85.14_{0.38}$ \\
    \bottomrule
    \end{tabular}
    } 
    }
    \caption{Evaluation results on GLUE using $\text{RoBERTa\textsubscript{BASE}}$. (Top) STL models, only learning a single task at a time. (Middle) Joint MTL methods, learning all tasks simultaneously. (Bottom) Two-stage MTL methods, composing the knowledge of several source adapters. The overall best results are \underline{underlined}, and the best results among the two-stage MTL models are \textbf{bold}.
   }
    \label{tab:results-glue}
\end{table*}

\begin{table*}[t]
    \centering
    \scalebox{1}{
    \resizebox{1\textwidth}{!}{%
    \begin{tabular}{lllllllll|l}
    \toprule
    \textbf{Model} & \multicolumn{1}{c}{\textbf{ReCoRD}} & \multicolumn{1}{c}{\textbf{MultiRC}} & \multicolumn{1}{c}{\textbf{BoolQ}} & \multicolumn{1}{c}{\textbf{WiC}}  & \multicolumn{1}{c}{\textbf{WSC}}  & \multicolumn{1}{c}{\textbf{COPA}} & \multicolumn{1}{c}{\textbf{CB}} & \multicolumn{1}{c|}{\textbf{RTE}} & \multicolumn{1}{c}{\textbf{Avg.}} \\\midrule
    \modelfine & $71.61_{0.84}$ & $71.64_{1.15}$ & $76.80_{1.34}$ & $66.38_{2.08}$ & \underline{$63.46$}$_{0.00}$ & $68.60_{6.74}$ & $81.96_{4.33}$ & $73.47_{2.05}$ & $71.74_{2.32}$ \\
    \modeladapter & $79.02_{0.62}$ & $72.84_{0.48}$ & $76.71_{1.38}$ & $65.58_{1.56}$ & \underline{$63.46$}$_{0.00}$ & $70.20_{4.13}$ & $84.82_{3.18}$ & $72.89_{2.54}$ & $73.19_{1.74}$ \\
    \modelpropetl & \underline{$80.29$}$_{0.24}$ & $73.07_{0.49}$ & $76.58_{0.78}$ & $66.60_{1.65}$ & \underline{$63.46$}$_{0.00}$ & $70.60_{3.44}$ & $84.46_{3.86}$ & $74.19_{2.03}$ & $73.69_{1.53}$ \\
    \modelcompact & $77.69_{2.67}$ & $70.44_{0.57}$ & $75.88_{0.96}$ & $66.46_{1.63}$ & \underline{$63.46$}$_{0.00}$ & $68.30_{4.00}$ & $87.68_{3.62}$ & $72.02_{2.21}$ & $72.74_{1.96}$ \\
    \rev{\modelia} & $\rev{75.27_{0.23}}$ & $\rev{70.32_{0.49}}$ & $\rev{76.31_{0.79}}$ & $\rev{67.07_{1.68}}$ & $\rev{63.35_{0.32}}$ & $\rev{69.30_{3.37}}$ & $\rev{87.32_{4.57}}$ & $\rev{72.56_{2.23}}$ & $\rev{72.69_{1.71}}$ \\
    \modellora & $79.60_{0.46}$ & $71.96_{0.36}$ & $76.58_{0.74}$ & $65.14_{1.17}$ & \underline{$63.46$}$_{0.00}$ & $68.20_{4.05}$ & $86.43_{3.17}$ & $70.40_{8.33}$ & $72.72_{2.28}$ \\    \midrule
    \modelfineMT & $72.21_{0.28}$ & $72.11_{0.68}$ & $76.39_{3.07}$ & $52.19_{1.11}$ & \underline{$63.46$}$_{0.00}$ & $74.33_{3.40}$ & $84.52_{0.84}$ & $74.85_{7.42}$ & $71.26_{2.10}$ \\
    \modeladapterMT & $72.43_{0.64}$ & $72.46_{0.43}$ & $75.32_{2.78}$ & $51.99_{1.74}$ & $59.94_{2.97}$ & $71.67_{3.40}$ & $86.31_{1.68}$ & $76.53_{1.06}$ & $70.83_{1.84}$ \\
    \modelpropetlMT & $73.14_{0.19}$ & $72.07_{0.58}$ & $73.91_{3.27}$ & $50.73_{0.99}$ & $59.62_{5.44}$ & $74.00_{3.27}$ & $82.14_{1.46}$ & $73.65_{3.83}$ & $69.91_{2.38}$ \\
    \modelhyper & $65.93_{4.47}$ & $33.54_{33.54}$ & $74.01_{1.10}$ & $55.49_{1.72}$ & $52.88_{10.58}$ & $55.50_{2.50}$ & $71.43_{7.14}$ & $61.73_{9.03}$ & $58.81_{8.76}$ \\
    \modelhyperPP & $24.50_{8.13}$ & $19.47_{27.53}$ & $62.17_{0.00}$ & $50.00_{0.00}$ & \underline{$63.46$}$_{0.00}$ & $54.33_{3.30}$ & $49.40_{0.84}$ & $49.09_{2.56}$ & $46.55_{5.30}$ \\
    \midrule
    \modeladapterfusion & $78.82_{0.49}$ & $71.79_{1.67}$ & $76.72_{0.55}$ & $66.57_{1.24}$ & \underline{$\bm{63.46}$}$_{0.00}$ & $73.10_{4.51}$ & $82.32_{2.85}$ & $76.03_{2.38}$ & $73.60_{1.71}$ \\
    \rev{\modeladaptersoup} & $\rev{64.26_{0.13}}$ & $\rev{33.62_{4.28}}$ & $\rev{68.84_{0.31}}$ & $\rev{58.53_{0.60}}$ & \rev{\underline{$\bm{63.46}$}$_{0.00}$} & $\rev{52.40_{2.41}}$ & $\rev{70.89_{0.86}}$ & $\rev{57.83_{0.93}}$ & $\rev{58.73_{1.19}}$ \\
    \modeladapteroursVecSplit & $79.52_{0.06}$ & \underline{$\bm{73.22}$}$_{0.44}$ & \underline{$\bm{77.27}$}$_{0.68}$ & $66.35_{1.20}$ & \underline{$\bm{63.46}$}$_{0.00}$ & $74.80_{2.15}$ & $90.89_{2.59}$ & $78.88_{2.14}$ & $75.55_{1.16}$ \\
    \modeladapteroursScalarSplit & $\bm{80.13}_{0.38}$ & $71.91_{0.60}$ & $76.06_{0.41}$ & $67.37_{1.22}$ & $62.50_{1.27}$ & $71.20_{1.23}$ & $89.11_{1.97}$ & $75.31_{0.90}$ & $74.20_{1.00}$ \\
    \modeladapteroursVecShare & $\bm{80.13}_{0.09}$ & $72.71_{0.57}$ & $76.44_{0.53}$ & $67.13_{1.24}$ & $62.26_{2.28}$ & \underline{$\bm{75.20}$}$_{1.93}$ & \underline{$\bm{93.04}$}$_{2.14}$ & \underline{$\bm{79.03}$}$_{0.95}$ & \underline{$\bm{75.74}$}$_{1.22}$ \\
    \modeladapteroursScalarShare & $79.79_{0.14}$ & $71.75_{0.38}$ & $76.13_{0.52}$ & \underline{$\bm{67.87}$}$_{0.89}$ & \underline{$\bm{63.46}$}$_{0.00}$ & $74.00_{1.70}$ & $91.61_{2.53}$ & $74.84_{1.58}$ & $74.93_{0.97}$ \\
    \bottomrule
    \end{tabular}
    }
    }
    \caption{Evaluation results on SuperGLUE using $\text{RoBERTa\textsubscript{BASE}}$.
    }
    \label{tab:results-superglue}
\end{table*}

\begin{table*}[t]
    \centering
    \scalebox{1}{
    \scriptsize
    \begin{tabular}{llllll|l}
    \toprule
    \textbf{Model} & \multicolumn{1}{c}{\textbf{Sectors}} & \multicolumn{1}{c}{\textbf{Pillars 1D}}  & \multicolumn{1}{c}{\textbf{Subpillars 1D}} & \multicolumn{1}{c}{\textbf{Pillars 2D}} & \multicolumn{1}{c|}{\textbf{Subpillars 2D}} & \multicolumn{1}{c}{\textbf{Avg.}} \\\midrule
    
    \modelfine & $71.99_{0.32}$ & $50.40_{0.24}$ & $43.76_{0.67}$ & $61.04_{0.26}$ & $41.68_{0.62}$ & $53.77_{0.42}$ \\
    \modeladapter & $71.38_{0.28}$ & $51.02_{1.23}$ & $43.26_{0.82}$ & $61.43_{0.91}$ & $42.46_{0.51}$ & $53.91_{0.75}$ \\
    \modelpropetl & $71.69_{0.86}$ & $49.69_{1.30}$ & $41.63_{0.84}$ & $60.58_{0.91}$ & $39.85_{1.10}$ & $52.69_{1.00}$ \\
    \modelcompact & $69.97_{1.89}$ & $37.37_{7.99}$ & $37.76_{2.14}$ & $58.13_{1.64}$ & $33.10_{9.00}$ & $47.26_{4.53}$ \\
    \rev{\modelia} & $\rev{70.22_{0.97}}$ & $\rev{45.55_{3.43}}$ & $\rev{40.05_{3.15}}$ & $\rev{58.54_{1.38}}$ & $\rev{39.27_{1.01}}$ & $\rev{50.73_{1.99}}$ \\
    \modellora & $71.08_{0.44}$ & $33.96_{29.41}$ & $42.75_{0.31}$ & $60.33_{0.52}$ & $\underline{42.81}_{0.63}$ & $50.19_{6.23}$ \\    \midrule
    \modelfineMT & $51.75_{3.62}$ & $22.65_{12.88}$ & $13.54_{6.06}$ & $33.27_{21.23}$ & $12.42_{3.39}$ & $26.73_{9.44}$ \\
    \modeladapterMT & $56.20_{2.72}$ & $28.53_{14.56}$ & $16.53_{9.46}$ & $35.90_{17.36}$ & $18.89_{2.64}$ & $31.21_{9.35}$ \\
    \modelpropetlMT & $59.80_{10.09}$ & $26.10_{14.36}$ & $29.57_{7.40}$ & $37.53_{12.08}$ & $30.35_{5.91}$ & $36.67_{9.97}$ \\
    \modelhyper & $71.08_{1.04}$ & $40.65_{6.93}$ & $34.16_{3.37}$ & $46.22_{14.11}$ & $32.47_{4.46}$ & $44.92_{5.98}$ \\
    \modelhyperPP & $60.42_{9.79}$ & $22.07_{7.45}$ & $20.35_{7.04}$ & $30.55_{19.83}$ & $18.90_{10.84}$ & $30.46_{10.99}$ \\
    \midrule
    \modeladapterfusion & $72.05_{0.12}$ & $49.63_{0.53}$ & $43.15_{0.38}$ & $60.68_{0.23}$ & $42.14_{0.46}$ & $53.53_{0.35}$ \\
    \rev{\modeladaptersoup}  & $\rev{56.81_{1.90}}$ & $\rev{30.09_{0.40}}$ & $\rev{21.84_{0.55}}$ & $\rev{40.71_{0.98}}$ & $\rev{17.89_{2.02}}$ & $\rev{33.47_{1.17}}$ \\
    \modeladapteroursVecSplit & $72.36_{0.05}$ & $51.63_{0.61}$ & $44.06_{0.37}$ & $61.52_{0.11}$ & $\bm{42.81}_{0.63}$ & $\underline{\bm{54.48}}_{0.35}$ \\
    \modeladapteroursScalarSplit & $72.20_{0.14}$ & $50.08_{0.79}$ & $42.97_{0.70}$ & $60.62_{0.16}$ & $41.95_{0.60}$ & $53.56_{0.48}$ \\
    \modeladapteroursVecShare & $\underline{\bm{72.38}}_{0.27}$ & $\underline{\bm{51.66}}_{0.27}$ & $\underline{\bm{44.23}}_{0.50}$ & $\underline{\bm{61.66}}_{0.13}$ & $42.21_{0.21}$ & $54.43_{0.28}$ \\
    \modeladapteroursScalarShare & $72.02_{0.32}$ & $50.78_{0.41}$ & $42.60_{0.85}$ & $60.82_{0.14}$ & $42.14_{0.72}$ & $53.67_{0.49}$ \\
    \bottomrule
    \end{tabular}
    }
    \caption{Evaluation results on HumSet using $\text{XLM-R\textsubscript{BASE}}$.}
    \label{tab:results-humset}
\end{table*}

\subsection{Parameter-efficiency analysis}
Table~\ref{tab:parameter-efficiency} provides a comprehensive overview of the number of learnable parameters of the models in our experiment setting on GLUE and SuperGLUE: $\text{RoBERTa\textsubscript{BASE}}$ as the backbone LM, $8$ source tasks, and the same $8$ tasks as target tasks ($|S|\!=\!|T|\!=\!8$). Starting from the STL models, the left and right columns report the number of trainable parameters for one and all tasks, respectively. The joint MTL models learn all tasks simultaneously, and hence only contain values in the right column. For the two-stage MTL models, we report the number of trainable parameters of the transfer layer for one target task ($\Omega_t$) in the first column and the same for all target tasks on the right ($\Omega$).
We deliberately organize the transfer parameters of the two-stage models ($\Omega$) under the corresponding numbers of other models in the right column since the two-stage paradigm benefits from already trained adapters and only needs to learn the transfer layer.
If the adapters should also be trained, we provide an extra comparison with the corresponding additional parameters in Appendix~\ref{sec:all-hyperparams}.

When comparing the results of the two-stage MTL methods in the transfer layer, \modeladapterfusion is expectedly far less parameter-efficient than \modelours models, where \modeladapteroursScalarShare only requires 64 parameters. The variants of \modelours add considerably fewer transfer parameters compared to the overall parameters of the particularly efficient joint MTL methods. Moreover, the \modelours models still remain comparable when also taking into account the source adapter parameters. Considering these results, in the following we report and discuss the evaluation results in transfer learning and few-shot learning on the respective benchmarks. 


\subsection{Transfer Learning Performance}

\vspace{0.5em}  
\noindent\textbf{Results on GLUE.} Table~\ref{tab:results-glue} shows the evaluation results on the GLUE benchmark using $\text{RoBERTa\textsubscript{BASE}}$. The evaluation metrics are Pearson's correlation for STS-B, Matthews' correlation for CoLA, and accuracy for the rest. We average the results over several runs and report the corresponding standard deviation in the subscripts. Overall, the two-stage models obtain strong gains, outperforming STL and joint MTL models. Remarkably, all variants of \modelours, including the highly parameter-efficient \modeladapteroursScalarShare achieve similarly good results with only a fraction of the parameters of \modeladapterfusion. 
Comparing the different variations of our method, while \modelours shows the best results, the other models also perform highly competitively. 

\vspace{0.5em}  
\noindent\textbf{Results on SuperGLUE.} Table~\ref{tab:results-superglue} shows the results on SuperGLUE for all methods considered. The evaluation metrics are F1 for MultiRC and ReCoRD and accuracy for the other tasks.
We observe similar patterns on this benchmark: two-stage models generally outperform other baselines. In this benchmark, \modelours and \modeladapteroursVecShare improve upon \modeladapterfusion by 2 percentage points of the average results. 
Notably, we observe performance drops for various joint MTL models in comparison to other models (up to $-27$\% when comparing \modelhyperPP and \modeladapter). This may be a signal of the sensitivity of these models to the selection of tasks.
\rev{Furthermore, the subpar performance of AdapterSoup suggests that calculating weights using sentence similarity is not appropriate for our specific problem setup.}
In contrast, the other two-stage MTL models (and, in particular, our \modelours models) do not show any considerable performance decreases. 

\vspace{0.5em}  
\noindent\textbf{Results on HumSet.} Table~\ref{tab:results-humset} shows the results on HumSet using $\text{XLM-R\textsubscript{BASE}}$ with the F1-score as the evaluation metric. Similarly, \modelours performs the best among all the methods, whereas the more parameter-efficient variants of \modelours are only marginally weaker in performance. On this benchmark, in particular, all joint MTL methods show poor performance, highlighting the sensitivity of these methods to task selection (up to $-27$\% for STL and MTL versions of \modelfine).

We conduct an ablation study on the effect on different combinatorial operators in \modelours, reported in Appendix~\ref{sec:additional-ablation}. In Appendix~\ref{sec:additional-results}, we provide further experiments and analyses of the results along with the results of GLUE and SuperGLUE using $\text{RoBERTa\textsubscript{LARGE}}$, HumSet using $\text{XLM-R\textsubscript{LARGE}}$, and for the combination of all tasks from GLUE and SuperGLUE. Finally, we provide an analysis of the scaling coefficients of \modeladapteroursScalarSplit and \modeladapteroursScalarShare in Appendix~\ref{sec:activation-vis}, revealing the effect of various source adapters on a target task. 

\subsection{Few-shot Transfer Learning}\label{sec:few-shot}

\begin{figure*}[h]
    \center
    \begin{subfigure}{0.32\textwidth}
        \centering
        \includegraphics[width=1\textwidth]{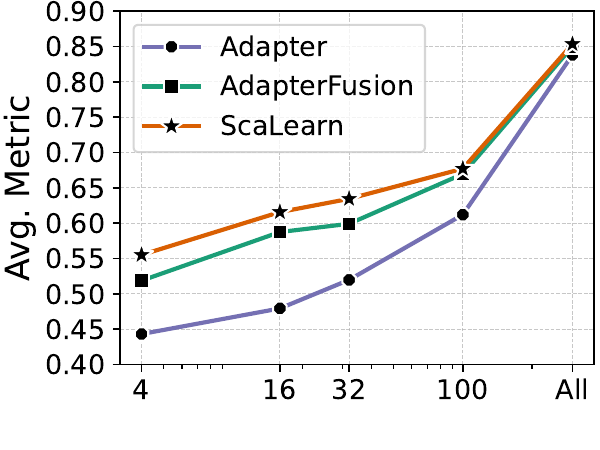}
        \caption{GLUE}
        \label{fig:fig-few.a}
    \end{subfigure}
    \begin{subfigure}{0.32\textwidth}
        \centering
        \includegraphics[width=1\textwidth]{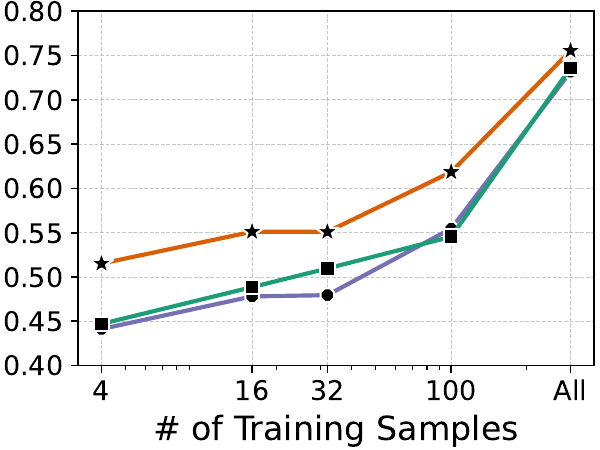}
        \caption{SuperGLUE}
        \label{fig:fig-few.b}
    \end{subfigure}
     \begin{subfigure}{0.32\textwidth}
        \centering
        \includegraphics[width=1\textwidth]{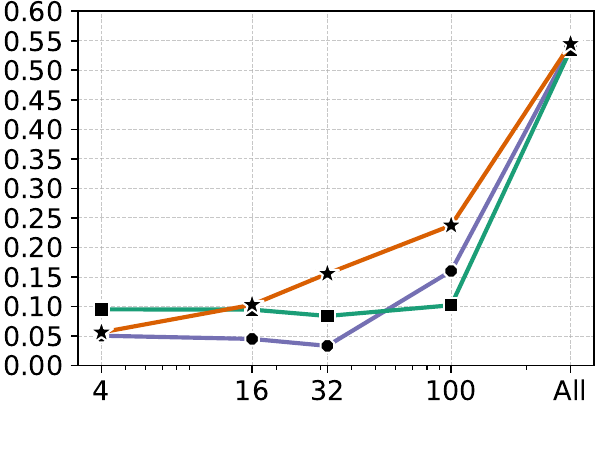}
        \caption{HumSet}
        \label{fig:fig-few.c}
    \end{subfigure}
    \caption{Few-shot transfer learning results with $k=\text{\{4,16,32,100\}}$ training samples for each target task using the $\text{BASE}$ models of $\text{RoBERTa}$ and $\text{XLM-R}$. 
     Full results over several runs are provided in Appendix~\ref{sec:more-fs-results}.}
    \label{fig:few-shot}
\end{figure*}

We further assess the applicability of \modelours in a few-shot setting, where we assume that only $k=\text{\{4,16,32,100\}}$ training samples are available for a given target task. 
For two-stage MTL methods, for a given benchmark, we use the source adapters of all tasks except the one corresponding to the target task, where we use a source adapter trained on only $k$ samples.
On the basis of this set of source adapters, we then train a transfer layer on the target task using $k$ data points.

Figure~\ref{fig:few-shot} shows the performance of \modeladapter, \modeladapterfusion, and \modelours on the GLUE, SuperGLUE, and HumSet benchmarks, averaged over 5 runs. We observe that \modelours consistently outperforms \modeladapter and \modeladapterfusion in all benchmarks and values of $k$ (except for $k=4$ on HumSet) pointing to the strength of our method for data-lean settings.
We provide the full results, including per-dataset ones, other variations of \modelours, and on $\text{RoBERTa\textsubscript{LARGE}}$ in \S\ref{sec:more-fs-results}.

\section{Related Work}
\label{sec:related}
\vspace{0.5em}  
\noindent\textbf{Parameter-efficient task learning in NLP.}
Various parameter-efficient methods have emerged as a more sustainable alternative to full fine-tuning, enabling modularization, efficient sharing, and reusability of knowledge. 
A common modularization approach is to introduce a small number of additional parameters into an LM, realized by various methods such as Adapters~\citep{RebuffiBV17, houlsby2019parameterefficient}, 
Compacter~\citep{compacter}, and 
ProPETL-Adapter~\citep{zeng-etal-2023-one}.
Similarly, LoRA~\citep{hu2022lora} injects trainable low-rank matrices into each transformer layer, and BitFit~\citep{ben-zaken-etal-2022-bitfit} updates only the bias terms.
Another line of research identifies sparse subnetworks within the model to tune~\citep{ansell-etal-2022-composable, guo-etal-2021-parameter,hauzenberger2023parameter, DBLP:journals/corr/abs-2401-16405}, 
while \citet{unified-adapter} and \citet{mao-unipelt} propose to merge various distinct modules. We refer to \citet{modularDeepL} for a full survey on this topic.

\vspace{0.5em}
\noindent\textbf{Learning by scaling.}
Besides the common approach of learning a feed-forward layer for a (non--)\xspace linear transformation of an input vector, several recent methods explore the merit of learning a scaling vector applied to the input vector in various scenarios. \citet{liu-ia3} learn a modular network for STL that rescales LM vectors through element-wise multiplication. 
\citet{gabriel:task_arithmetic} and \citet{guillermo:NTK_task_arithmetic} introduce task arithmetic to control LM behavior by extracting task vectors from pre- and post-fine-tuning model weights, then scaling and combining them to improve MTL performance.
\citet{congater:eacl} learn a gating adapter that adjusts the scaling of representations to control the behavior of the model at inference time. Finally, \citet{lian-ssf} learn to shift and scale the output vectors of a vision transformer in an STL setting. Our work contributes to this line of research by leveraging scaling for highly parameter-efficient and effective MTL.

\vspace{0.5em}  
\noindent\textbf{Joint MTL.}
Interference and imbalance between tasks have been shown to impede performance in joint MTL~\citep{kirkpatrick-catastrophic, kendall2018multi, modularDeepL}.
Several studies have aimed to address these issues and improve generalization. For example, \citep{liu-mtdnn} learn representations across multiple NLU tasks using context from a semantic similarity model, and \citet{DBLP:conf/iclr/PilaultEP21} introduce a parameter-efficient model that uses modules facilitating weight sharing.
Moreover, \citet{DBLP:conf/icml/Stickland019} use an adapter for each task while also updating the LM parameters.
\citet{Zhang2022SkillNetNLUAS} further focus on modularity by only activating a subset of task-specific modules at once; however, tasks must be mapped a priori to a given high-level skill.
\rev{\citet{ponti-etal-2023-combining} and \citet{caccia2022multi} loosen this constraint by learning a task-skill allocation matrix for cross-task generalization, but rely on a multi-task pre-training stage.} Finally, 
\citet{hyperformer}
leverage a hypernetwork~\citep{ha:hypernetworks} that generates modular task-specific parameters.

\vspace{0.5em}  
\noindent\textbf{Two-stage MTL.}
Various methods have been proposed to extract task-specific information and compose this knowledge. \citet{adapterSoup} studies transfer learning in generative LMs by first selecting source adapters based on different heuristics and merging their weights to create a new combined adapter. \citet{holtermann:wtw} provide further insights into how to combine adapters effectively and efficiently for zero-shot knowledge compositions.
Furthermore, \citet{lorahub} introduce LoraHub with the aim of composing LoRA~\citep{hu2022lora} modules for cross-task generalization using black-box optimization and an additional pre-filtering stage. 
\citet{asai-attempt} and \citet{wang-MPT} leverage continuous prompts learned on large-scale source tasks, leading to competitive performance in MTL benchmarks, although both methods depend on the selection of typically high-resource source tasks. In contrast to the mentioned methods that highly depend on the selection of tasks and/or apply the combination to the weights, 
\citet{pfeiffer-etal-2021-adapterfusion} combines the output representations of several independent source adapters through an attention mechanism. Our work is directly related to this line of research and introduces a novel highly parameter-efficient transfer layer applied to the output representation. 
\section{Conclusion}  
We propose \modelours, a highly parameter-efficient and effective two-stage MTL method leveraging simple scaling of output vectors. Our proposed approach directly learns the coefficients that scale the representations of source adapters and combines them simply by taking the sum. 
We conduct transfer learning experiments \rev{using encoder LMs} on the three benchmarks of GLUE, SuperGLUE, and HumSet, consisting of a diverse set of tasks, domains, and languages. Our results show that \modelours and even its extremely parameter-efficient variants obtain strong improvement over existing
MTL methods without any negative cross-task effects. We further show that these improvements are also present in few-shot transfer learning. 

\section*{Limitations}

The first limitation of our work concerns the selection of benchmarks -- we conducted experiments only on the GLUE, SuperGLUE, and HumSet benchmarks. While these already cover a vast number of tasks and domains of varying sizes in different languages, they still do not fully represent the myriad of tasks, domains, and languages within the NLP domain. However, we strongly believe that our findings also hold for other transfer learning corpora, including different tasks, domains, and languages, especially since \modelours\xspace* models are agnostic concerning this selection.
Related to this aspect, we focused on transformer-based encoder LMs as the backbone for our experiments and did not experiment with other architectures, \eg convolutional or recurrent networks, or transformer-based decoder LMs. 
Finally, we relied on adapters as arguably the most popular modularization technique (cf. \citet{pfeiffer-etal-2021-adapterfusion, adapterSoup}). Due to the large number of additional experiments required and related environmental concerns, we did not experiment with other modularization methods (e.g., LoRA or \modelia). However, our method clearly shows the usefulness of simply scaling output representations of modules for transfer learning.

\section*{Ethical Considerations}
The nature of our work is manifold, and so are the ethical aspects touched by our research. First, we acknowledge the potential of NLP datasets and models for encoding unfair stereotypical~\citep{blodgett2020} and exclusive~\citep{dev-etal-2021-harms} biases that may lead to representational and allocational harms~\citep{barocas2017problem}. This potential is a general property of pre-trained language models, and the models and datasets we use in this research are no exception to this danger. We thus strongly advise practitioners to carefully consider the sociotechnical context before deploying any models (with or without \modelours), and, aligned with the specific deployment scenario, to take measures against unfair discrimination. Examples of such measures include the use of bias measurement~\citep{nangia-etal-2020-crows} and mitigation~\citep{bordia-bowman-2019-identifying} approaches. 
Second, the core of this work deals with efficiency aspects. On the one hand, given the well-known relationship between model training (and inference) effort and potential CO$_2$ emissions~\citep{strubell-etal-2019-energy}, our work directly contributes to reaching the goals of Green AI by making parameter-efficient MTL more environmentally sustainable. On the other hand, since the training of language models often comes with high infrastructure requirements exclusive to certain user groups~\citep{StochasticParrots}, we hope that our work also contributes to the ongoing democratization of language technology by reducing resource-related usage barriers.


\section*{Acknowledgements}
This work received financial support by the State of Upper Austria and the Federal Ministry of Education, Science, and Research, through grant LIT-2021-YOU-215. This work was also funded by the Austrian Science Fund (FWF): P36413, P33526, and DFH-23. The work of Carolin Holtermann and Anne Lauscher is funded under the Excellence Strategy of the German Federal Government and the States. The authors would like to thank Benjamin Minixhofer for his invaluable feedback on the manuscript.
\bibliography{cites}

\clearpage
\appendix
\section{Appendix}
\label{sec:appendix}
\subsection{Complete Experiment Details}\label{sec:all-hyperparams}

\begin{table}[h]
    \centering
    \small
    \begin{tabular}{l|ccc}
    \toprule
    Name & $\vert$Train$\vert$ & $\vert$Validation$\vert$ & $\vert$Test$\vert$ \\
    \midrule
    MNLI & 353,431 & 39,270 & 9,815 \\
    QQP & 327,461 & 36,384 & 40,430 \\
    QNLI & 94,268 & 10,474 & 5,463 \\
    SST-2 & 60,614 & 6,734 & 872 \\
    STS-B & 5,174 & 574 & 1,500 \\
    MRPC & 3,301 & 366 & 408 \\
    RTE & 2,241 & 249 & 277 \\
    CoLA & 7,695 & 855 & 1,043 \\
    \midrule
    ReCoRD & 100,730 & 10,000 & 10,000 \\
    MultiRC & 24,518 & 2,724 & 4,848 \\
    BoolQ & 8,484 & 942 & 3,270 \\
    WiC & 4,885 & 542 & 638 \\
    WSC & 498 & 55 & 104 \\
    COPA & 360 & 40 & 100 \\
    CB & 225 & 25 & 56 \\
    \midrule
    Sectors & 117,435 & 16,039 & 15,147 \\
    Pillars 1D & 117,435 & 16,039 & 15,147 \\
    Subpillars 1D & 117,435 & 16,039 & 15,147 \\
    Pillars 2D & 117,435 & 16,039 & 15,147 \\
    Subpillars 2D & 117,435 & 16,039 & 15,147 \\
    \bottomrule
    \end{tabular}
    \caption{Number of used samples for each dataset and used split. (Top) GLUE tasks. (Middle) SuperGLUE tasks. (Bottom) HumSet tasks.}
    \label{tab:dataset-n-samples}
\end{table}

\begin{table*}[h]
    \centering
    \resizebox{1\textwidth}{!}{%
    \begin{tabular}{l|llll}
    \toprule
    Name & Category & Task & Domain & Metric \\\midrule
    MNLI & GLUE & NLI & various & accuracy   \\
    QQP & GLUE & paraphrase detection & social QA & accuracy \& F1 \\
    QNLI & GLUE & NLI & Wikipedia & accuracy \\
    SST-2 & GLUE & sentiment analysis & Movie Reviews & accuracy  \\
    STS-B &  GLUE & sentence similarity &  various & \underline{Pearson} \& Spearman corr. \\
    MRPC & GLUE & paraphrase detection  & news & \underline{accuracy} \& {F1}  \\
    RTE & GLUE & NLI & News, Wikipedia & accuracy \\
    CoLA & GLUE & acceptability & various & Matthews' corr.  \\
    \midrule
    ReCoRD & SuperGLUE & cloze-style QA & news (CNN, Daily Mail) & F1 \& EM \\
    MultiRC & SuperGLUE & QA & various & \underline{F1} \& EM \\
    BoolQ & SuperGLUE & boolean QA &  Wikipedia & accuracy   \\
    WiC & SuperGLUE & word sense disambiguation  & lexical databases & accuracy \\
    WSC & SuperGLUE & coreference / commonsense & fiction books & accuracy\\
    COPA & SuperGLUE & commonsense reasoning & various & accuracy \\
    CB & SuperGLUE & NLI & various & accuracy \\
    \midrule
    Sectors & HumSet & classification & humanitarian crisis response & \underline{F1} \& precision \\
    Pillars 1D & HumSet & classification & humanitarian crisis response & \underline{F1} \& precision\\
    Subpillars 1D & HumSet & classification & humanitarian crisis response & \underline{F1} \& precision\\
    Pillars 2D & HumSet & classification & humanitarian crisis response & \underline{F1} \& precision\\
    Subpillars 2D & HumSet & classification & humanitarian crisis response & \underline{F1} \& precision\\
     \bottomrule
     \end{tabular}
     }
    \caption{Details of all datasets. Lexical databases for WiC include WordNet, VerbNet, Wiktionary. For datasets where two metrics are officially used, we use the \underline{underlined} metric as our main metric. (Top) GLUE tasks. (Middle) SuperGLUE tasks. (Bottom) HumSet tasks.}
    \label{tab:task-overview}
\end{table*}

\vspace{0.5em}  
\noindent\textbf{Dataset Details.} As has been mentioned, we are using the GLUE, SuperGLUE, and HumSet benchmarks for our experiments. Table~\ref{tab:task-overview} summarizes the tasks contained in each of the datasets. We use the \texttt{datasets} library~\citep{lhoest-etal-2021-datasets} to load each dataset for our experiments. We set the maximum length of the input sequence to 128 tokens for all tasks in GLUE, SuperGLUE, and HumSet. However, for MultiRC and ReCoRD, we set the maximum length to 324 and 256, respectively, due to their significantly longer context lengths. Note that we treat HumSet as five separate tasks, following~\citep{fekih-etal-2022-humset}.
The GLUE and SuperGLUE benchmarks only contain the training and validation split publicly, so we follow \citet{chen-etal-2022-revisiting} and use 10\% of the training samples from the training split as the validation set and the remaining 90\% for training. We split the datasets with the \texttt{datasets} library~\citep{lhoest-etal-2021-datasets} using seed \texttt{42} and shuffle the samples. Then, the original validation split is taken as the test set on which we report the performance of all models. For HumSet, we use the original train/validation/test splits, as all of them are publicly available, including labels. Details about the train/validation/test splits can be found in Table~\ref{tab:dataset-n-samples}.

\vspace{0.5em}  
\noindent\textbf{Computing Infrastructure.} We run all experiments with $\text{RoBERTa\textsubscript{BASE}}$ and $\text{XLM-R\textsubscript{BASE}}$ on a single Nvidia GTX1080Ti GPU and Intel Xeon CPU E5-2640 v4 CPUs, and the experiments with $\text{RoBERTa\textsubscript{LARGE}}$ and $\text{XLM-R\textsubscript{LARGE}}$ on a single Nvidia RTX5000 GPU and Intel Xeon Silver 4216 CPUs.

\vspace{0.5em}  
\noindent\textbf{Implementation Details.} We use \texttt{PyTorch}~\citep{pytorch} for all experiments. For the joint multi-task learning methods, we adapt the codebase of \citet{karimi-mahabadi-etal-2021-parameter} and \citet{zeng-etal-2023-one}, both of which rely on the \texttt{transformers}~\citep{wolf-etal-2020-transformers} library. For all other models, we make use of the \texttt{adapter-transformers} library~\citep{adapterhub} library, a wrapper around the \texttt{transformers} library. Our code is released under the MIT License, ensuring open access to the community for further development.

\vspace{0.5em}  
\noindent\textbf{Training and optimization.} We train all methods with a batch size of $32$. All STL and two-stage MTL methods are trained for a maximum of $30$ epochs with early stopping and patience of 5. \footnote{The exception is ReCoRD, which we train on 3 epochs due to its size.} We use $10$ seeds for low-resource and $3$ seeds for high-resource tasks when using $\text{RoBERTa\textsubscript{BASE}}$, and on $5$ and $2$ seeds for low- and high-resource tasks, respectively, when using $\text{RoBERTa\textsubscript{LARGE}}$. We define tasks with more than $10k$ training samples as high-resource and as low-resource otherwise. All joint MTL models are trained on $3$ seeds. We report the mean and standard deviations across all runs. We use the AdamW~\citep{AdamW, DBLP:conf/iclr/LoshchilovH19} optimizer with default PyTorch hyperparameters ($\text{weight decay}=0.01$, $\beta_1=0.9$, $\beta_2=0.99$, $\epsilon=1\cdot 10^{-6}$). We use seeds \texttt{\{0,1\}} for instances with two seeds, \texttt{\{0,1,2\}} for instances with three seeds, seeds \texttt{\{0,1,2,3,4\}} for instances with five seeds, and \texttt{\{0,1,2,3,4,5,6,7,8,9\}} for instances with ten seeds.

\begin{table*}[t]
\centering
\scriptsize
\begin{tabular}{l r@{\hskip3pt}l r @{\hskip3pt}l r @{\hskip3pt}l}
    \toprule
    \textbf{Model} & \multicolumn{2}{c}{\textbf{\makecell{Parameters\\(one task)}}} & \multicolumn{2}{c}{\textbf{\makecell{Parameters\\(all tasks)}}} & \multicolumn{2}{c}{\textbf{\makecell{Task ($\Theta$) + Transfer ($\Omega$)\\(source adapters + transfer layers)}}} \\
    \midrule
    \modeladapterfusion & 17.05\% & (21M) & 136.40\% & (170M) & $5.74\%+136.40\%=\!142.14$\% &  (177M) \\
    \modeladapteroursVecSplit & 0.06\% & (74K) & 0.47\% & (590K) & $5.74\%+0.47\%=\!6.21$\% &  (8M) \\
    \modeladapteroursScalarSplit & 0.00\% & (96) & 0.00\% & (768) & $5.74\%+0.00\%=\!5.74$\% &  (7M) \\
    \modeladapteroursVecShare & 0.00\% & (6K) & 0.04\% & (49K) & $5.74\%+0.04\%=\!5.79$\% &  (7M) \\
    \modeladapteroursScalarShare & 0.00\% & (8) & 0.00\% & (64) & $5.74\%+0.00\%=\!5.74$\% &  (7M) \\
    \bottomrule
\end{tabular}
\caption{Percentage and number of trainable parameters for Two-Stage MTL models in total.}
\label{tab:two-stage-mtl}
\end{table*}

\vspace{0.5em}  
\noindent\textbf{Single-task learning hyperparameters.} We train \modelfine with a learning rate of 2e-5, \modeladapter with a learning rate of 3e-4, \modelcompact with a learning rate of 3e-3, and \modelpropetl with a learning rate of 1e-3, a mask learning rate of 5e-3, a sparsity rate of $0.5$, and a weight decay of $0.1$, which we found to be the most suitable for our setup. Furthermore, we train \modelia with a learning rate of 5e-3. For \modellora, we use a learning rate of 3e-4 in combination with rank $r=32$ and scaling factor $\alpha=64$. Moreover, we follow \citet{hu2022lora} and apply LoRA on the query and value matrices of the transformer. Each of them is trained with a linear learning rate decay. 

For $\text{RoBERTa\textsubscript{LARGE}}$, we add a linear learning rate warmup for the first 10\% of training, as we notice it improves stability. For early stopping, we use the loss on the validation set, except for HumSet, where we use the F1-score, and in the few-shot setting, where we use the main metric for the respective dataset, as shown in Table~\ref{tab:task-overview}. In the few-shot setting, we train for a maximum of 1,000 steps, apply an early stopping patience of $20$, and use a maximum of 5,000 samples for validation. Note that, while the layer normalization parameters of the LM have also been updated~\citep{compacter, hyperformer}, following \citet{pfeiffer-etal-2021-adapterfusion}, we keep them frozen. This approach improves modularity, while still allowing LMs to efficiently adapt to new tasks. Note that the same hyperparameters as outlined here are also used for \modeladapter in our probing analyses (cf. Appendix~\ref{sec:analyses}).

\vspace{0.5em}  
\noindent\textbf{Joint MTL hyperparameters.} In all joint multi-task learning methods, we sample tasks with conventional temperature-based sampling with temperature $\tau=10$, following \citet{hyperformer} and \citet{zeng-etal-2023-one}. Specifically, a task is sampled with probability $p_t^{1/\tau}$, where $p_t=\frac{N_t}{\sum_{i=1}^\tau N_t}$, $N_t$ the number of training samples of task $t$, and $\tau=10$. Using this sampling strategy, we train each model for a total of 375,000 steps to ensure convergence and evaluate every 7,500 steps. We train each model with early stopping and patience of $10$. In the end, the model checkpoint with the lowest average validation loss is loaded and evaluated on the test set. We train \modelfineMT with a learning rate of 2e-5, \modeladapterMT, \modelhyper, and \modelhyperPP with a learning rate of 3e-4, and \modelpropetlMT with a learning rate of 3e-4 and a mask learning rate of 3e-3, a sparsity rate of $0.3$, and no weight decay. We train each of them with a linear learning rate warmup for the first 10\% of training, followed by a linear learning rate decay. For the remaining hyperparameters of \modelpropetlMT, \modelhyper, and \modelhyperPP, we follow the respective original implementations, but always use a reduction factor of $16$ for a fair comparison. 

\vspace{0.5em}  
\noindent\textbf{Two-stage MTL hyperparameters}. We train each variant of \modelours$\!$* with a learning rate of 6e-3 and train \modeladapterfusion with a learning rate of 5e-5, following \citet{pfeiffer-etal-2021-adapterfusion}. Both \modelours$\!$* and \modeladapterfusion are trained with a linear learning rate decay and no warmup. Early stopping is the same as in the single-task learning setting. We initialize the parameters of \modelours$\!$* with $\mathcal{N}\left(\frac{2}{T}, 0.001\right)$,\footnote{We also test out \{$\mathcal{N}\left(\frac{1}{T}, 0.001\right)$, $\mathcal{N}\left(\frac{3}{T}, 0.001\right)$, $\mathcal{N}\left(1, 0.001\right)$\}.} and apply a dropout rate of $0.3$ to increase robustness for \modelours and \modeladapteroursVecShare.
\rev{For AdapterSoup, we first calculate the cosine similarity of sentence embeddings for each task from the training set using the \texttt{sentence-transformers}~\citep{reimers-gurevych-2019-sentence} library and the \texttt{all-mpnet-base-v2} model. In contrast to \citet{adapterSoup}, who only select $100$ samples for each domain, we select $10000$ samples for each task, as our sequences corresponding to tasks are meaningfully shorter than the sequences corresponding to domains. Using these similarities, we select the top $5$ most similar tasks to the target task, normalize the similarity scores to obtain the weights, and perform weight-space averaging of the adapter parameters, following \citet{adapterSoup}. Note that we also include the corpus of the target task when calculating the similarities for weight-space averaging, and hence also the target adapter during weight-space averaging, and train a new task head on the target task to allow a more fair comparison to other two-stage MTL methods.
We use a learning rate of 3e-4 when training the target task head with \modeladaptersoup.}

\vspace{0.5em}  
\noindent\textbf{Efficiency of two-stage MTL methods.} We provide a comprehensive comparison of all trainable parameters of two-stage MTL methods if all the adapters should also be trained in Table~\ref{tab:two-stage-mtl}.

\subsection{Analysis on Scaling Output Representations}\label{sec:analyses}

\begin{figure*}[t]
    \centering
    \begin{subfigure}{0.24\textwidth}
        \centering
        \includegraphics[width=\textwidth]{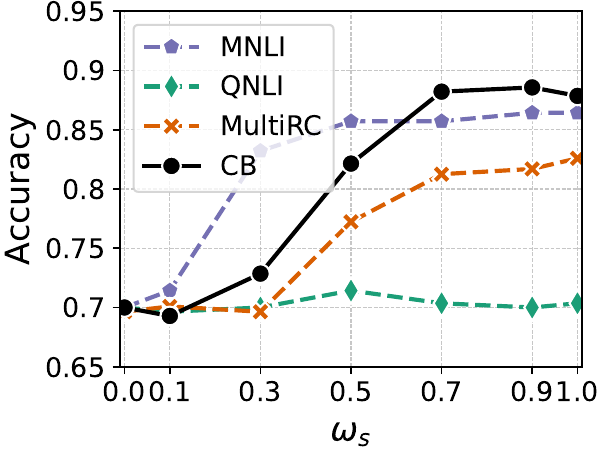}
        \includegraphics[width=\textwidth]{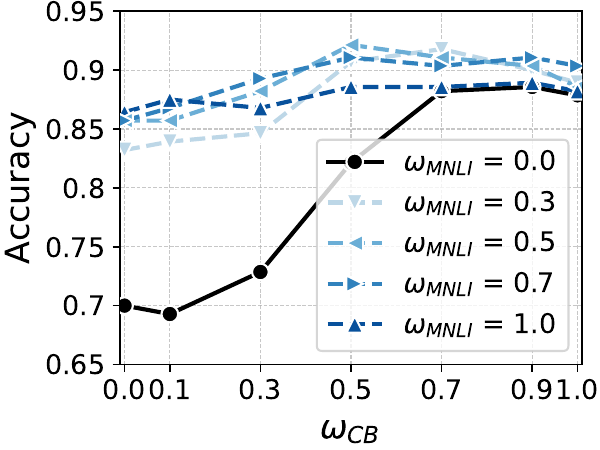}
        \caption{CB}
        \label{fig:fig1.a}
    \end{subfigure}
    \begin{subfigure}{0.24\textwidth}
        \centering
        \includegraphics[width=\textwidth]{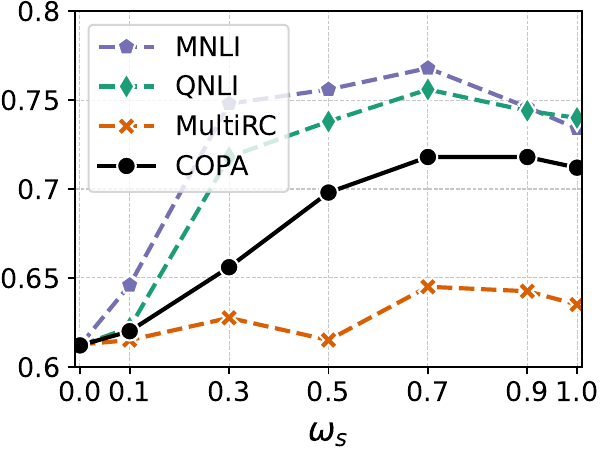}
        \includegraphics[width=\textwidth]{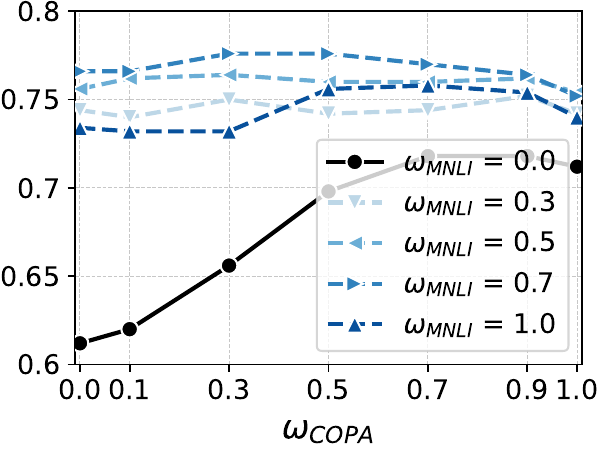}
        \caption{COPA}
        \label{fig:fig1.b}
    \end{subfigure}
    \begin{subfigure}{0.24\textwidth}
        \centering
        \includegraphics[width=\textwidth]{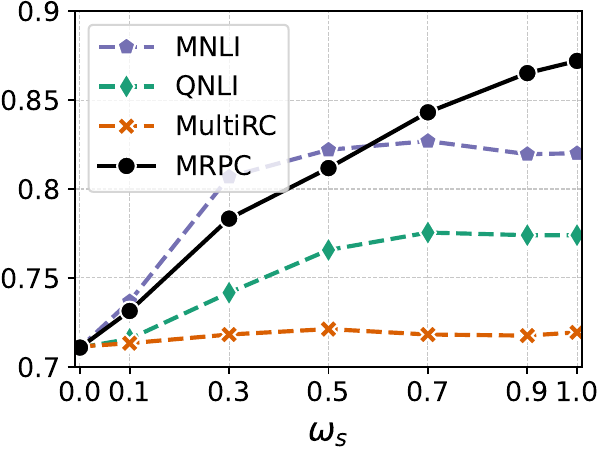}
        \includegraphics[width=\textwidth]{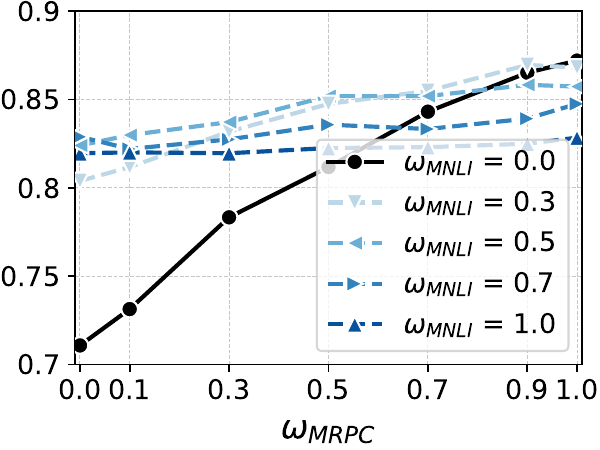}
        \caption{MRPC}
        \label{fig:fig1.c}
    \end{subfigure}
    \begin{subfigure}{0.24\textwidth}
        \centering
        \includegraphics[width=\textwidth]{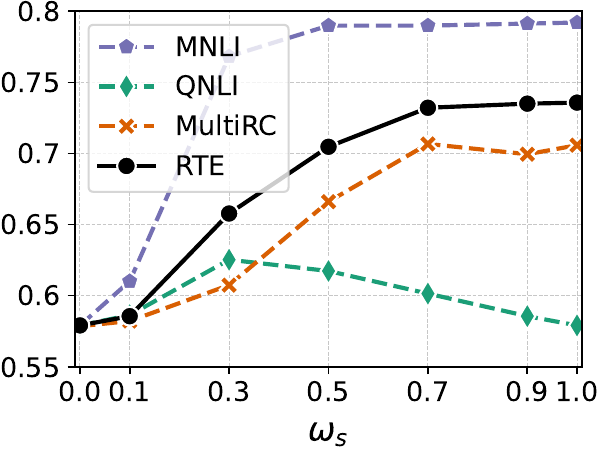}
        \includegraphics[width=\textwidth]{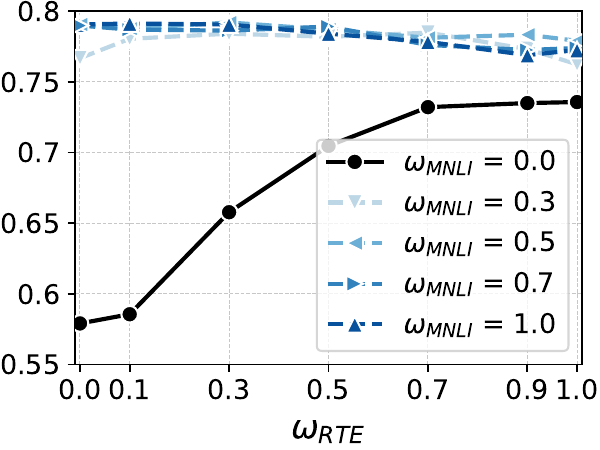}
        \caption{RTE}
        \label{fig:fig1.d}
    \end{subfigure}
    
    \caption{Probing results of 4 target tasks in various transfer learning conditions. (Top) 
    Effect of scaling the output representations of adapters by weight $\omega_s$ using different source adapters.
    (Bottom)
    Effect of combining independently scaled output representations of two adapters trained on the target task and MNLI, respectively. Each point shows the mean over 5 seeds.}
    \label{fig:fig1}
\end{figure*}

As mentioned in \S~\ref{sec:method}, we conducted preliminary experiments in which we scaled the output representations of adapters -- in isolation and combining two of them each.
We use the GLUE~\citep{GLUE} and SuperGLUE~\citep{SuperGLUE} benchmarks (cf. Appendix~\ref{sec:all-hyperparams}) and train a Pfeiffer adapter~\citep{pfeiffer-etal-2021-adapterfusion} on each task using \rev{the encoder LM} $\text{RoBERTa\textsubscript{BASE}}$~\citep{roberta}. In our probing-like setup~\citep{tenney-etal-2019-bert}, we freeze both the backbone and adapter weights and train a new task head on target task $t$ each time we change the scaling factor. 
For full clarity, we first show the effect of scaling output representations of adapters on a subset of tasks from GLUE and SuperGLUE in Figure~\ref{fig:fig1}, and then show the remaining ones in Figure~\ref{fig:fig.a1} as well as Figure~\ref{fig:fig.a4}.
Complete descriptions of the datasets, hyperparameters, and training procedure are provided in \S~\ref{sec:setup} and Appendix~\ref{sec:all-hyperparams}. 


We start by analyzing the performance change of a target task when scaling the output representations of the adapter of \emph{one} given source task. We define $\omega_s$ as the scaling value in the range of $[0,1]$, multiplied by the output representations $\vo^{l}_{s}$ of the source task $s$ in all layers, such that $\vo^{l}_{t} = \omega_{s}\vo^{l}_{s}$. 
Figure~\ref{fig:fig1} (Top) shows the probing results on four target tasks (each column), given various scaling weights applied to four source tasks (one of which is the respective target task). The results show that, while increasing the scaling weights generally improves the performance, 
the optimal value is not necessarily at $\omega_s=1$. In particular, there exist instances with $0<\omega_s<1$ reaching better performance than $\omega_s = 1$. This suggests that \emph{partial knowledge transfer} of tasks may be more beneficial.
Notably, and as also reported in previous studies~\citep{poth-pretrain, intermediate-task-transfer}, some source tasks such as MNLI show strong transfer learning abilities. 

Next, we go one step further by assessing the scaled \emph{combination} of the output vectors of two adapters. We focus on MNLI as one of the source tasks given its observed benefit in transfer learning, and set the second source adapter (denoted by $s$) to the one corresponding to the target task. We use two scaling parameters $\omega_{\text{MNLI}}$ and $\omega_{s}$ to scale $\vo^{l}_{\text{MNLI}}$ and $\vo^{l}_{s}$, respectively. The resulting output vector is defined as: $\vo^{l}_{t}=\omega_{s}\vo^{l}_{s}+\omega_{\text{MNLI}}\vo^{l}_{\text{MNLI}}$. 
Figure~\ref{fig:fig1} (Bottom) shows the results for various values of $\omega_{\text{MNLI}}$ and $\omega_{s}$. 
Combining the information encapsulated within multiple adapters through scaling can result in improved performance. 
Interestingly, in some cases, the best combination of $\omega_{\text{MNLI}}$ and $\omega_{s}$ does not add up to $1$, i.e., $\omega_{t} + \omega_{s} \neq 1$. These initial experiments -- while only covering a simple combination of up to two source tasks -- provide insights into the benefits of scaling representations for transfer learning.

\begin{table*}[t]
    \centering
    \resizebox{\linewidth}{!}{%
    \begin{tabular}{llllllllll|l}
    \toprule
    \textbf{Model} & \textbf{Constraint} & \multicolumn{1}{c}{\textbf{MNLI}} & \multicolumn{1}{c}{\textbf{QQP}}  & \multicolumn{1}{c}{\textbf{QNLI}} & \multicolumn{1}{c}{\textbf{SST-2}} & \multicolumn{1}{c}{\textbf{STS-B}} & \multicolumn{1}{c}{\textbf{MRPC}} & \multicolumn{1}{c}{\textbf{RTE}}  & \multicolumn{1}{c}{\textbf{CoLA}} & \multicolumn{1}{c}{\textbf{Avg.}} \\\midrule
    \modelours & None (original) & $86.97_{0.09}$ & $90.32_{0.10}$ & $\bm{92.51}_{0.17}$ & $93.88_{0.18}$ & $\bm{90.96}_{0.16}$ & $\bm{87.75}_{0.58}$ & $\bm{82.06}_{1.37}$ & $58.47_{1.76}$ & $\bm{85.36}_{0.55}$ \\
    \modelours & Mean & $\bm{87.03}_{0.01}$ & $90.36_{0.30}$ & $92.34_{0.09}$ & $92.60_{1.38}$ & $90.62_{0.25}$ & $87.11_{0.79}$ & $79.21_{1.82}$ & $\bm{59.87}_{2.95}$ & $84.89_{0.95}$ \\
    \modelours & Softmax & $86.85_{0.05}$ & $\bm{90.60}_{0.05}$ & $92.74_{0.22}$ & $93.75_{0.08}$ & $90.66_{0.10}$ & $85.83_{1.09}$ & $79.28_{1.04}$ & $58.43_{1.98}$ & $84.77_{0.58}$ \\
    \bottomrule
    \end{tabular}
    }
    \caption{Effect of adding various constraints to the scaling values of \modeladapteroursVecSplit, evaluated on GLUE using $\text{RoBERTa\textsubscript{BASE}}$. The constraints \textit{mean} and \textit{softmax} are applied over the task dimension, enforcing ${\sum_{s=1}^{|S|} \bm{\omega}^{l}_{s}=1}$. The best results are shown in \textbf{bold}.}
    \label{tab:results-constraints-glue}
\end{table*}

\begin{figure*}[t]
    \center
    \begin{subfigure}{0.28\textwidth}
        \centering
        \includegraphics[width=\textwidth]{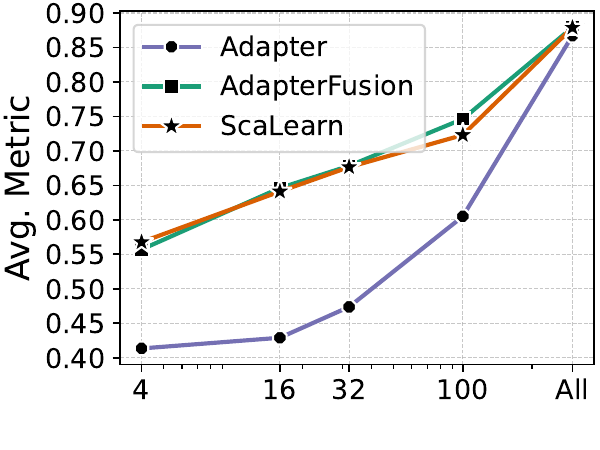}
        \caption{GLUE}
        \label{fig:fig-few.a.l}
    \end{subfigure}
    \begin{subfigure}{0.28\textwidth}
        \centering
        \includegraphics[width=\textwidth]{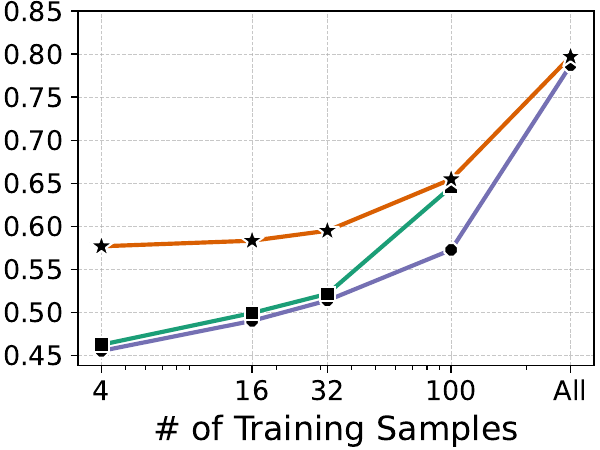}
        \caption{SuperGLUE}
        \label{fig:fig-few..l}
    \end{subfigure}
     \begin{subfigure}{0.28\textwidth}
        \centering
        \includegraphics[width=\textwidth]{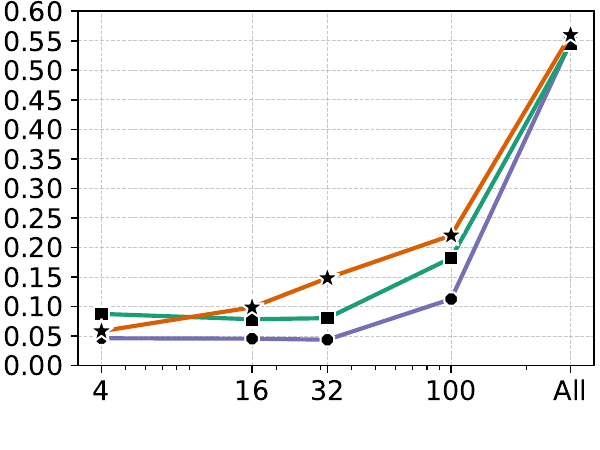}
        \caption{HumSet}
        \label{fig:fig-few.c.l}
    \end{subfigure}
    \caption{Few-shot learning results ($k= \text{\{4,16,32,100\}}$) comparing \modeladapter, \modeladapterfusion, and \modelours using $\text{RoBERTa\textsubscript{LARGE}}$ on three benchmarks. We show the mean across 5 seeds. For \modeladapterfusion and \modelours, we assume that there is a Pfeiffer adapter trained on the target task on $k$ samples and a Pfeiffer adapter trained on all samples for all other tasks available.}
    \label{fig:few-shot-large}
\end{figure*}

\subsection{Ablation Study}\label{sec:additional-ablation}

Table~\ref{tab:results-constraints-glue} shows the effect of adding constraints on the distributional values of scaling coefficient in \modelours, evaluated on GLUE using $\text{RoBERTa\textsubscript{BASE}}$. In particular, we change the original \modelours model by adding the constraints \textit{mean} and \textit{softmax} over the source task dimension, thus enforcing $\sum_{s=1}^{|S|} \bm{\omega}^{l}_{s}=1$. The results indicate that both constraints reduce average performance compared to those having no constraints, confirming our choice of directly learning the scaling coefficients without imposing any restrictions.

\subsection{Scaling Coefficient Visualizations}\label{sec:activation-vis}
\modeladapteroursScalarSplit and \modeladapteroursScalarShare utilize uniform scaling and learn coefficients that are directly used to scale the output representations of the source adapters. In the following, we leverage this characteristic to provide an analysis of the potential degrees of effects of source tasks on target tasks. We present the adapter weights learned using $\text{RoBERTa\textsubscript{BASE}}$ for GLUE and SuperGLUE, and using $\text{XLM-R\textsubscript{BASE}}$ for HumSet with the random seed set to \texttt{0}.

The learned coefficients of each LM layer on GLUE, SuperGLUE, and HumSet of \modeladapteroursScalarSplit are shown in Figure~\ref{fig:fig.act.1}, Figure~\ref{fig:fig.act.2}, and Figure~\ref{fig:fig.act.3}, respectively. The weights reveal that in most cases, the actual target task adapter is activated most strongly across the layers. Among the source tasks, most weights are close to $0$, while some source tasks also show high values, particularly in some of the higher layers of the LM. 
Interestingly, some of the scaling coefficients go beyond or even below $1$, which would not have been possible in the traditional paradigm where scaling coefficients combining multiple vectors are restricted to sum up to 1.

The learned weights on GLUE, SuperGLUE, and HumSet of \modeladapteroursScalarShare are shown in Figure~\ref{fig:fig.act.4}. \modeladapteroursScalarShare also mostly activates the actual target task adapter, whereas this effect is comparatively weaker in SuperGLUE and stronger in HumSet. As is the case with \modeladapteroursScalarSplit, many scaling coefficients exceed or go below $1$.

\subsection{Additional Results}\label{sec:additional-results}

\vspace{0.5em}  
\noindent\textbf{More results using $\text{RoBERTa\textsubscript{BASE}}$.}  Table~\ref{tab:results-gsg-full} shows the results when training on the combination of all GLUE and SuperGLUE tasks, resulting in a total of $15$ tasks. 

\vspace{0.5em}  
\noindent\textbf{Results using $\text{RoBERTa\textsubscript{LARGE}}$.} 
We further validate our method and its variations on the encoder LM $\text{RoBERTa\textsubscript{LARGE}}$.
Table~\ref{tab:results-glue-large} shows the corresponding results, including all baselines, on the GLUE benchmark. Table~\ref{tab:results-superglue-large} shows the results on SuperGLUE. Table~\ref{tab:results-humset-avg-large} shows the results on HumSet. Finally, Table~\ref{tab:results-gsg-full-large} shows the results when training on the combination of all GLUE and SuperGLUE tasks, resulting in a total of $15$ tasks.


\subsection{Complete Few-Shot Results}\label{sec:more-fs-results}

To obtain a more complete understanding of the few-shot capabilities of \modeladapter, \modeladapterfusion, and \modelours, we show few-shot transfer learning results for each dataset, as well as for every variant of \modelours (cf. \S~\ref{sec:few-shot}).

\vspace{0.5em}  
\noindent\textbf{Few-shot results using $\text{RoBERTa\textsubscript{BASE}}$.}
Table~\ref{tab:results-few-glue-full} shows the few-shot transfer learning performance of the methods on the GLUE benchmark using $k=\text{\{4,16,32,100\}}$ samples. Table~\ref{tab:results-few-superglue-full} shows the performance of the methods on SuperGLUE. Table~\ref{tab:results-few-humset-avg} shows the performance of the methods on HumSet (on $\text{XLM-R)\textsubscript{BASE}}$. Finally, Table~\ref{tab:results-few-gsg-full} shows the results when training on the combination of all GLUE and SuperGLUE tasks, resulting in $|S|=15$ source tasks.

\vspace{0.5em}  
\noindent\textbf{Few-shot results using $\text{RoBERTa\textsubscript{LARGE}}$.}
Figure~\ref{fig:few-shot-large} provides an overview, comparing the few-shot learning capabilities of \modeladapter, \modeladapterfusion, and \modelours when using $\text{RoBERTa\textsubscript{LARGE}}$.
Moreover, Table~\ref{tab:results-few-glue-full-large} shows the few-shot learning performance of the methods on the GLUE benchmark using $k=\text{\{4,16,32,100\}}$ samples. Table~\ref{tab:results-few-superglue-full-large} shows the performance of the methods on SuperGLUE. Table~\ref{tab:results-few-humset-avg-large} shows the performance of the methods on HumSet (on $\text{XLM-R\textsubscript{LARGE}}$). Finally, Table~\ref{tab:results-few-gsg-full-large} shows the results when training on the combination of all GLUE and SuperGLUE tasks.


\begin{figure}[h]
    \centering
    \begin{subfigure}{0.43\textwidth}
        \centering
        \includegraphics[width=\textwidth]{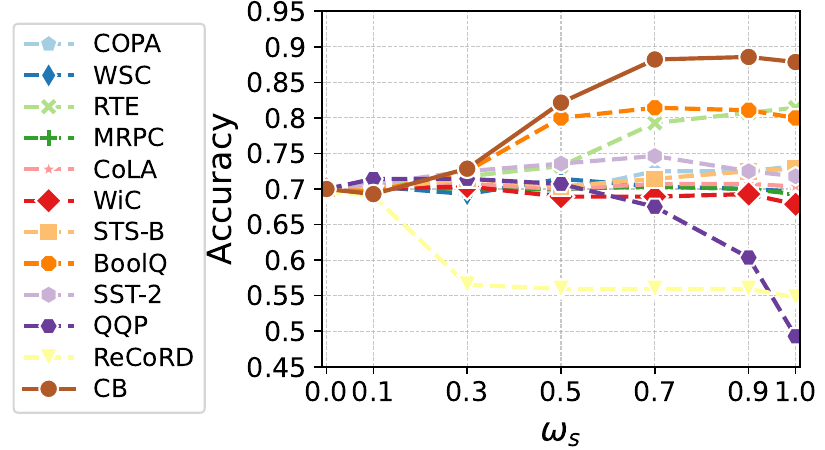}
        \caption{CB}
        \label{fig:fig.a1.a}
    \end{subfigure}
    \begin{subfigure}{0.43\textwidth}
        \centering
        \includegraphics[width=\textwidth]{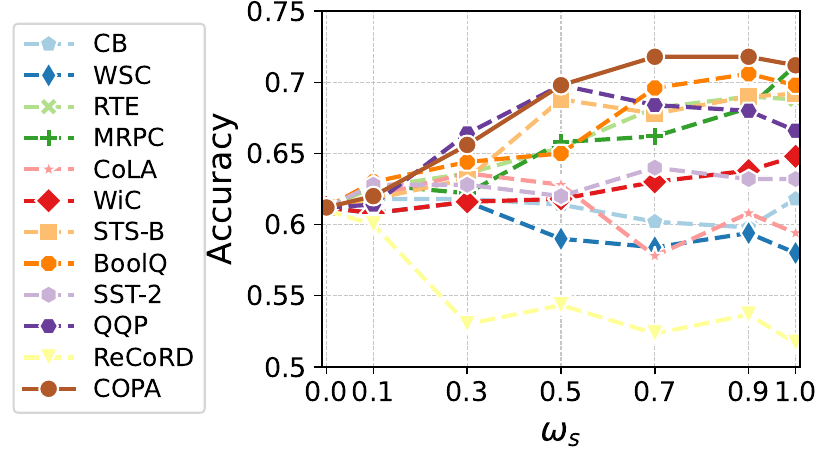}
        \caption{COPA}
        \label{fig:fig.a1.b}
    \end{subfigure}
    \begin{subfigure}{0.43\textwidth}
        \centering
        \includegraphics[width=\textwidth]{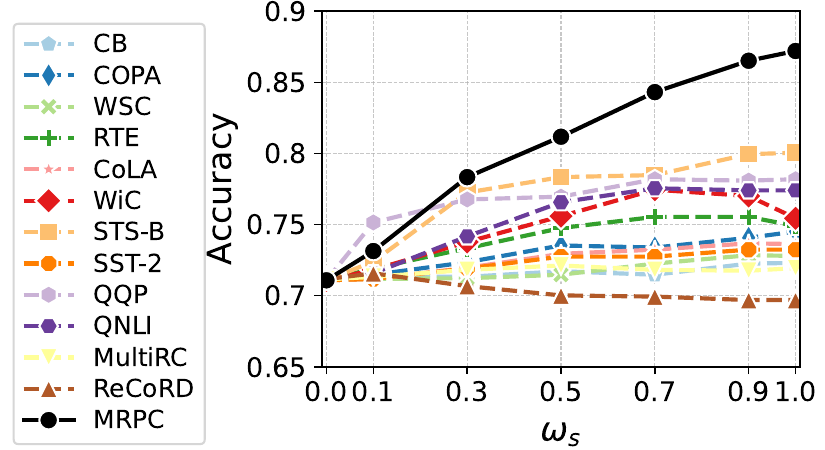}
        \caption{MRPC}
        \label{fig:fig.a1.c}
    \end{subfigure}
    \begin{subfigure}{0.43\textwidth}
        \centering
        \includegraphics[width=\textwidth]{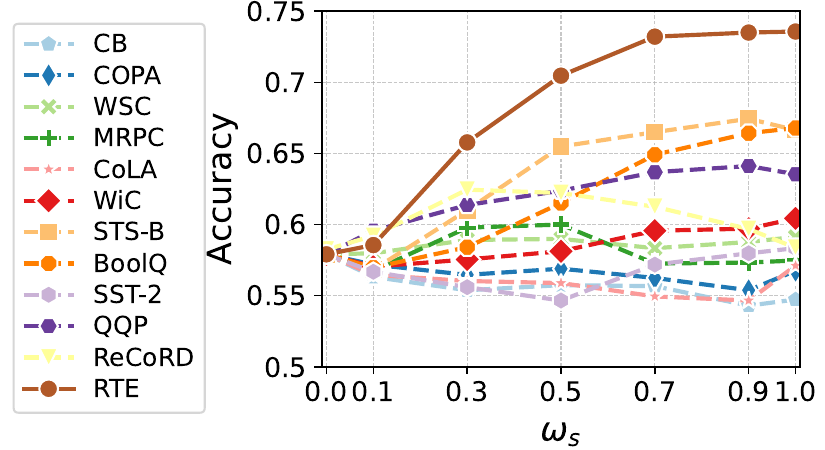}
        \caption{RTE}
        \label{fig:fig.a1.d}
    \end{subfigure}
    \begin{subfigure}{0.43\textwidth}
        \centering
        \includegraphics[width=\textwidth]{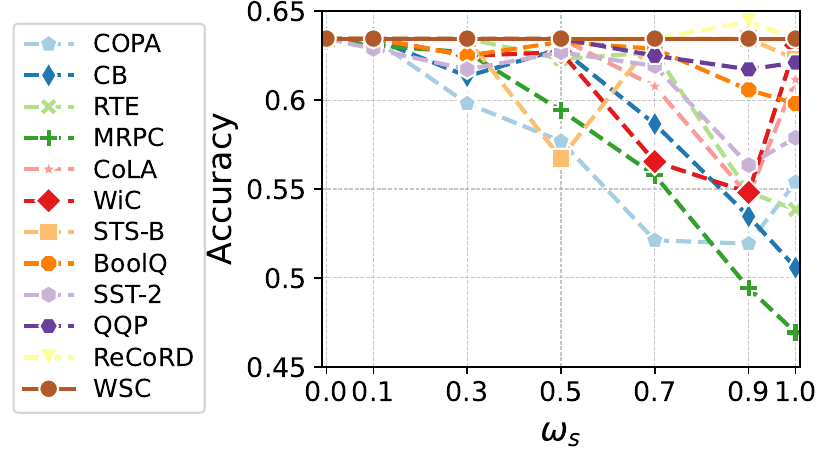}
        \caption{WSC}
        \label{fig:fig.a1.e}
    \end{subfigure}
    \caption{Effect of scaling the output representations $\vo^{l}_{s}$ of adapters by weight $\omega_s$ using different source adapters from all other tasks from GLUE and SuperGLUE.
    Each point shows the mean over 5 seeds.
    }
    \label{fig:fig.a1}
\end{figure}

\begin{figure}[h]
    \centering
    \begin{subfigure}{0.26\textwidth}
        \centering
        \includegraphics[width=\textwidth]{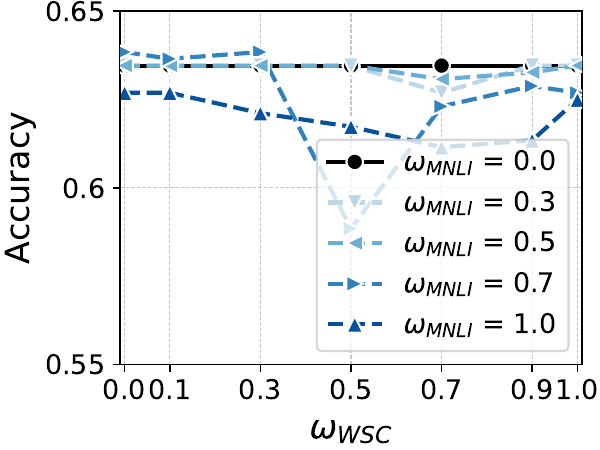}
        \caption{WSC}
        \label{fig:fig.a4.a}
    \end{subfigure}
    \begin{subfigure}{0.26\textwidth}
        \centering
        \includegraphics[width=\textwidth]{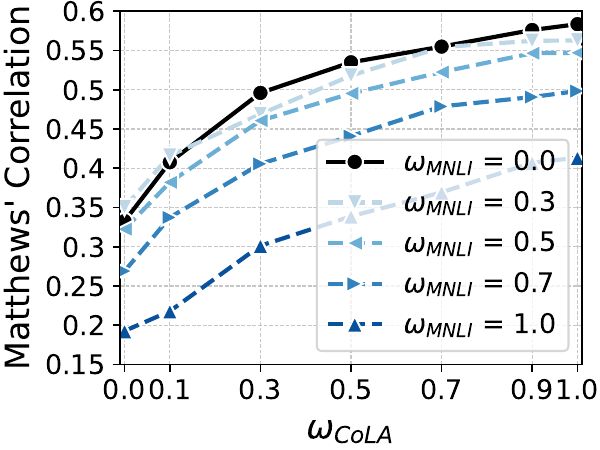}
        \caption{CoLA}
        \label{fig:fig.a4.b}
    \end{subfigure}
    \begin{subfigure}{0.26\textwidth}
        \centering
        \includegraphics[width=\textwidth]{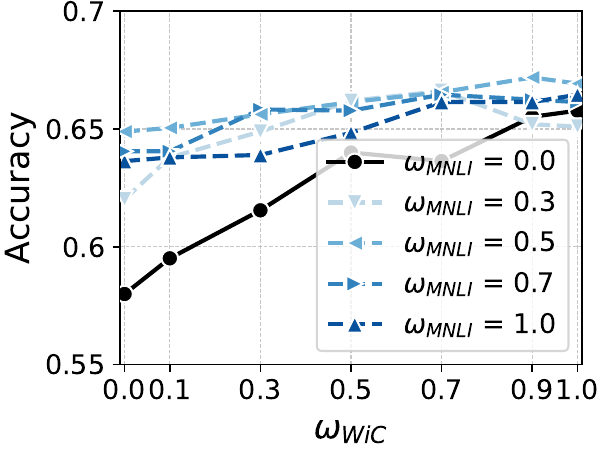}
        \caption{WiC}
        \label{fig:fig.a4c}
    \end{subfigure}
    \begin{subfigure}{0.26\textwidth}
        \centering
        \includegraphics[width=\textwidth]{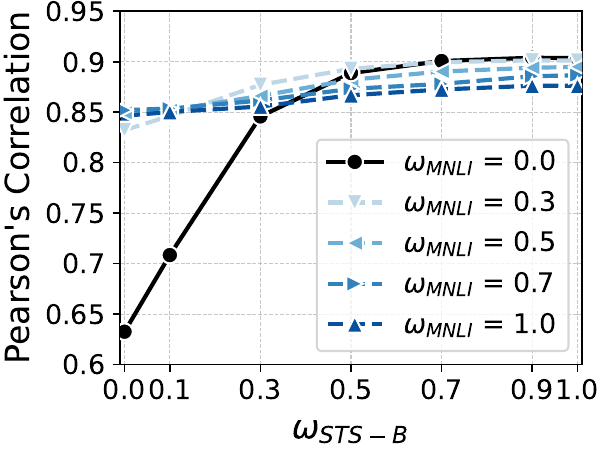}
        \caption{STS-B}
        \label{fig:fig.a4.d}
    \end{subfigure}
    \begin{subfigure}{0.26\textwidth}
        \centering
        \includegraphics[width=\textwidth]{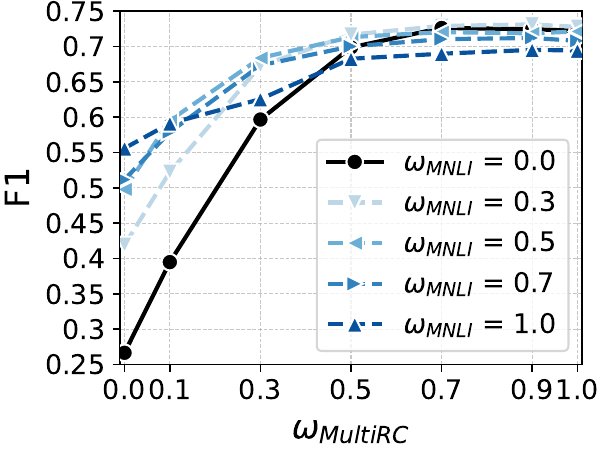}
        \caption{MultiRC}
        \label{fig:fig.a4.e}
    \end{subfigure}
    \begin{subfigure}{0.26\textwidth}
        \centering
        \includegraphics[width=\textwidth]{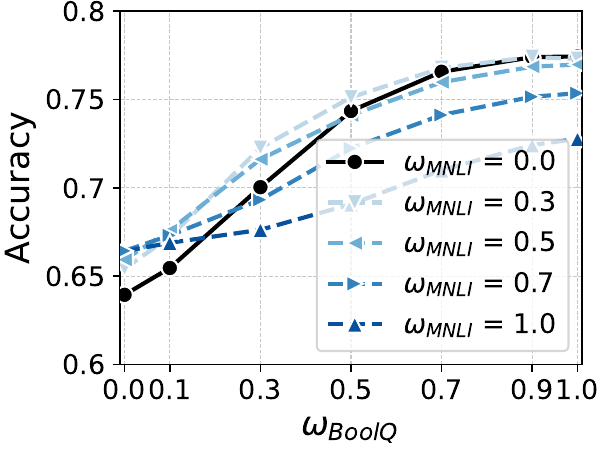}
        \caption{BoolQ}
        \label{fig:fig.a4f}
    \end{subfigure}
    \caption{
    Effect of combining independently scaled output representations of two adapters trained on the target task and MNLI, respectively, on additional tasks from GLUE and SuperGLUE. Each point shows the mean over 5 seeds.
    }
    \label{fig:fig.a4}
\end{figure}

\onecolumn
\begin{figure}[h]
    \centering
    \begin{subfigure}{\textwidth}
        \centering
        \includegraphics[width=0.9\textwidth]{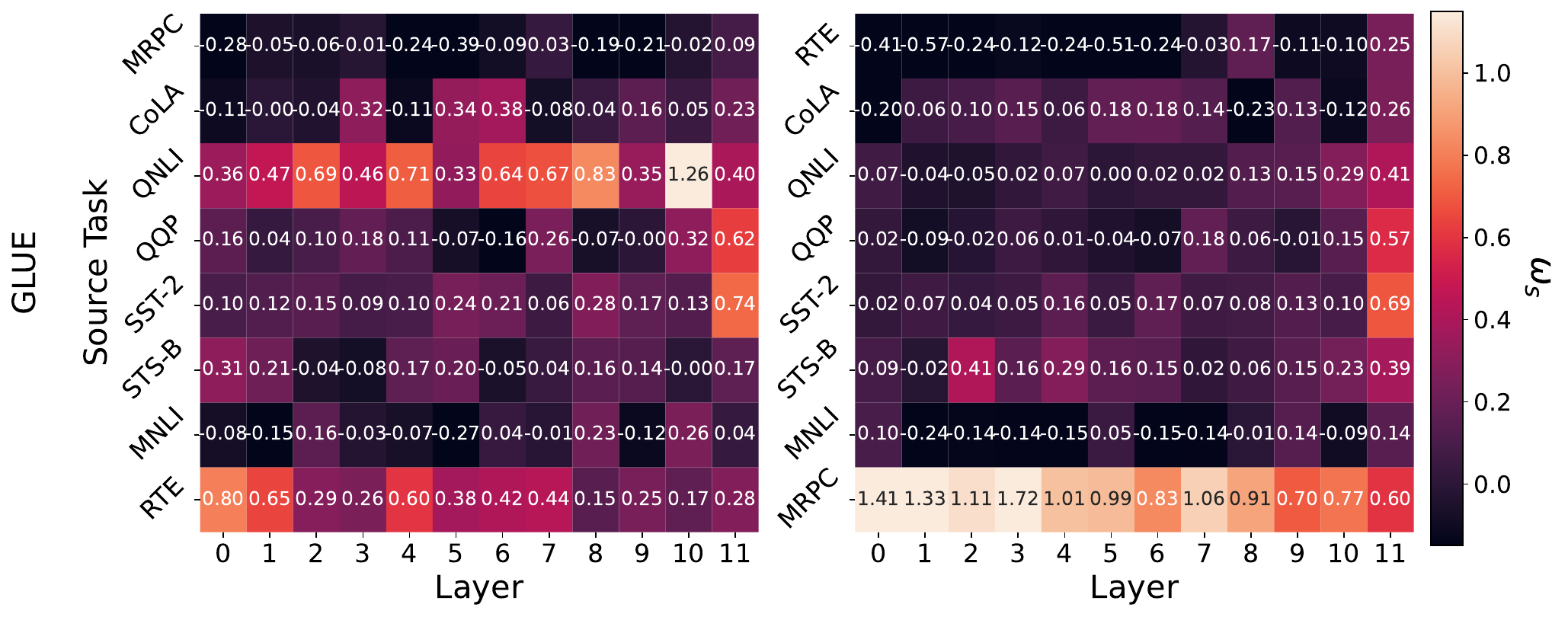}
        \label{fig:fig.act.u1}
    \end{subfigure}
    \begin{subfigure}{\textwidth}
        \centering
        \includegraphics[width=0.9\textwidth]{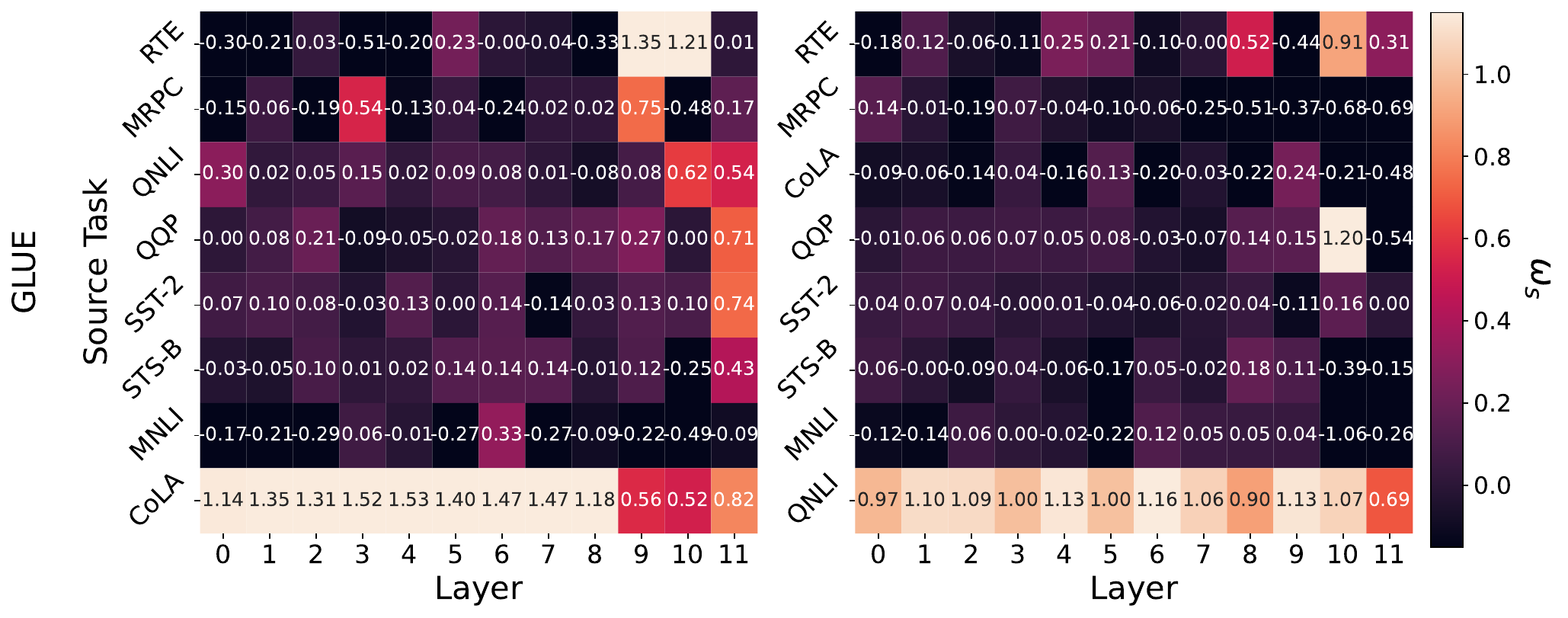}
        \label{fig:fig.act.u2}
    \end{subfigure}
    \begin{subfigure}{\textwidth}
        \centering
        \includegraphics[width=0.9\textwidth]{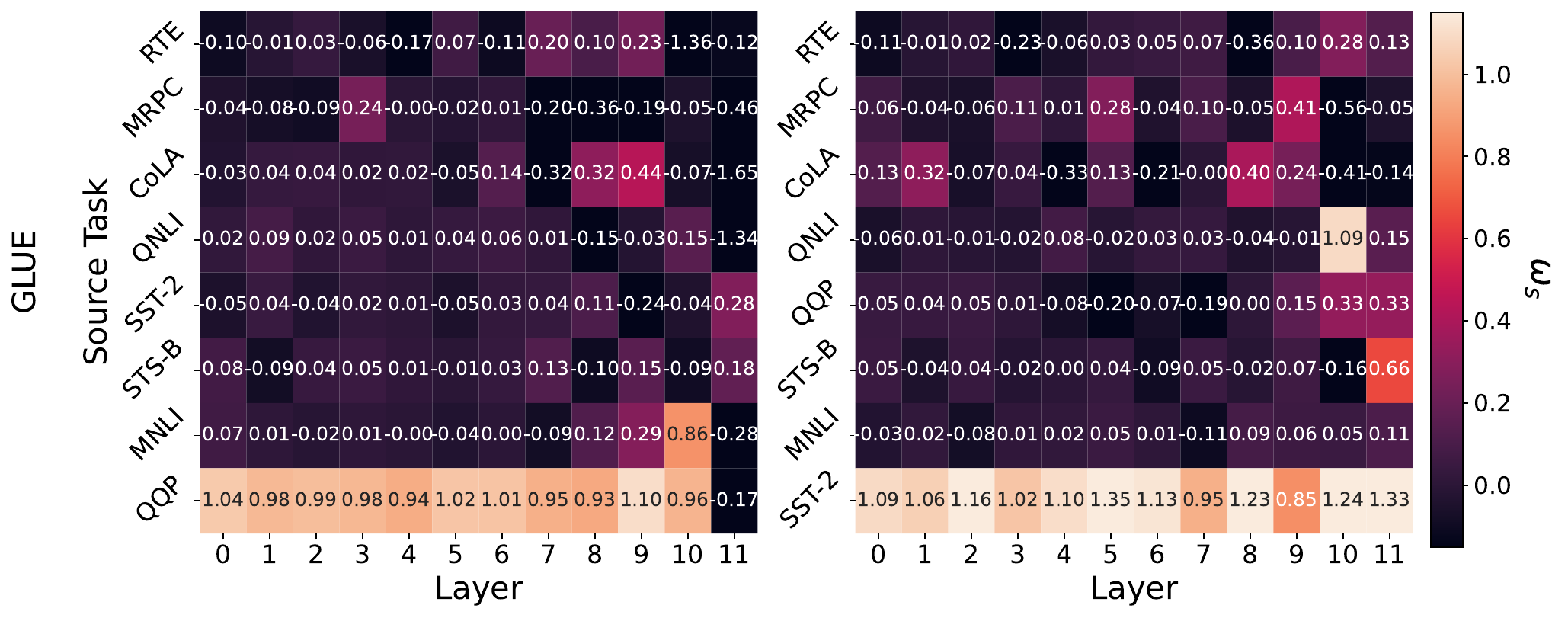}
        \label{fig:fig.act.u3}
    \end{subfigure}
    \begin{subfigure}{\textwidth}
        \centering
        \includegraphics[width=0.9\textwidth]{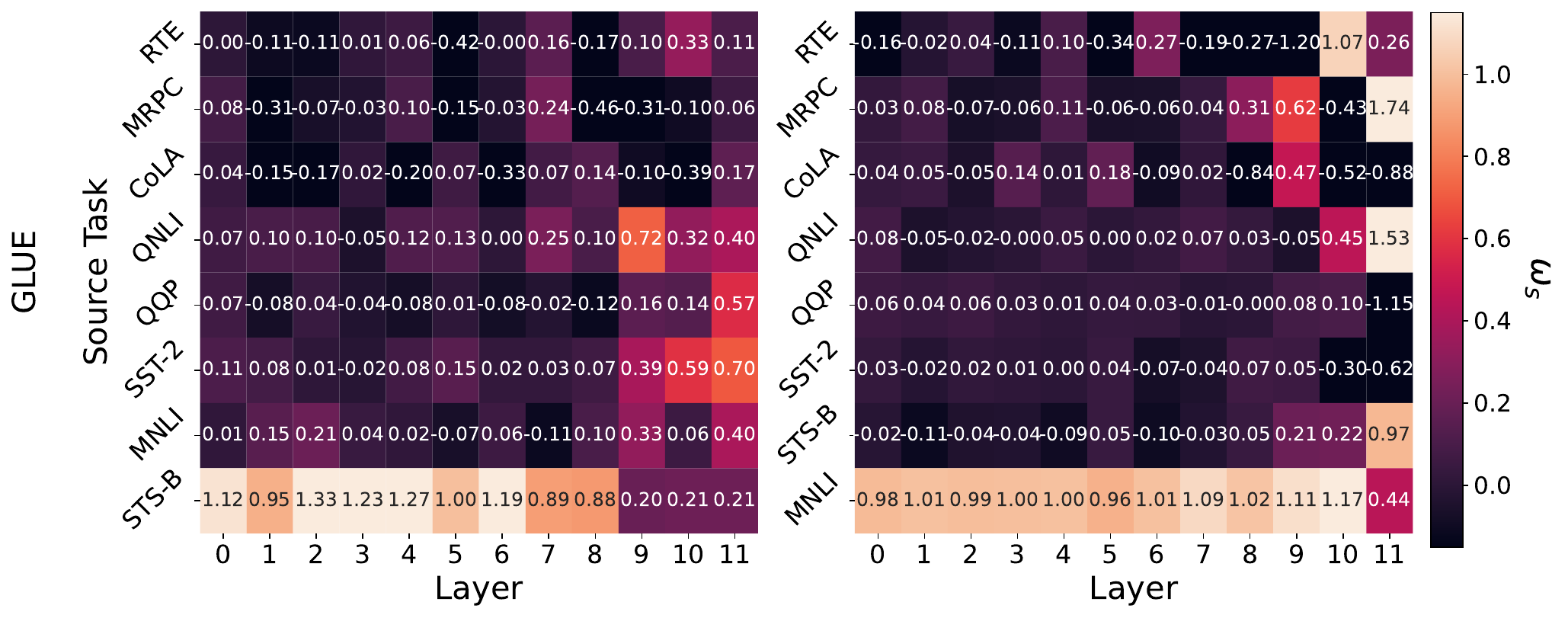}
        \label{fig:fig.act.u4}
    \end{subfigure}
    \caption{\modeladapteroursScalarSplit scaling coefficients on GLUE using $\text{RoBERTa\textsubscript{BASE}}$ on seed $\texttt{0}$. Target tasks are shown in the last index of each heatmap.}
    \label{fig:fig.act.1}
\end{figure}

\begin{figure}[t]
    \centering
    \begin{subfigure}{\textwidth}
        \centering
        \includegraphics[width=0.9\textwidth]{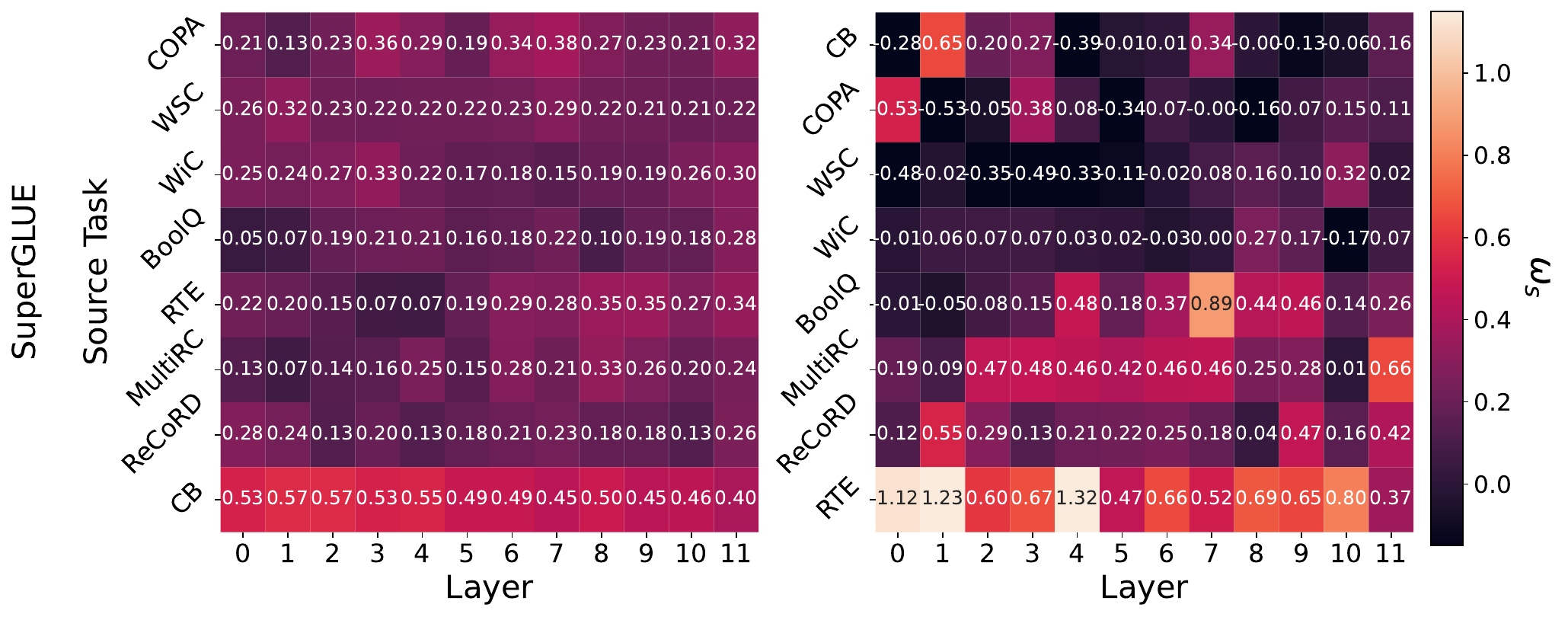}
        \label{fig:fig.act.u5}
    \end{subfigure}
    \begin{subfigure}{\textwidth}
        \centering
        \includegraphics[width=0.9\textwidth]{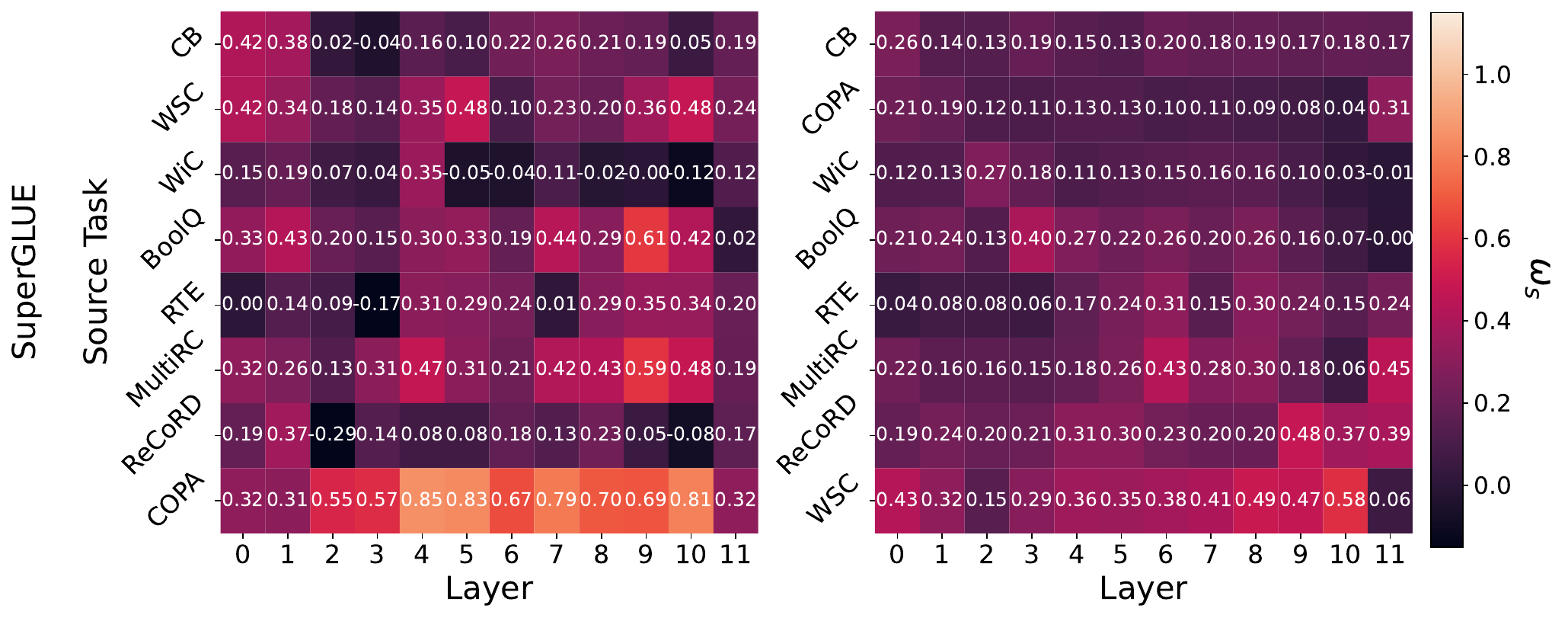}
        \label{fig:fig.act.u6}
    \end{subfigure}
    \begin{subfigure}{\textwidth}
        \centering
        \includegraphics[width=0.9\textwidth]{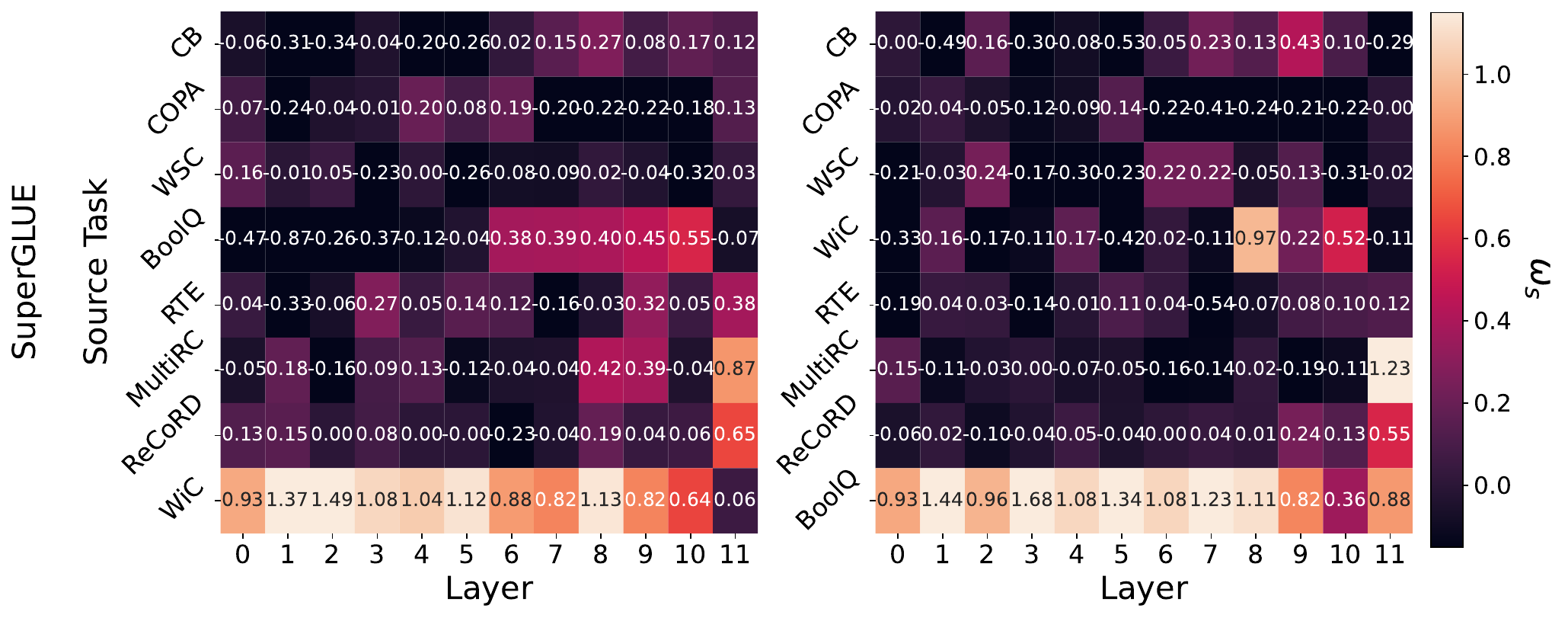}
        \label{fig:fig.act.u7}
    \end{subfigure}
    \begin{subfigure}{\textwidth}
        \centering
        \includegraphics[width=0.9\textwidth]{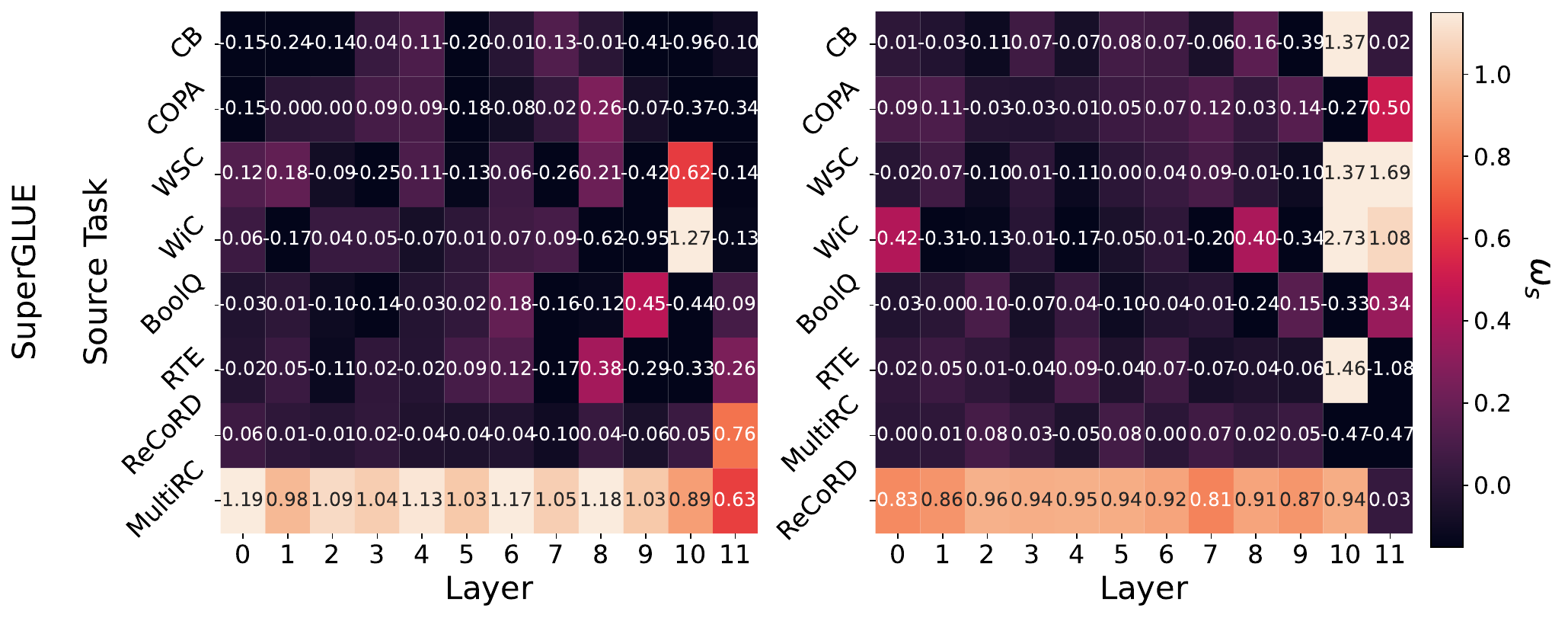}
        \label{fig:fig.act.u8}
    \end{subfigure}
    
    \caption{\modeladapteroursScalarSplit scaling coefficients on SuperGLUE using $\text{RoBERTa\textsubscript{BASE}}$ on seed $\texttt{0}$. Target tasks are shown in the last index of each heatmap.}
    \label{fig:fig.act.2}
\end{figure}

\begin{figure}[t]
    \centering
    \begin{subfigure}{\textwidth}
        \centering
        \includegraphics[width=0.9\textwidth]{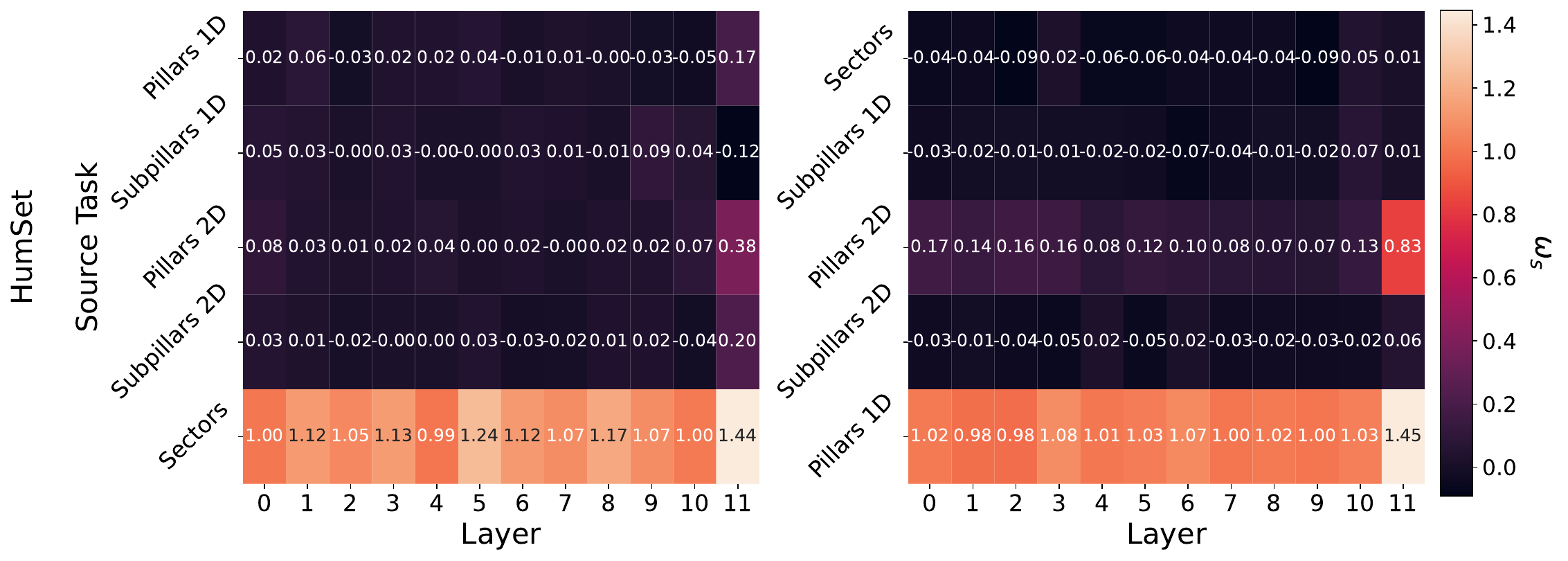}
        \label{fig:fig.act.h1}
    \end{subfigure}
    \begin{subfigure}{\textwidth}
        \centering
        \includegraphics[width=0.9\textwidth]{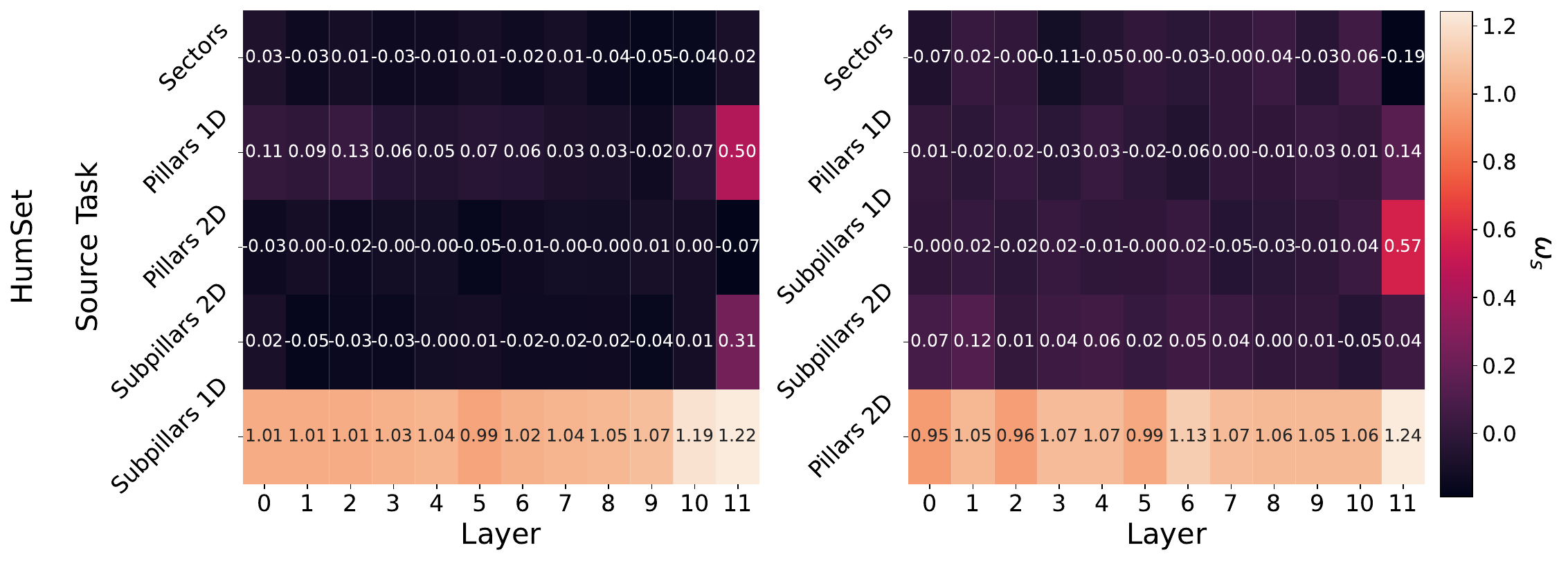}
        \label{fig:fig.act.h2}
    \end{subfigure}
    \begin{subfigure}{\textwidth}
        \centering
        \includegraphics[width=0.5\textwidth]{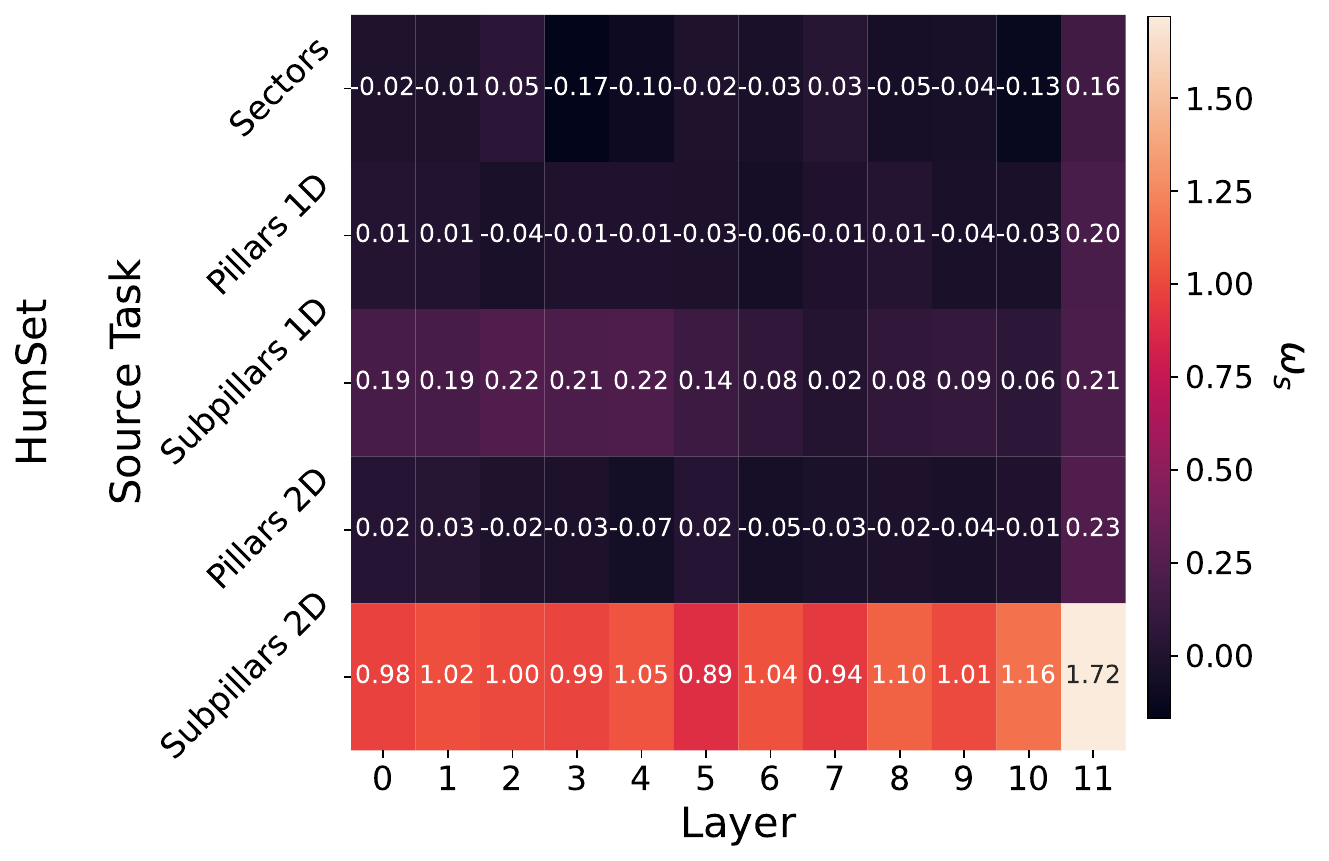}
        \label{fig:fig.act.h3}
    \end{subfigure}
    
    \caption{\modeladapteroursScalarSplit scaling coefficients on HumSet using $\text{XLM-R\textsubscript{BASE}}$ on seed $\texttt{0}$. Target tasks are shown in the last index of each heatmap.}
    \label{fig:fig.act.3}
\end{figure}

\begin{figure}[h]
    \centering
    \begin{subfigure}{\textwidth}
        \centering
        \includegraphics[width=\textwidth]{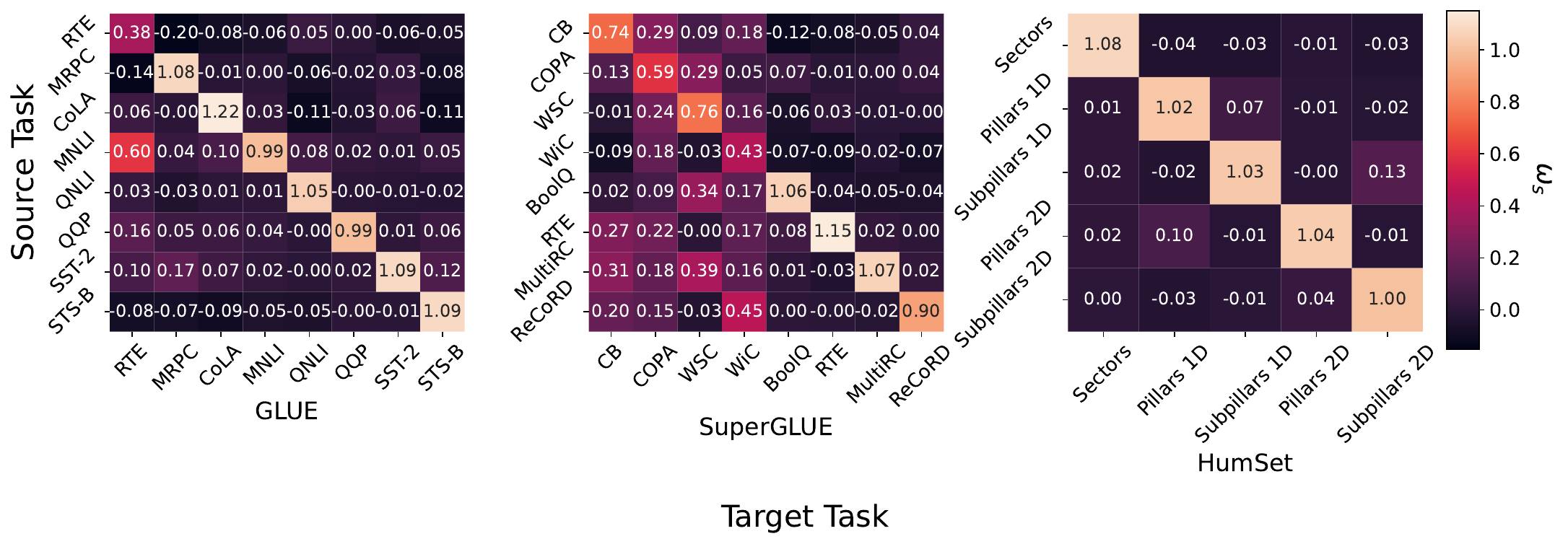}
        \label{fig:fig.act.uPP1}
    \end{subfigure}
    
    \caption{\modeladapteroursScalarShare scaling coefficients on GLUE, SuperGLUE, and HumSet using $\text{RoBERTa\textsubscript{BASE}}$ for GLUE and SuperGLUE and $\text{XLM-R\textsubscript{BASE}}$ for HumSet on seed $\texttt{0}$.}
    \label{fig:fig.act.4}
\end{figure}

\twocolumn

\begin{table*}[t]
\centering
\resizebox{1\linewidth}{!}{%
\begin{tabular}{lcccccccc|c}
\toprule
\textbf{Model} & \textbf{MNLI} & \textbf{QQP}  & \textbf{QNLI} & \textbf{SST-2} & \textbf{STS-B} & \textbf{MRPC} & \textbf{RTE}  & \textbf{CoLA} & \textbf{Avg.} \\\midrule

\modelfine & $89.57_{0.36}$ & $89.75_{1.03}$ & $93.91_{0.43}$ & $95.30_{0.65}$ & $91.89_{0.35}$ & $86.27_{1.15}$ & $81.52_{3.19}$ & $60.15_{2.89}$ & $86.04_{1.26}$ \\
\modeladapter & $89.62_{0.18}$ & $89.87_{0.67}$ & $94.13_{0.06}$ & $95.24_{0.08}$ & $91.81_{0.29}$ & $87.82_{2.11}$ & $81.23_{2.92}$ & $64.07_{1.97}$ & $86.72_{1.04}$ \\
\modelpropetl & $89.78_{0.24}$ & $89.23_{0.77}$ & \underline{$94.32$}$_{0.09}$ & $95.41_{0.00}$ & $91.45_{0.39}$ & $87.65_{0.73}$ & $84.55_{2.14}$ & $65.85_{2.10}$ & $87.28_{0.81}$ \\
\modelcompact & $89.15_{0.67}$ & $87.33_{2.39}$ & $92.93_{1.42}$ & $95.41_{0.00}$ & $91.46_{0.35}$ & $87.84_{1.23}$ & $79.71_{4.58}$ & $65.66_{2.08}$ & $86.19_{1.59}$ \\ 
\rev{\modelia} & $\rev{88.69_{0.61}}$ & $\rev{87.79_{0.72}}$ & $\rev{91.72_{0.79}}$ & $\rev{94.95_{0.16}}$ & $\rev{91.39_{0.45}}$ & $\rev{86.37_{1.65}}$ & $\rev{80.79_{3.16}}$ & $\rev{64.70_{3.20}}$ & $\rev{85.80_{1.34}}$ \\
\modellora & $89.66_{0.27}$ & $89.66_{0.10}$ & $94.20_{0.28}$ & $95.47_{0.24}$ & $91.98_{0.13}$ & $87.70_{1.14}$ & $80.51_{2.03}$ & $63.80_{2.71}$ & $86.62_{0.86}$ \\\midrule
\modelfineMT & $87.95_{0.39}$ & $89.82_{0.77}$ & $92.58_{0.32}$ & $94.88_{0.94}$ & $87.04_{0.68}$ & $81.37_{1.00}$ & $84.36_{1.19}$ & $55.32_{0.78}$ & $84.16_{0.76}$ \\
\modeladapterMT & $89.10_{0.36}$ & $89.35_{0.09}$ & $93.64_{0.05}$ & $94.90_{0.17}$ & $88.40_{0.32}$ & $83.09_{0.25}$ & $86.64_{0.00}$ & $56.38_{0.79}$ & $85.19_{0.25}$ \\
\modelpropetlMT & $88.98_{0.33}$ & $89.03_{0.15}$ & $94.14_{0.11}$ & $95.15_{0.05}$ & $91.56_{0.23}$ & $87.83_{1.10}$ & \underline{$88.45$}$_{0.29}$ & $60.99_{1.03}$ & $87.01_{0.41}$ \\
\modelhyper & $89.66_{0.40}$ & $90.15_{0.63}$ & $93.95_{0.13}$ & $95.80_{0.62}$ & $91.68_{0.35}$ & $86.60_{1.22}$ & $86.28_{0.29}$ & $61.18_{4.76}$ & $86.91_{1.05}$ \\
\modelhyperPP & $89.79_{0.21}$ & $89.54_{0.43}$ & $93.95_{0.54}$ & $95.22_{0.11}$ & $91.62_{0.29}$ & $88.07_{1.86}$ & $86.28_{1.06}$ & $65.16_{0.61}$ & $87.45_{0.64}$ \\
\midrule
\modeladapterfusion & $89.57_{0.17}$ & \underline{$\bm{90.88}$}$_{0.06}$ & $94.15_{0.04}$ & $95.87_{0.00}$ & $91.86_{0.15}$ & $88.97_{0.78}$ & $85.70_{1.13}$ & $66.39_{1.83}$ & $87.93_{0.52}$ \\
\rev{\modeladaptersoup} & $\rev{65.83_{0.51}}$ & $\rev{82.37_{0.00}}$ & $\rev{74.06_{1.01}}$ & $\rev{93.98_{0.24}}$ & $\rev{81.67_{1.63}}$ & $\rev{73.37_{0.51}}$ & $\rev{67.27_{1.63}}$ & $\rev{43.70_{1.62}}$ & $\rev{72.78_{0.89}}$ \\
\modeladapteroursVecSplit & $90.09_{0.09}$ & $90.51_{0.26}$ & $94.18_{0.03}$ & $95.41_{0.16}$ & $92.32_{0.15}$ & $88.09_{0.82}$ & $\bm{87.08}_{0.54}$ & $65.40_{2.62}$ & $87.91_{0.55}$ \\
\modeladapteroursScalarSplit & $90.11_{0.04}$ & $90.05_{0.28}$ & $\bm{94.23}_{0.08}$ & $95.41_{0.16}$ & $92.11_{0.06}$ & $88.63_{1.72}$ & $84.40_{3.93}$ & $66.98_{0.58}$ & $87.74_{0.86}$ \\
\modeladapteroursVecShare & \underline{$\bm{90.31}$}$_{0.10}$ & $90.59_{0.03}$ & $94.05_{0.03}$ & \underline{$\bm{95.93}$}$_{0.24}$ & \underline{$\bm{92.48}$}$_{0.15}$ & $88.48_{1.26}$ & $86.28_{1.05}$ & \underline{$\bm{67.13}$}$_{0.59}$ & \underline{$\bm{88.16}$}$_{0.43}$ \\
\modeladapteroursScalarShare & $90.08_{0.01}$ & $90.49_{0.02}$ & $94.12_{0.16}$ & $95.18_{0.16}$ & $92.12_{0.09}$ & \underline{$\bm{90.05}$}$_{0.54}$ & $84.98_{1.32}$ & $64.97_{0.85}$ & $87.75_{0.39}$ \\

\bottomrule
\end{tabular}
} 
\caption{Evaluation results on GLUE using $\text{RoBERTa\textsubscript{LARGE}}$. (Top) STL models, only learning a single task at a time. (Middle) Joint MTL methods, learning all tasks simultaneously. (Bottom) Two-stage MTL methods, composing the knowledge of several source adapters. The overall best results are \underline{underlined}, and the best results among the two-stage MTL models are shown in \textbf{bold}.}
\label{tab:results-glue-large}
\end{table*}

\begin{table*}[t]
    \centering
    \resizebox{1\textwidth}{!}{%
    \begin{tabular}{lcccccccc|c}
    \toprule
    \textbf{Model} & \textbf{ReCoRD} & \textbf{MultiRC} & \textbf{BoolQ} & \textbf{WiC}  & \textbf{WSC}  & \textbf{COPA} & \textbf{CB} & \textbf{RTE}  & \textbf{Avg.} \\\midrule
    \modelfine & $81.60_{1.25}$ & $79.03_{0.02}$ & $81.65_{0.30}$ & $69.72_{2.16}$ & \underline{$63.46$}$_{0.00}$ & $52.00_{8.28}$ & $90.36_{2.99}$ & $81.52_{3.19}$ & $74.92_{2.27}$ \\
    \modeladapter & $88.52_{0.09}$ & $80.73_{0.69}$ & $82.36_{0.72}$ & $69.16_{1.31}$ & $63.25_{0.64}$ & $71.90_{13.63}$ & $92.68_{1.78}$ & $81.23_{2.92}$ & $78.73_{2.72}$ \\
    \modelpropetl & $87.86_{2.59}$ & \underline{$81.19$}$_{0.99}$ & $81.61_{0.86}$ & $69.62_{2.16}$ & \underline{$63.46$}$_{0.00}$ & $69.00_{18.96}$ & $94.11_{4.04}$ & $84.55_{2.14}$ & $78.92_{3.97}$ \\
    \modelcompact & $88.34_{0.97}$ & $79.18_{0.29}$ & $79.53_{6.13}$ & $69.26_{1.51}$ & $62.26_{1.43}$ & $79.00_{9.74}$ & $87.50_{7.48}$ & $79.71_{4.58}$ & $78.10_{4.02}$ \\
    \rev{\modelia} & $\rev{87.47_{0.21}}$ & $\rev{77.91_{0.43}}$ & $\rev{80.97_{0.75}}$ & $\rev{68.65_{2.55}}$ & $\rev{60.58_{0.00}}$ & $\rev{77.00_{0.00}}$ & $\rev{90.00_{3.91}}$ & $\rev{80.79_{3.16}}$ & $\rev{77.93_{1.35}}$ \\
    \modellora & $88.30_{0.36}$ & $79.10_{0.29}$ & $78.02_{8.88}$ & $68.46_{2.07}$ & $62.12_{1.46}$ & $76.60_{19.22}$ & $92.86_{1.79}$ & $80.51_{2.03}$ & $73.58_{11.06}$ \\    \midrule
    \modelfineMT & $83.57_{0.81}$ & $78.08_{0.55}$ & $81.70_{0.65}$ & $53.03_{0.37}$ & $49.36_{9.50}$ & $86.67_{2.36}$ & $82.14_{2.92}$ & $83.87_{2.01}$ & $74.80_{2.39}$ \\
    \modeladapterMT & $86.76_{0.32}$ & $75.15_{0.24}$ & $77.18_{2.22}$ & $51.57_{1.12}$ & $53.21_{9.75}$ & $67.67_{1.25}$ & $80.95_{1.68}$ & $77.38_{1.36}$ & $71.23_{2.24}$ \\
    \modelpropetlMT & $84.83_{0.40}$ & $79.60_{0.37}$ & $82.02_{1.11}$ & $55.33_{0.46}$ & $59.62_{9.05}$ & $86.67_{4.03}$ & $88.10_{2.23}$ & $85.56_{0.29}$ & $77.71_{2.24}$ \\
    \modelhyper & $84.38_{1.00}$ & $79.68_{0.97}$ & $81.87_{0.97}$ & $53.81_{2.48}$ & \underline{$63.46$}$_{8.64}$ & $82.33_{6.94}$ & $83.93_{2.53}$ & \underline{$86.88$}$_{0.90}$ & $77.04_{3.05}$ \\
    \modelhyperPP & $13.66_{0.00}$ & $40.21_{40.21}$ & $71.50_{9.33}$ & $49.14_{0.86}$ & $62.98_{0.48}$ & $54.00_{3.00}$ & $67.86_{17.86}$ & $66.97_{19.68}$ & $53.29_{11.43}$ \\
    \midrule
    \modeladapterfusion & \underline{$\bm{89.21}$}$_{0.17}$ & $80.52_{0.24}$ & $82.21_{0.30}$ & $69.09_{1.68}$ & \underline{$\bm{63.46}$}$_{0.68}$ & $81.20_{16.07}$ & \underline{$\bm{95.71}$}$_{0.98}$ & $\bm{86.06}_{1.07}$ & $80.93_{2.65}$ \\
    \rev{\modeladaptersoup} & $\rev{70.33_{0.28}}$ & $\rev{38.42_{12.42}}$ & $\rev{73.20_{0.16}}$ & $\rev{62.23_{1.17}}$ & \rev{\underline{$\bm{63.46}$}$_{0.00}$} & $\rev{54.50_{5.74}}$ & $\rev{68.75_{1.03}}$ & $\rev{61.37_{3.97}}$ & $\rev{61.53_{3.06}}$ \\
    \modeladapteroursVecSplit & $87.85_{0.01}$ & $78.40_{0.70}$ & $80.29_{2.52}$ & $68.56_{1.68}$ & $62.98_{0.68}$ & $85.40_{3.78}$ & $92.86_{1.79}$ & $84.91_{0.59}$ & $80.16_{1.47}$ \\
    \modeladapteroursScalarSplit & $88.85_{0.22}$ & $80.42_{0.06}$ & $81.85_{0.21}$ & $69.91_{1.15}$ & $61.54_{0.00}$ & $82.00_{3.08}$ & $90.00_{1.60}$ & $84.04_{1.66}$ & $79.83_{1.00}$ \\
    \modeladapteroursVecShare & $88.28_{0.23}$ & $\bm{80.76}_{0.58}$ & \underline{$\bm{83.08}$}$_{0.31}$ & $69.59_{1.89}$ & $62.98_{0.68}$ & \underline{$\bm{87.80}$}$_{1.10}$ & $91.07_{1.79}$ & $85.70_{0.32}$ & \underline{$\bm{81.16}$}$_{0.86}$ \\
    \modeladapteroursScalarShare & $88.85_{0.22}$ & $80.70_{0.04}$ & $82.13_{0.21}$ & \underline{$\bm{70.19}$}$_{0.26}$ & $62.98_{0.68}$ & $83.60_{2.88}$ & $91.07_{2.82}$ & $84.84_{1.02}$ & $80.54_{1.02}$ \\        
    \bottomrule
    \end{tabular}
    }
    \caption{Evaluation results on SuperGLUE using $\text{RoBERTa\textsubscript{LARGE}}$.}
    \label{tab:results-superglue-large}
\end{table*}

\begin{table*}[h]
    \centering
    \resizebox{1\linewidth}{!}{%
    \begin{tabular}{lccccc|c}
    \toprule
    \textbf{Model} & \textbf{Sectors} & \textbf{Pillars 1D}  & \textbf{Subpillars 1D} & \textbf{Pillars 2D} & \textbf{Subpillars 2D} &\textbf{Avg.} \\\midrule
    
    \modelfine & $72.99_{0.17}$ & $51.38_{0.39}$ & $44.84_{0.89}$ & $61.90_{0.20}$ & $43.49_{0.86}$ & $54.92_{0.50}$ \\
    \modeladapter & $72.29_{0.59}$ & $49.31_{1.27}$ & \underline{$45.25$}$_{0.03}$ & $62.58_{0.67}$ & $44.36_{0.66}$ & $54.76_{0.65}$ \\
    \modelpropetl & $73.20_{0.32}$ & $51.58_{0.40}$ & $45.10_{0.92}$ & $61.52_{2.29}$ & $41.98_{0.70}$ & $54.68_{0.92}$ \\
    \modelcompact & $61.77_{12.63}$ & $8.17_{5.92}$ & $6.37_{11.00}$ & $20.39_{24.91}$ & $15.36_{2.71}$ & $22.41_{11.43}$ \\
    \rev{\modelia} & $\rev{64.72_{1.83}}$ & $\rev{38.26_{7.27}}$ & $\rev{26.77_{2.79}}$ & $\rev{55.57_{1.48}}$ & $\rev{31.11_{2.53}}$ & $\rev{43.29_{3.18}}$ \\
    \modellora & $72.22_{0.82}$ & $52.15_{0.25}$ & $0.00_{0.00}$ & $61.34_{1.35}$ & $0.00_{0.00}$ & $37.14_{0.48}$ \\    \midrule
    \modelfineMT & $59.04_{7.86}$ & $22.95_{12.78}$ & $10.75_{5.31}$ & $29.76_{21.25}$ & $9.65_{1.25}$ & $26.43_{9.69}$ \\
    \modeladapterMT & $65.66_{7.13}$ & $37.65_{11.25}$ & $28.51_{7.80}$ & $43.40_{16.06}$ & $27.44_{1.68}$ & $40.53_{8.78}$ \\
    \modelpropetlMT & $70.56_{1.06}$ & $41.58_{6.27}$ & $35.91_{3.46}$ & $42.20_{14.55}$ & $29.67_{6.92}$ & $43.98_{6.45}$ \\
    \modelhyper & $47.74_{20.72}$ & $29.06_{11.76}$ & $22.16_{8.44}$ & $35.92_{17.37}$ & $22.58_{10.58}$ & $31.49_{13.77}$ \\
    \modelhyperPP & $0.00_{0.00}$ & $0.00_{0.00}$ & $0.00_{0.00}$ & $0.00_{0.00}$ & $0.00_{0.00}$ & $0.00_{0.00}$ \\
    \midrule
    \modeladapterfusion & $72.53_{0.45}$ & $51.33_{0.23}$ & $43.75_{0.52}$ & $62.31_{0.25}$ & $42.78_{2.11}$ & $54.54_{0.71}$ \\
    \rev{\modeladaptersoup} & $\rev{52.54_{1.61}}$ & $\rev{24.07_{2.18}}$ & $\rev{20.62_{0.28}}$ & $\rev{31.16_{1.40}}$ & $\rev{12.84_{0.49}}$ & $\rev{28.25_{1.19}}$ \\
    \modeladapteroursVecSplit & \underline{$\bm{73.32}$}$_{0.08}$ & \underline{$\bm{53.94}$}$_{0.13}$ & $44.14_{0.75}$ & \underline{$\bm{63.89}$}$_{0.16}$ & $44.75_{0.47}$ & \underline{$\bm{56.01}$}$_{0.32}$ \\
    \modeladapteroursScalarSplit & $72.56_{0.20}$ & $50.59_{0.10}$ & $44.62_{0.00}$ & $62.66_{0.00}$ & $45.16_{0.00}$ & $55.12_{0.06}$ \\
    \modeladapteroursVecShare & $73.18_{0.04}$ & $51.41_{0.36}$ & $44.10_{0.09}$ & $63.37_{0.02}$ & \underline{$\bm{45.43}$}$_{0.24}$ & $55.50_{0.15}$ \\
    \modeladapteroursScalarShare & $73.02_{0.20}$ & $50.84_{0.30}$ & \bm{$44.88$}$_{0.39}$ & $62.87_{0.01}$ & $44.45_{0.02}$ & $55.21_{0.18}$ \\
    
    \bottomrule
    \end{tabular}
    } 
    \caption{Evaluation results on HumSet using $\text{XLM-R\textsubscript{LARGE}}$.}
    \label{tab:results-humset-avg-large}
\end{table*}

\begin{sidewaystable*}[]
    \centering
    \resizebox{1\textwidth}{!}{%
\begin{tabular}{lcccccccc|ccccccc|c}
    \toprule
    \textbf{Model} & \textbf{MNLI} & \textbf{QQP}  & \textbf{QNLI} & \textbf{SST-2} & \textbf{STS-B} & \textbf{MRPC} & \textbf{RTE}  & \textbf{CoLA} & \textbf{ReCoRD} & \textbf{MultiRC} & \textbf{BoolQ} & \textbf{WiC}  & \textbf{WSC}  & \textbf{COPA} & \textbf{CB}   & \textbf{Avg.} \\\midrule
    
\modelfine & $86.61_{0.51}$ & \underline{$90.32$}$_{0.15}$ & $91.78_{0.28}$ & $93.33_{0.48}$ & $90.53_{0.22}$ & $86.94_{1.52}$ & $73.47_{2.05}$ & $58.46_{4.03}$ & $71.61_{0.85}$ & $71.64_{1.15}$ & $76.80_{1.34}$ & $66.38_{2.08}$ & \underline{$63.46$}$_{0.00}$ & $68.60_{6.74}$ & $81.96_{4.33}$ & $78.12_{1.72}$ \\ 
\modeladapter & $86.50_{0.33}$ & $90.18_{0.11}$ & $92.25_{0.19}$ & $93.65_{0.71}$ & $90.23_{0.41}$ & $86.64_{1.07}$ & $72.89_{2.54}$ & $58.28_{2.50}$ & $79.02_{0.62}$ & $72.84_{0.48}$ & $76.71_{1.38}$ & $65.58_{1.56}$ & \underline{$63.46$}$_{0.00}$ & $70.20_{4.13}$ & $84.82_{3.18}$ & $78.88_{1.28}$ \\ 
\modelpropetl & $86.19_{0.25}$ & $88.88_{0.48}$ & $92.05_{0.80}$ & $93.81_{0.72}$ & $90.03_{0.35}$ & $85.93_{1.22}$ & $74.19_{2.03}$ & $59.29_{2.07}$ & \underline{$80.29$}$_{0.24}$ & \underline{$73.07$}$_{0.49}$ & $76.58_{0.78}$ & $66.60_{1.65}$ & \underline{$63.46$}$_{0.00}$ & $70.60_{3.44}$ & $84.46_{3.86}$ & $79.05_{1.21}$ \\ 
\modelcompact & $85.62_{0.42}$ & $88.84_{0.70}$ & $91.79_{0.39}$ & $93.58_{0.34}$ & $89.67_{0.54}$ & $87.21_{0.61}$ & $72.02_{2.21}$ & $58.49_{2.58}$ & $77.69_{2.67}$ & $70.44_{0.57}$ & $75.88_{0.96}$ & $66.46_{1.63}$ & \underline{$63.46$}$_{0.00}$  & $68.30_{4.00}$ & $87.68_{3.62}$ & $78.48_{1.42}$ \\ 
\rev{\modelia} & $\rev{84.24_{1.01}}$ & $\rev{88.37_{0.20}}$ & $\rev{90.57_{0.38}}$ & $\rev{93.35_{0.30}}$ & $\rev{89.93_{0.30}}$ & $\rev{87.11_{1.14}}$ & $\rev{72.56_{2.23}}$ & $\rev{56.57_{5.39}}$ & $\rev{75.27_{0.23}}$ & $\rev{70.32_{0.49}}$ & $\rev{76.31_{0.79}}$ & $\rev{67.07_{1.68}}$ & $\rev{63.35_{0.32}}$ & $\rev{69.30_{3.37}}$ & $\rev{87.32_{4.57}}$ & $\rev{78.11_{1.49}}$ \\
\bottomrule
\end{tabular}
}
    \caption{Evaluation results on the combination of all GLUE and SuperGLUE tasks using $\text{RoBERTa\textsubscript{BASE}}$. (Top) STL models, only learning a single task at a time. (Middle) Joint MTL methods, learning all tasks simultaneously. (Bottom) Two-stage MTL methods, composing the knowledge of several source adapters. The overall best results are \underline{underlined}, and the best results among the two-stage MTL models are shown in \textbf{bold}.}
\label{tab:results-gsg-full}

\centering
\resizebox{1\textwidth}{!}{%
\begin{tabular}{lcccccccc|ccccccc|c}
\toprule
\textbf{Model} & \textbf{MNLI} & \textbf{QQP}  & \textbf{QNLI} & \textbf{SST-2} & \textbf{STS-B} & \textbf{MRPC} & \textbf{RTE}  & \textbf{CoLA} & \textbf{ReCoRD} & \textbf{MultiRC} & \textbf{BoolQ} & \textbf{WiC}  & \textbf{WSC}  & \textbf{COPA} & \textbf{CB}   & \textbf{Avg.} \\\midrule

\modelfine & $89.57_{0.36}$ & $89.75_{1.03}$ & $93.91_{0.43}$ & $95.30_{0.65}$ & $91.89_{0.35}$ & $86.27_{1.15}$ & $81.52_{3.19}$ & $60.15_{2.89}$ & $81.60_{1.25}$ & $79.03_{0.02}$ & $81.65_{0.30}$ & $69.72_{2.16}$ & $63.46_{0.00}$ & $52.00_{8.28}$ & $90.36_{2.99}$ & $80.41_{1.67}$ \\ 
\modeladapter & $89.62_{0.18}$ & $89.87_{0.67}$ & $94.13_{0.06}$ & $95.24_{0.08}$ & $91.81_{0.29}$ & $87.82_{2.11}$ & $81.23_{2.92}$ & $64.07_{1.97}$ & $88.52_{0.09}$ & $80.73_{0.69}$ & $82.36_{0.72}$ & $69.16_{1.31}$ & $63.25_{0.64}$ & $71.90_{13.63}$ & $92.68_{1.78}$ & $82.83_{1.81}$ \\ 
\modelpropetl & $89.78_{0.24}$ & $89.23_{0.77}$ & $94.32_{0.09}$ & $95.41_{0.00}$ & $91.45_{0.39}$ & $87.65_{0.73}$ & $84.55_{2.14}$ & $65.85_{2.10}$ & $87.86_{2.59}$ & \underline{$81.19$}$_{0.99}$ & $81.61_{0.86}$ & $69.62_{2.16}$ & $63.46_{0.00}$ & $69.00_{18.96}$ & $94.11_{4.04}$ & $83.01_{2.40}$ \\ 
\modelcompact & $89.15_{0.67}$ & $87.33_{2.39}$ & $92.93_{1.42}$ & $95.41_{0.00}$ & $91.46_{0.35}$ & $87.84_{1.23}$ & $79.71_{4.58}$ & $65.66_{2.08}$ & $88.34_{0.97}$ & $79.18_{0.29}$ & $79.53_{6.13}$ & $69.26_{1.51}$ & $62.26_{1.43}$ & $79.00_{9.74}$ & $87.50_{7.48}$ & $82.30_{2.69}$ \\
\bottomrule
\end{tabular}
}
\caption{Evaluation results on the combination of all GLUE and SuperGLUE tasks using $\text{RoBERTa\textsubscript{LARGE}}$.}
\label{tab:results-gsg-full-large}
\end{sidewaystable*}

\begin{table*}[b]
\centering
\resizebox{1\textwidth}{!}{%
\begin{tabular}{lccccccccc|c}
\toprule
\textbf{Model} & \textbf{Samples} & \textbf{MNLI} & \textbf{QQP}  & \textbf{QNLI} & \textbf{SST-2} & \textbf{STS-B} & \textbf{MRPC} & \textbf{RTE}  & \textbf{CoLA}  & \textbf{Avg.} \\\midrule

\modeladapter & 4 & $33.65_{1.39}$ & $63.27_{0.11}$ & $50.53_{0.04}$ & $50.92_{0.00}$ & $32.12_{9.28}$ & $68.38_{0.00}$ & $52.71_{0.00}$ & $2.93_{3.88}$ & $44.31_{1.84}$ \\
\modeladapter & 16 & $34.78_{0.58}$ & $63.18_{0.00}$ & $50.46_{0.20}$ & $57.18_{1.23}$ & $55.53_{10.12}$ & $68.38_{0.00}$ & $53.72_{1.29}$ & $0.25_{0.56}$ & $47.94_{1.75}$ \\
\modeladapter & 32 & $33.56_{0.66}$ & $63.18_{0.00}$ & $51.86_{0.33}$ & $70.46_{2.25}$ & $73.78_{1.30}$ & $68.38_{0.00}$ & $54.58_{1.81}$ & $0.00_{0.00}$ & $51.98_{0.80}$ \\
\modeladapter & 100 & $40.71_{2.67}$ & $71.74_{0.50}$ & $58.77_{4.13}$ & $85.00_{2.25}$ & $82.51_{1.21}$ & $73.09_{1.27}$ & $56.17_{1.95}$ & $21.69_{3.94}$ & $61.21_{2.24}$ \\
\modeladapter & All & $86.50_{0.33}$ & $90.18_{0.11}$ & $92.25_{0.19}$ & $93.65_{0.71}$ & $90.23_{0.41}$ & $86.64_{1.07}$ & $72.89_{2.54}$ & $58.28_{2.50}$ & $83.83_{0.98}$ \\
\midrule
\modeladapterfusion & 4 & $33.94_{2.09}$ & $72.01_{5.39}$ & $52.36_{2.75}$ & $50.92_{0.00}$ & $77.17_{2.44}$ & $72.99_{4.28}$ & $52.78_{0.16}$ & $2.79_{3.54}$ & $51.87_{2.58}$ \\
\modeladapterfusion & 16 & $49.12_{2.76}$ & $76.26_{1.20}$ & $61.95_{11.04}$ & $59.29_{6.12}$ & $83.51_{1.79}$ & $78.28_{0.37}$ & $60.65_{2.27}$ & $0.92_{1.82}$ & $58.75_{3.42}$ \\
\bottomrule
\end{tabular}
}
\caption{Complete few-shot transfer learning results on GLUE with $k=\text{\{4,16,32,100\}}$ training samples for each target task using $\text{RoBERTa\textsubscript{BASE}}$.}
\label{tab:results-few-glue-full}
\end{table*}

\begin{table*}[t]
\centering
\resizebox{1\textwidth}{!}{%
\begin{tabular}{lccccccccc|c}
\toprule
\textbf{Model} & \textbf{Samples} & \textbf{ReCoRD} & \textbf{Multi} & \textbf{BoolQ} & \textbf{WiC}  & \textbf{WSC}  & \textbf{COPA} & \textbf{CB} & \textbf{RTE}  & \textbf{Avg.} \\\midrule

\modeladapter & 4 & $9.65_{2.79}$ & $24.92_{6.71}$ & $62.05_{0.27}$ & $49.44_{1.26}$ & $41.92_{12.04}$ & $50.20_{3.63}$ & $62.14_{8.12}$ & $52.71_{0.00}$ & $44.13_{4.35}$ \\
\modeladapter & 16 & $13.82_{6.06}$ & $37.48_{8.48}$ & $62.17_{0.00}$ & $50.53_{1.18}$ & $42.50_{5.46}$ & $53.00_{5.48}$ & $69.29_{2.93}$ & $53.72_{1.29}$ & $47.81_{3.86}$ \\
\modeladapter & 32 & $17.64_{12.76}$ & $38.55_{3.74}$ & $62.16_{0.03}$ & $52.26_{1.78}$ & $36.54_{0.00}$ & $51.20_{2.39}$ & $70.71_{1.60}$ & $54.58_{1.81}$ & $47.95_{3.01}$ \\
\modeladapter & 100 & $37.69_{2.61}$ & $51.56_{3.89}$ & $61.51_{1.27}$ & $54.04_{1.01}$ & $50.38_{10.12}$ & $58.40_{5.18}$ & $73.93_{4.11}$ & $56.17_{1.95}$ & $55.46_{3.77}$ \\
\modeladapter & All & $79.02_{0.62}$ & $72.84_{0.48}$ & $76.71_{1.38}$ & $65.58_{1.56}$ & $63.46_{0.00}$ & $70.20_{4.13}$ & $84.82_{3.18}$ & $72.89_{2.54}$ & $73.19_{1.74}$ \\
\midrule
\modeladapterfusion & 4 & $8.51_{2.73}$ & $44.50_{24.40}$ & $62.16_{0.03}$ & $50.31_{1.04}$ & $38.08_{3.44}$ & $50.40_{2.19}$ & $51.07_{2.40}$ & $52.64_{1.31}$ & $44.71_{4.69}$ \\
\modeladapterfusion & 16 & $13.71_{10.75}$ & $48.86_{14.98}$ & $62.12_{0.27}$ & $50.16_{1.84}$ & $38.46_{4.30}$ & $56.80_{7.22}$ & $67.86_{3.99}$ & $52.92_{3.71}$ & $48.86_{5.88}$ \\
\bottomrule
\end{tabular}
}
\caption{Complete few-shot transfer learning results on SuperGLUE with $k=\text{\{4,16,32,100\}}$ training samples for each target task using $\text{RoBERTa\textsubscript{BASE}}$.}
\label{tab:results-few-superglue-full}
\end{table*}

\begin{table*}[t]
    \centering
    \resizebox{1\linewidth}{!}{%
    \begin{tabular}{lcccccc|c}
    \toprule
    \textbf{Model} & \textbf{Samples} & \textbf{Sectors} & \textbf{Pillars 1D}  & \textbf{Subpillars 1D} & \textbf{Pillars 2D} & \textbf{Subpillars 2D} &\textbf{Avg.} \\\midrule
    
    \modeladapter & 4 & $5.78_{2.05}$ & $4.21_{1.16}$ & $0.69_{0.34}$ & $11.07_{2.07}$ & $3.58_{0.49}$ & $5.07_{1.22}$ \\
    \modeladapter & 16 & $8.22_{6.21}$ & $2.59_{2.28}$ & $0.78_{0.42}$ & $8.42_{4.12}$ & $2.59_{1.34}$ & $4.52_{2.87}$ \\
    \modeladapter & 32 & $4.65_{1.88}$ & $2.30_{2.71}$ & $0.82_{0.15}$ & $5.96_{7.43}$ & $2.97_{1.52}$ & $3.34_{2.74}$ \\
    \modeladapter & 100 & $44.26_{1.22}$ & $10.59_{9.70}$ & $0.00_{0.00}$ & $25.26_{1.36}$ & $0.01_{0.02}$ & $16.02_{2.46}$ \\
    \modeladapter & All & $71.38_{0.28}$ & $51.02_{1.23}$ & $43.26_{0.82}$ & $61.43_{0.91}$ & $42.46_{0.51}$ & $53.91_{0.75}$ \\
    \midrule
    \modeladapterfusion & 4 & $13.60_{1.29}$ & $7.20_{2.19}$ & $2.45_{0.37}$ & $16.24_{2.77}$ & $8.16_{1.00}$ & $9.53_{1.53}$ \\
    \modeladapterfusion & 16 & $13.27_{1.27}$ & $8.38_{0.99}$ & $2.17_{0.67}$ & $15.98_{2.41}$ & $7.63_{0.73}$ & $9.48_{1.21}$ \\
\bottomrule
    \end{tabular}
    } 
    \caption{Complete few-shot transfer learning results on HumSet with $k=\text{\{4,16,32,100\}}$ training samples for each target task using $\text{XLM-R\textsubscript{BASE}}$.}
    \label{tab:results-few-humset-avg}
\end{table*}

\begin{sidewaystable*}[t]
\centering
\resizebox{1\textwidth}{!}{%
\begin{tabular}{lcccccccccccccccc|c}
\toprule
\textbf{Model} & \textbf{Samples} & \textbf{MNLI} & \textbf{QQP}  & \textbf{QNLI} & \textbf{SST-2} & \textbf{STS-B} & \textbf{MRPC} & \textbf{RTE}  & \textbf{CoLA} & \textbf{ReCoRD} & \textbf{Multi} & \textbf{BoolQ} & \textbf{WiC}  & \textbf{WSC}  & \textbf{COPA} & \textbf{CB}   & \textbf{Avg.} \\\midrule

\modeladapter & 4 & $33.65_{1.39}$ & $63.27_{0.11}$ & $50.53_{0.04}$ & $50.92_{0.00}$ & $32.12_{9.28}$ & $68.38_{0.00}$ & $52.71_{0.00}$ & $2.93_{3.88}$ & $9.65_{2.79}$ & $24.92_{6.71}$ & $62.05_{0.27}$ & $49.44_{1.26}$ & $41.92_{12.04}$ & $50.20_{3.63}$ & $62.14_{8.12}$ & $43.66_{3.3}$ \\
\modeladapter & 16 & $34.78_{0.58}$ & $63.18_{0.00}$ & $50.46_{0.20}$ & $57.18_{1.23}$ & $55.53_{10.12}$ & $68.38_{0.00}$ & $53.72_{1.29}$ & $0.25_{0.56}$ & $13.82_{6.06}$ & $37.48_{8.48}$ & $62.17_{0.00}$ & $50.53_{1.18}$ & $42.50_{5.46}$ & $53.00_{5.48}$ & $69.29_{2.93}$ & $47.48_{2.9}$ \\
\modeladapter & 32 & $33.56_{0.66}$ & $63.18_{0.00}$ & $51.86_{0.33}$ & $70.46_{2.25}$ & $73.78_{1.30}$ & $68.38_{0.00}$ & $54.58_{1.81}$ & $0.00_{0.00}$ & $17.64_{12.76}$ & $38.55_{3.74}$ & $62.16_{0.03}$ & $52.26_{1.78}$ & $36.54_{0.00}$ & $51.20_{2.39}$ & $70.71_{1.60}$ & $49.66_{1.91}$ \\
\modeladapter & 100 & $40.71_{2.67}$ & $71.74_{0.50}$ & $58.77_{4.13}$ & $85.00_{2.25}$ & $82.51_{1.21}$ & $73.09_{1.27}$ & $56.17_{1.95}$ & $21.69_{3.94}$ & $37.69_{2.61}$ & $51.56_{3.89}$ & $61.51_{1.27}$ & $54.04_{1.01}$ & $50.38_{10.12}$ & $58.40_{5.18}$ & $73.93_{4.11}$ & $58.48_{3.07}$ \\
\modeladapter & All & $86.50_{0.33}$ & $90.18_{0.11}$ & $92.25_{0.19}$ & $93.65_{0.71}$ & $90.23_{0.41}$ & $86.64_{1.07}$ & $72.89_{2.54}$ & $58.28_{2.50}$ & $79.02_{0.62}$ & $72.84_{0.48}$ & $76.71_{1.38}$ & $65.58_{1.56}$ & $63.46_{0.00}$ & $70.20_{4.13}$ & $84.82_{3.18}$ & $78.88_{1.28}$ \\
\bottomrule
\end{tabular}
}
\caption{Complete few-shot transfer learning results on the combination of all GLUE and SuperGLUE tasks with $k=\text{\{4,16,32,100\}}$ training samples for each target task using $\text{RoBERTa\textsubscript{BASE}}$.}
\label{tab:results-few-gsg-full}
\end{sidewaystable*}

\begin{table*}[t]
\centering
\resizebox{1\textwidth}{!}{%
\begin{tabular}{lccccccccc|c}
\toprule
\textbf{Model} & \textbf{Samples} & \textbf{MNLI} & \textbf{QQP}  & \textbf{QNLI} & \textbf{SST-2} & \textbf{STS-B} & \textbf{MRPC} & \textbf{RTE}  & \textbf{CoLA}  & \textbf{Avg.} \\\midrule

\modeladapter & 4 & $34.09_{0.48}$ & $62.00_{2.54}$ & $50.46_{1.12}$ & $50.92_{0.00}$ & $10.02_{2.34}$ & $68.33_{0.11}$ & $51.48_{2.74}$ & $3.47_{3.01}$ & $41.35_{1.54}$ \\
\modeladapter & 16 & $35.12_{1.00}$ & $63.11_{0.18}$ & $49.59_{0.24}$ & $59.38_{3.42}$ & $12.41_{5.51}$ & $68.38_{0.00}$ & $52.64_{1.06}$ & $2.55_{3.07}$ & $42.90_{1.81}$ \\
\modeladapter & 32 & $34.05_{0.94}$ & $63.88_{1.40}$ & $51.30_{0.98}$ & $74.70_{2.59}$ & $27.16_{13.89}$ & $68.77_{0.71}$ & $51.62_{1.75}$ & $7.47_{10.36}$ & $47.37_{4.08}$ \\
\modeladapter & 100 & $41.39_{2.59}$ & $71.35_{0.81}$ & $53.75_{1.18}$ & $83.67_{2.22}$ & $76.84_{4.07}$ & $69.07_{1.48}$ & $56.97_{2.30}$ & $30.96_{5.72}$ & $60.50_{2.55}$ \\
\modeladapter & All & $89.62_{0.18}$ & $89.87_{0.67}$ & $94.13_{0.06}$ & $95.24_{0.08}$ & $91.81_{0.29}$ & $87.82_{2.11}$ & $81.23_{2.92}$ & $64.07_{1.97}$ & $86.72_{1.04}$ \\
\midrule
\modeladapterfusion & 4 & $39.26_{6.48}$ & $79.28_{0.71}$ & $65.13_{11.67}$ & $51.03_{0.23}$ & $76.40_{12.07}$ & $69.95_{2.76}$ & $54.08_{3.07}$ & $4.93_{1.85}$ & $55.01_{4.85}$ \\
\modeladapterfusion & 16 & $49.94_{8.89}$ & $80.37_{0.13}$ & $78.85_{3.67}$ & $56.65_{3.82}$ & $83.96_{0.85}$ & $77.50_{1.62}$ & $70.47_{4.04}$ & $16.08_{3.34}$ & $64.23_{3.29}$ \\
\bottomrule
\end{tabular}
}
\caption{Complete few-shot transfer learning results on GLUE with $k=\text{\{4,16,32,100\}}$ training samples for each target task using $\text{RoBERTa\textsubscript{LARGE}}$.}
\label{tab:results-few-glue-full-large}
\end{table*}

\begin{table*}[t]
\centering
\resizebox{1\textwidth}{!}{%
\begin{tabular}{lccccccccc|c}
\toprule
\textbf{Model} & \textbf{Samples} & \textbf{ReCoRD} & \textbf{Multi} & \textbf{BoolQ} & \textbf{WiC}  & \textbf{WSC}  & \textbf{COPA} & \textbf{CB} & \textbf{RTE}  & \textbf{Avg.} \\\midrule

\modeladapter & 4 & $15.58_{3.93}$ & $31.78_{15.80}$ & $61.83_{0.58}$ & $49.75_{0.56}$ & $50.38_{8.93}$ & $49.60_{5.59}$ & $53.93_{4.96}$ & $51.48_{2.74}$ & $45.54_{5.39}$ \\
\modeladapter & 16 & $17.42_{7.21}$ & $40.46_{3.08}$ & $61.64_{0.54}$ & $51.38_{1.37}$ & $54.04_{2.30}$ & $53.60_{5.46}$ & $61.07_{3.19}$ & $52.64_{1.06}$ & $49.03_{3.03}$ \\
\modeladapter & 32 & $22.04_{14.70}$ & $41.11_{5.21}$ & $62.17_{0.01}$ & $52.88_{1.91}$ & $47.69_{3.76}$ & $66.20_{7.60}$ & $67.50_{2.33}$ & $51.62_{1.75}$ & $51.40_{4.66}$ \\
\modeladapter & 100 & $31.01_{19.22}$ & $51.93_{4.94}$ & $62.17_{0.00}$ & $55.96_{2.03}$ & $52.88_{5.81}$ & $65.20_{13.86}$ & $82.14_{5.65}$ & $56.97_{2.30}$ & $57.28_{6.73}$ \\
\modeladapter & All & $88.52_{0.09}$ & $80.73_{0.69}$ & $82.36_{0.72}$ & $69.16_{1.31}$ & $63.25_{0.64}$ & $71.90_{13.63}$ & $92.68_{1.78}$ & $81.23_{2.92}$ & $78.73_{2.72}$ \\
\midrule
\modeladapterfusion & 4 & $19.21_{4.17}$ & $24.07_{20.35}$ & $61.77_{0.18}$ & $50.63_{1.49}$ & $43.27_{12.03}$ & $57.00_{7.42}$ & $61.43_{11.75}$ & $52.71_{0.00}$ & $46.26_{7.17}$ \\
\modeladapterfusion & 16 & $14.28_{5.34}$ & $28.09_{4.31}$ & $61.51_{0.35}$ & $51.10_{3.13}$ & $47.31_{9.26}$ & $66.20_{12.44}$ & $77.86_{4.48}$ & $53.21_{1.37}$ & $49.95_{5.09}$ \\
\bottomrule
\end{tabular}
}
\caption{Complete few-shot transfer learning results on SuperGLUE with $k=\text{\{4,16,32,100\}}$ training samples for each target task using $\text{RoBERTa\textsubscript{LARGE}}$.}
\label{tab:results-few-superglue-full-large}
\end{table*}

\begin{table*}[t]
    \centering
    \resizebox{1\linewidth}{!}{%
    \begin{tabular}{lcccccc|c}
    \toprule
    \textbf{Model} & \textbf{Samples} & \textbf{Sectors} & \textbf{Pillars 1D}  & \textbf{Subpillars 1D} & \textbf{Pillars 2D} & \textbf{Subpillars 2D} &\textbf{Avg.} \\\midrule
    
    \modeladapter & 4 & $4.80_{0.60}$ & $4.33_{0.18}$ & $0.60_{0.08}$ & $10.87_{1.72}$ & $2.56_{0.56}$ & $4.63_{0.63}$ \\
    \modeladapter & 16 & $7.12_{2.11}$ & $1.35_{1.85}$ & $0.45_{0.32}$ & $11.08_{0.59}$ & $2.82_{0.82}$ & $4.56_{1.14}$ \\
    \modeladapter & 32 & $6.60_{3.21}$ & $0.58_{0.54}$ & $0.52_{0.24}$ & $11.82_{1.44}$ & $2.40_{0.92}$ & $4.39_{1.27}$ \\
    \modeladapter & 100 & $24.66_{13.33}$ & $12.38_{3.57}$ & $0.00_{0.00}$ & $16.21_{1.14}$ & $3.13_{2.91}$ & $11.27_{4.19}$ \\
    \modeladapter & All & $72.29_{0.59}$ & $49.31_{1.27}$ & $45.25_{0.03}$ & $62.58_{0.67}$ & $44.36_{0.66}$ & $54.76_{0.65}$ \\
    \midrule
    \modeladapterfusion & 4 & $12.43_{2.84}$ & $7.58_{0.95}$ & $2.11_{0.12}$ & $14.59_{0.57}$ & $7.10_{1.13}$ & $8.76_{1.12}$ \\
    \modeladapterfusion & 16 & $11.06_{2.41}$ & $6.49_{2.35}$ & $2.30_{0.26}$ & $13.08_{1.04}$ & $6.33_{1.79}$ & $7.85_{1.57}$ \\
\bottomrule
    \end{tabular}
    } 
    \caption{Complete few-shot transfer learning results on HumSet with $k=\text{\{4,16,32,100\}}$ training samples for each target task using $\text{XLM-R\textsubscript{LARGE}}$.}
    \label{tab:results-few-humset-avg-large}
\end{table*}

\begin{sidewaystable*}[t]
\centering
\resizebox{1\textwidth}{!}{%
\begin{tabular}{lcccccccccccccccc|c}
\toprule
\textbf{Model} & \textbf{Samples} & \textbf{MNLI} & \textbf{QQP}  & \textbf{QNLI} & \textbf{SST-2} & \textbf{STS-B} & \textbf{MRPC} & \textbf{RTE}  & \textbf{CoLA} & \textbf{ReCoRD} & \textbf{Multi} & \textbf{BoolQ} & \textbf{WiC}  & \textbf{WSC}  & \textbf{COPA} & \textbf{CB}   & \textbf{Avg.} \\\midrule

\modeladapter & 4 & $34.09_{0.48}$ & $62.00_{2.54}$ & $50.46_{1.12}$ & $50.92_{0.00}$ & $10.02_{2.34}$ & $68.33_{0.11}$ & $51.48_{2.74}$ & $3.47_{3.01}$ & $15.58_{3.93}$ & $31.78_{15.80}$ & $61.83_{0.58}$ & $49.75_{0.56}$ & $50.38_{8.93}$ & $49.60_{5.59}$ & $53.93_{4.96}$ & $42.91_{3.51}$ \\
\modeladapter & 16 & $35.12_{1.00}$ & $63.11_{0.18}$ & $49.59_{0.24}$ & $59.38_{3.42}$ & $12.41_{5.51}$ & $68.38_{0.00}$ & $52.64_{1.06}$ & $2.55_{3.07}$ & $17.42_{7.21}$ & $40.46_{3.08}$ & $61.64_{0.54}$ & $51.38_{1.37}$ & $54.04_{2.30}$ & $53.60_{5.46}$ & $61.07_{3.19}$ & $45.52_{2.51}$ \\
\modeladapter & 32 & $34.05_{0.94}$ & $63.88_{1.40}$ & $51.30_{0.98}$ & $74.70_{2.59}$ & $27.16_{13.89}$ & $68.77_{0.71}$ & $51.62_{1.75}$ & $7.47_{10.36}$ & $22.04_{14.70}$ & $41.11_{5.21}$ & $62.17_{0.01}$ & $52.88_{1.91}$ & $47.69_{3.76}$ & $66.20_{7.60}$ & $67.50_{2.33}$ & $49.24_{4.54}$ \\
\modeladapter & 100 & $41.39_{2.59}$ & $71.35_{0.81}$ & $53.75_{1.18}$ & $83.67_{2.22}$ & $76.84_{4.07}$ & $69.07_{1.48}$ & $56.97_{2.30}$ & $30.96_{5.72}$ & $31.01_{19.22}$ & $51.93_{4.94}$ & $62.17_{0.00}$ & $55.96_{2.03}$ & $52.88_{5.81}$ & $65.20_{13.86}$ & $82.14_{5.65}$ & $59.02_{4.79}$ \\
\modeladapter & All & $89.62_{0.18}$ & $89.87_{0.67}$ & $94.13_{0.06}$ & $95.24_{0.08}$ & $91.81_{0.29}$ & $87.82_{2.11}$ & $81.23_{2.92}$ & $64.07_{1.97}$ & $88.52_{0.09}$ & $80.73_{0.69}$ & $82.36_{0.72}$ & $69.16_{1.31}$ & $62.79_{1.57}$ & $71.90_{13.63}$ & $92.68_{1.78}$ & $82.80_{1.87}$ \\
\bottomrule
\end{tabular}
}
\caption{Complete few-shot transfer learning results on the combination of all GLUE and SuperGLUE tasks with $k=\text{\{4,16,32,100\}}$ training samples for each target task using $\text{RoBERTa\textsubscript{LARGE}}$.}
\label{tab:results-few-gsg-full-large}
\end{sidewaystable*}

\end{document}